%% file: main.tex
\documentclass[letterpaper]{article} 
\usepackage{aaai24}  
\usepackage{times}  
\usepackage{helvet}  
\usepackage{courier}  
\usepackage[hyphens]{url}  
\usepackage{graphicx} 
\usepackage{natbib}  
\usepackage{caption} 
\usepackage{microtype}
\usepackage{graphicx}
\usepackage{enumitem}
\usepackage{subfigure}
\usepackage{booktabs}

\usepackage{thm-restate}

\usepackage[algo2e,ruled,vlined,lined,linesnumbered]{algorithm2e}

\usepackage{amsmath}
\usepackage{amssymb}
\usepackage{mathtools}
\usepackage{amsthm}

\usepackage[utf8]{inputenc} 
\usepackage[T1]{fontenc}    

\usepackage{url}            
\usepackage{booktabs}       
\usepackage{amsfonts}       
\usepackage{nicefrac}       
\usepackage{microtype}      
\usepackage{xcolor}         

\usepackage{subfigure}
\usepackage{amsmath}
\usepackage{amssymb}
\usepackage{mathtools}
\usepackage{amsthm}
\usepackage{tcolorbox}
\definecolor{babyblueeyes}{rgb}{0.63, 0.79, 0.95}
\definecolor{celadon}{rgb}{0.67, 0.88, 0.69}

\usepackage{cleveref}

\theoremstyle{theorem}
\newtheorem{theorem}{Theorem}[section]

\newtheorem{corollary}[theorem]{Corollary}
\theoremstyle{definition}

\newtheorem{assumption}{Assumption}
\theoremstyle{remark}
\newtheorem{remark}[theorem]{Remark}
\usepackage{algorithm}
\usepackage{algorithmic}
\DeclareMathOperator*{\argmax}{arg\,max}

\usepackage{tcolorbox}
\usepackage{multirow}
\usepackage{multicol}

\title{On the Role of Server Momentum in Federated Learning}
\author{
    Jianhui Sun\textsuperscript{\rm 1}\equalcontrib,
    Xidong Wu\textsuperscript{\rm 2}\equalcontrib,
    Heng Huang\textsuperscript{\rm 3},
    Aidong Zhang\textsuperscript{\rm 1}
}
\affiliations {
    \textsuperscript{\rm 1}Computer Science, University of Virginia, VA, USA\\
    \textsuperscript{\rm 2}Electrical and Computer Engineering, University of Pittsburgh, PA, USA\\
    \textsuperscript{\rm 3}Computer Science, University of Maryland College Park, MD, USA\\
    js9gu@virginia.edu, xidong\_wu@outlook.com, heng@umd.edu, aidong@virginia.edu
}
\usepackage{bibentry}

\begin{document}

\maketitle

\begin{abstract}
Federated Averaging (FedAvg) is known to experience convergence issues when encountering significant clients system heterogeneity and data heterogeneity. Server momentum has been proposed as an effective mitigation. However, existing server momentum works are restrictive in the momentum formulation, do not properly schedule hyperparameters and focus only on system homogeneous settings, which leaves the role of server momentum still an under-explored problem. In this paper, we propose a general framework for server momentum, that (a) covers a large class of momentum schemes that are unexplored in federated learning (FL), (b) enables a popular stagewise hyperparameter scheduler, (c) allows heterogeneous and asynchronous local computing. We provide rigorous convergence analysis for the proposed framework. To our best knowledge, this is the first work that thoroughly analyzes the performances of server momentum with a hyperparameter scheduler and system heterogeneity. Extensive experiments validate the effectiveness of our proposed framework.
\end{abstract}

\input{subfiles/1-intro.tex}
\input{subfiles/2-background.tex}

\input{subfiles/3-fedgm.tex}

\input{subfiles/4-multistage.tex}
\input{subfiles/5-free.tex}

\input{subfiles/6-experiments.tex}

\input{subfiles/7-conclusion.tex}

\section{Acknowledgments}
This work was partially supported by NSF 2217071, 2213700, 2106913, 2008208, 1955151 at UVA.
This work was partially supported by NSF IIS 2347592, 2348169, 2348159, 2347604, CNS 2347617, CCF 2348306, DBI 2405416 at Pitt and UMD.


\newpage

\bibliography{jianhui}

\newpage
\onecolumn
\appendix
\input{subfiles/8-appendix}

\end{document}

%% file: subfiles/1-intro.tex
\section{Introduction}
\label{sec:intro}

Federated Averaging (FedAvg) \citep{McMahan2017FedAvg}, which runs multiple epochs of Stochastic Gradient Descent (SGD) locally in each client and then averages the local model updates once in a while on the server, is probably the most popular algorithm to solve many federated learning (FL) problems, mainly due to its low communication cost and appealing convergence property.

Though it has seen great empirical success, vanilla FedAvg experiences an unstable and slow convergence when encountering \textit{client drift}, i.e., the local client models move away from globally optimal models due to client heterogeneity \citep{karimireddy2020scaffold}. On the server side, FedAvg is in spirit similar to an SGD with a constant learning rate one and updates the global model relying only on the averaged model update from the current round, thus extremely vulnerable to client drift. Note that in non-FL settings, SGD in its vanilla form has long been replaced by some momentum scheme (e.g. heavy ball momentum (SHB) and Nesterov's accelerated gradient (NAG)) in many tasks, as momentum schemes achieve an impressive training time saving and generalization performance boosting compared to competing optimizers \cite{Sutskever13Init,Wilson2017Generalization}, which promises a great potential to apply momentum in FL settings as well. Incorporating server momentum essentially integrates historical aggregates into the current update, which could conceptually stabilize the global update against dramatic local drifts. 

Though various efforts have been made to understand the role of server momentum in FL, e.g. \citep{Hsu2019MeasuringTE,rothchild20fetchsgd}, it is still largely an under-explored problem due to the following reasons:

(1) Lack of diversity in momentum schemes. Most existing server momentum works only focus on SHB (e.g. FedAvgM \citep{Hsu2019MeasuringTE}). It is unclear whether many momentum schemes that outperformed SHB in non-FL settings can also perform better in FL, and there is no unified analysis for momentum schemes other than SHB. 

(2) No hyperparameter schedule. Properly scheduling hyperparameters is key to train deep models more efficiently and an appropriate selection of server learning rate $\eta_t$ is also important in obtaining optimal convergence rate \citep{yang2021achieving}. Existing works either still employ a constant server learning rate one or consider a $\eta_t$ schedule that is uncommonly used in practice, e.g., polynomially decay (i.e., $\eta_t\propto\frac{1}{t^\alpha}$) \citep{khanduri2021stem}. Moreover, it is known that increasing momentum factor $\beta$ is also a critical technique in deep model training \citep{Sutskever13Init,Smith18DontDecay}, while to our best knowledge, there is no prior work considering time-varying $\beta$ in FL. 

(3) Ignoring client system heterogeneity. Existing works make unrealistic assumptions on system homogeneity and client synchrony, e.g., clients are sampled uniformly at random, all participating clients synchronize at each round $t$, and all clients run identical number of local epochs, none of which holds in most cross device FL deployments \citep{Kairouz21AdvancesProblems}. System heterogeneity (i.e., the violation of above assumptions), alongside with data heterogeneity, is also a main source client drift \citep{karimireddy2020scaffold}. Thus, ignoring it would provide an incomplete understanding of the role of server momentum. 

To address the above limitations, we propose a novel formulation which we refer to as Federated General Momentum (FedGM). FedGM includes the following hyperparameters, learning rate $\eta$, momentum factor $\beta$, and instant discount factor $\nu$. With different specifications of $(\eta,\beta,\nu)$, FedGM subsumes the FL version of many popular momentum schemes, most of which have never been explored in FL yet. 

We further incorporate a widely used hyperparameter scheduler ``constant and drop'' (a.k.a. ``step decay'') in FedGM. We refer to this framework as multistage FedGM. Specifically, with a prespecified set of hyperparameters $\{\eta_s,\beta_s,\nu_s\}_{s=1}^{S}$ and training lengths $\{T_s\}_{s=1}^{S}$, the training process is divided into $S$ stages, and at stage $s$, FedGM with $\{\eta_s,\beta_s,\nu_s\}$ is applied for $T_s$ rounds. Compared to many unrealistic schedule in existing works, ``constant and drop'' is the de-facto scheduler in most model training \citep{Sutskever13Init,He16Res,Huang2017DenseNet}. Multistage FedGM is extremely flexible, as it allows the momentum factor to vary stagewise as well, and subsumes single-stage training as a special case. We provide the convergence analysis of multistage FedGM. Our theoretical results reveal why stagewise training can provide empirically faster convergence. 

Furthermore, in order to understand how server momentum behaves in the presence of system heterogeneity, we propose a framework that we refer to as Autonomous Multistage FedGM, in which clients can do heterogeneous and asynchronous computing. Specifically, we allow each client to (a) update local models based on an asynchronous view of the global model, (b) run a time-varying, client-dependent number of local epochs, and (c) participate at will. We provide convergence analysis of Autonomous Multistage FedGM. Autonomous Multistage FedGM is a more realistic characterization of real-world cross-device FL applications.

Finally, we conduct extensive experiments that validate, (a) FedGM is a much more capable momentum scheme compared with existing FedAvgM in both with and without system heterogeneity settings; and (b) multistage hyperparameter scheduler further improves FedGM effectively.

Our main contributions can be summarized as follow,

\begin{itemize}[leftmargin=*]
    \item We propose FedGM, which is a general framework for server momentum and covers a large class of momentum schemes that are unexplored in FL. We further propose Multistage FedGM, which incorporates a popular hyperparameter scheduler to FedGM.
    \item We show the convergence of multistage FedGM in both full and partial participation settings. We also empirically validate the superiority of multistage FedGM. To our best knowledge, this is the first work that provides convergence analysis of server-side hyperparameter scheduler.
    \item We propose Autonomous Multistage FedGM, which requires much less coordination between server and workers than most existing algorithms, and theoretically analyze its convergence. Our work is the first to study the interplay between server momentum and system heterogeneity.
\end{itemize}

The rest of the paper is organized as follows. In Section \ref{sec:background}, we formally introduce federated optimization. In Section \ref{sec:fedgm}, we introduce Federated General Momentum (FedGM), followed by multistage FedGM and its convergence analysis in Section \ref{sec:stagewise}. In Section \ref{sec:autonomous}, we introduce Autonomous multistage FedGM and provide its convergence analysis. Section \ref{sec:experiments} presents the experimental results. Due to the page limit, we leave related work, all proofs, and additional experimental results to the Appendix.

%% file: subfiles/2-background.tex
\section{Background: FedOPT and FedAvg}
\label{sec:background}

Many FL tasks can be formulated as solving the following optimization problems, 
\begin{equation}
\begin{gathered}
\min_{x\in\mathbb{R}^d}f(x)\triangleq\frac{1}{n}\sum_{i=1}^n f_i(x), \quad
\text{where} \quad f_i(x)=\mathbb{E}_{\xi\sim \mathcal{D}_i} f_i(x,\xi).
\end{gathered}
\label{fed_min_objective}
\end{equation}
where $n$ is the total number of clients, $x$ is the model parameter with $d$ dimension. Each client $i$ has a local data distribution $\mathcal{D}_i$ and a local objective function $f_i(x)=\mathbb{E}_{\xi\sim \mathcal{D}_i} f_i(x,\xi)$. The global objective function is the averaged objective among all clients. $\mathcal{D}_i$ can be very different from $\mathcal{D}_j$ when $i\neq j$.

\begin{algorithm2e}[tb]
\SetAlgoVlined
\KwIn{
Number of clients $n$, objective function $f(x)=\frac{1}{n}\sum_{i=1}^n f_i(x)$, initialization $x_0$, Number of communication rounds $T$, \textbf{local} learning rate $\eta_l$, \textbf{local} number of updates $K$, \textbf{global} hyperparameters $\mathbb{H}$;}
\SetAlgoLined
\For{$t\in\{1,...,T\}$}
{   
    Randomly sample a subset $\mathcal{S}_t$ of clients
    
    Server sends $x_t$ to subset $\mathcal{S}_t$ of clients
    
    \For{each client $i\in \mathcal{S}_t$}
    {
    $\Delta_t^i=\textbf{LocalOPT}\left(i,\eta_l,K,x_t\right)$
    }
    
    Server aggregates $\Delta_t=\frac{1}{\lvert \mathcal{S}_t\rvert}\sum_{i\in \mathcal{S}_t}\Delta_t^i$
    
    $x_{t+1} = \textbf{ServerOPT}\left(x_t,\Delta_t,\mathbb{H}\right)$

}
return $x_T$
\caption{\textbf{FedOPT} \citep{reddi2020adaptive}: A Generic Formulation of Federated Optimization}
\label{alg:fedopt}
\end{algorithm2e}

\begin{algorithm2e}[tp]
\SetAlgoVlined
\KwIn{
client index $i$, data distribution $\mathcal{D}_i$,
\textbf{local} learning rate $\eta_l$, \textbf{local} updating number $K$,
round $t$, \textbf{global} model $x_t$\;
}
\SetAlgoLined
Initialize $x_{t,0}^i\gets x_t$
    
\For{$k\in\{0,1,...,K-1\}$}
{
Randomly sample a batch $\xi_{t,k}^i$ from $\mathcal{D}_i$
    
Compute $g_{t,k}^i=\nabla f_i(x_{t,k}^i, \xi_{t,k}^i)$
    
Update $x_{t,k+1}^i=x_{t,k}^i-\eta_l g_{t,k}^i$
}
$\Delta_t^i=x_t-x_{t,K}^i$
return $\Delta_t^i$
\caption{$\textbf{LocalOPT}\left(i,\eta_l,K,x_t\right)$}
\label{alg:localopt}
\end{algorithm2e}

FedAvg \citep{McMahan2017FedAvg} and its variants are a special case of a more flexible formulation, \textbf{FedOPT} \citep{reddi2020adaptive}, which is formalized in Algorithm \ref{alg:fedopt}. Suppose the total number of rounds is $T$, and the global model parameter is $\{x_t\}_{t=1}^T$. At each round $t$, the server randomly samples a subset of clients $\mathcal{S}_t$ and sends the global model $x_t$ to them. Upon receiving $x_t$, each participating client would do \textbf{LocalOPT} (Algorithm \ref{alg:localopt}). Specifically, each client $i$ would initialize their local model at $x_t$, run $K$ steps of local SGD with local $\eta_l$ and the local model is updated to $x_{t,K}^i$. The client then sends the local model update $\Delta_t^i=x_t-x_{t,K}^i$ back to the server. The server aggregates by averaging, i.e. $\Delta_t=\frac{1}{\lvert \mathcal{S}_t\rvert}\sum_{i\in \mathcal{S}_t}\Delta_t^i$, and then triggers server-side optimization \textbf{ServerOPT}, which takes $x_t$, aggregated model update $\Delta_t$, and a hyperparameter set $\mathbb{H}$ as input, and outputs the next round's global model parameter $x_{t+1}$.

In FedAvg, ServerOPT is simply $x_{t+1} = x_t - \Delta_t$, which is in spirit identical to SGD with a constant learning rate one if viewing $\Delta_t$ as a pseudo gradient.

%% file: subfiles/3-fedgm.tex
\section{FedGM: Federated Learning with General Momentum Acceleration}
\label{sec:fedgm}

Partially due to its equivalence of constant learning rate SGD, FedAvg has two main limitations, (a) it is extremely vulnerable to client drift, as FedAvg relies entirely on its current aggregate $\Delta_t$ and ignores historical directions; (b) FedAvg may not be the best option in many applications, e.g. training large-scale vision or language models \citep{devlin2018bert,dosovitskiy2021VIT} where its counterpart SGD is known to be inferior to momentum or adaptive optimizers in non-FL settings \citep{Wilson2017Generalization,Zhang20Adam_Attention}.

Note that in FedOPT, ServerOPT could in principle be any type of gradient-based optimizers. In non-FL settings, the momentum scheme is known to not only exhibit convincing accelerating effect in training, it has also achieved better generalizability in many tasks than adaptive optimizers like Adam \citep{Wilson2017Generalization,Cutkosky2020MomentumIN}, which provides a strong motivation to incorporate server momentum.

Moreover, server-side momentum basically integrates historical aggregates into the current update and therefore could potentially make the global model more robust to drastic local drifts.

Existing server momentum works mostly focus on one specific type of momentum, i.e. stochastic heavy ball momentum (SHB) \citep{Hsu2019MeasuringTE,rothchild20fetchsgd,khanduri2021stem}, while ignoring many other momentum schemes that outperform SHB in many non-FL settings.

In order to systematically understand the role of server momentum schemes in FL, we propose a new algorithm which we refer to as Federated General Momentum (FedGM). FedGM replaces the ServerOPT $x_{t+1}=x_t-\Delta_t$ in FedAvg with the following,
\begin{equation}
\label{fedgm_formulation}
\begin{gathered}
d_{t+1}=(1-\beta)\Delta_{t}+\beta d_{t},\quad
h_{t+1}=(1-\nu)\Delta_{t}+\nu d_{t+1},\\
x_{t+1}=x_t-\eta h_{t+1}.
\end{gathered}
\end{equation}
where the hyperparameter set $\mathbb{H}=\{\eta,\beta,\nu\}$. $\eta$ is server learning rate, $\beta$ and $\nu$ are two hyperparameters which we call momentum factor and instant discount factor. 

By setting $\nu$ as 0, FedGM becomes FedAvg with two-sided learning rates \citep{yang2021achieving}, i.e., choices of $\eta$ other than 1 is allowed, which we refer to as FedSGD.

By setting $\nu=1$, FedGM becomes FedAvgM \citep{Hsu2019MeasuringTE} (or FedSHB), which essentially applies server SHB, i.e. we update the model by a ``momentum buffer'' $d_{t+1}$. $\beta$ controls how slowly the momentum buffer is updated. FedGM could be interpreted as a $\nu$-weighted average of the FedAvgM update step and the plain FedAvg update step. $\nu$ is thus referred to as instant discount factor. 

FedGM leverages the general formulation of QHM \citep{ma2018quasihyperbolic} and is much more general than just FedAvg and FedAvgM. It subsumes many other momentum variants that are never explored in FL. For example, if $\nu=\beta$, FedGM becomes a new algorithm which can be naturally referred to as FedNAG, i.e. application of the popular optimizer Nesterov's accelerated gradient (NAG) to FL. Specifically, we update model by $x_{t+1}=x_t-\eta\left[(1-\beta)\Delta_{t}+\beta d_{t+1}\right]$, where $d_{t+1}$ is the momentum buffer.

FedGM could further recover the FL version of many other momentum schemes, e.g., SNV \citep{Lessard14SNV}, PID \citep{An18PID}, ASGD \citep{Kidambi18Insufficiency}, and Triple Momentum \citep{VanScoy18Triple}, with different $\eta,\beta,\nu$. Therefore, FedGM describes a family of momentum schemes, most of which have not been studied yet in FL.

%% file: subfiles/4-multistage.tex
\section{Multistage FedGM and Convergence}
\label{sec:stagewise}

\subsection{Proposed Algorithm: Multistage FedGM}

One major limitation in FedGM \eqref{fedgm_formulation} is that all server-side hyperparameters are held constant, which are inconsistent with common practice. Adaptively adjusting hyperparameters throughout the training is key to the success of many optimizers. Learning rate scheduling has been thoroughly studied in non-FL settings, e.g., \citep{Krizhevsky12ImageNet, He16Res, GoyalDGNWKTJH17LargeMinibatch, Smith17Cyclic}. Scheduling other hyperparameters (e.g. momentum factor and batch size) is also shown to be very effective in many settings. For example, \cite{Sutskever13Init,Smith18Bayesian,Smith18DontDecay} show a slowly increasing schedule for the momentum factor $\beta$ is crucial in training deep models faster.

\begin{algorithm2e}[tb]
\SetAlgoVlined
\KwIn{\\
Initialization $x_0$, number of rounds $T$, \textbf{local} learning rate $\eta_l$, \textbf{local} updating number $K$\;
\colorbox{babyblueeyes}{Number of stages $S$, stage lengths $\{T_s\}_{s=1}^{S}$}\;
\colorbox{babyblueeyes}{Stagewise hyperparameters $\{\eta_s,\beta_s,\nu_s\}_{s=1}^{S}$;}
}
\SetAlgoLined
\For{$s\in\{1,...,S\}$}
{
\For{$t $ \text{in stage} $s$}
{
    {   
        Randomly sample a subset $\mathcal{S}_t$ of clients
    
        Server sends $x_t$ to subset $\mathcal{S}_t$ of clients
    
        \For{each client $i \in \mathcal{S}_t$}
        {
        $\Delta_t^i=\text{LocalOPT}\left(i,\eta_l,K,x_t\right)$
        }
    
        Server aggregates $\Delta_t=\frac{1}{\lvert \mathcal{S}_t\rvert}\sum_{i\in \mathcal{S}_t}\Delta_t^i$
        
        \colorbox{babyblueeyes}{$d_{t+1}=(1-\beta_s)\Delta_{t}+\beta_s d_{t}$}

        \colorbox{babyblueeyes}{$h_{t+1}=(1-\nu_s)\Delta_{t}+\nu_s d_{t+1}$}
        
        \colorbox{babyblueeyes}{Update $x_{t+1}=x_t-\eta_s h_{t+1}$}
        
    }
}
}
return $x_T$
\caption{\colorbox{babyblueeyes}{Multistage FedGM}}
\label{multistage_FedGM_algorithm}
\end{algorithm2e}

We focus on a simple yet effective hyperparameter scheduler, ``constant and drop'' (a.k.a. ``step decay''). In its non-FL SGD version (a.k.a. multistage SGD), with a prespecified set of learning rates $\{\eta_s\}_{s=1}^{S}$ and training lengths $\{T_s\}_{s=1}^{S}$ (measured by number of iterations/epochs), the training process is divided into $S$ stages, and SGD with $\eta_s$ is applied for $T_s$ iterations/epochs at $s$-th stage, where $\{\eta_s\}_{s=1}^{S}$ is usually a non-increasing sequence \footnote{The name ``constant and drop'' refers to learning rate is dropped by some constant factor after each stage.}. We concentrate on ``constant and drop'' as it is the de-facto learning rate scheduler in most large-scale neural networks \citep{Krizhevsky12ImageNet,Sutskever13Init,He16Res,Huang2017DenseNet}, and has been theoretically shown to achieve near-optimal rate in non-FL settings \citep{ge19stepdecay,wang21stepdecay}.

The intuition behind ``constant and drop'' is straightforward: a large learning rate is held constant for a reasonably long period to take advantage of faster convergence until it saturates, and then the learning rate is dropped by a constant factor for more refined training.

We extend ``constant and drop'' to FedGM in Algorithm \ref{multistage_FedGM_algorithm}, which we refer to as Multistage FedGM. In Multistage FedGM (Algorithm \ref{multistage_FedGM_algorithm}), each stage has length $T_s$ ($T=\sum_{s=1}^{S}T_s$), and has its triplet of stagewise hyperparameters $\{\eta_s,\beta_s,\nu_s\}_{s=1}^{S}$. The convergence analysis in Sec \ref{subsec:convergence_analysis} also applies to single-stage FedGM by $S=1$.

To our best knowledge, there is no prior work giving definitive theoretical guarantee or empirical performances of any hyperparameter schedule in FL, especially considering multistage FedGM is an extremely flexible framework that allows both learning rate and momentum factor to evolve.

\subsection{Convergence Analysis of Multistage FedGM}
\label{subsec:convergence_analysis}

We now analyze the convergence of Algorithm \ref{multistage_FedGM_algorithm} under both full and partial participation settings. 

We aim to optimize objective \eqref{fed_min_objective}. Each local loss $f_i$ (and therefore $f$) may be general nonconvex function. We study the general \textit{non-i.i.d.} setting, i.e. $\mathcal{D}_i\neq\mathcal{D}_j$ when $i\neq j$. We state the assumptions that are needed in the analysis.

\begin{assumption}[Smoothness]
\label{smoothness_assumption}
Each local loss $f_i(x)$ is differentiable and has $L$-Lipschitz continuous gradients, i.e., $\forall x, x^\prime\in \mathbb{R}^d$, we have $\left\|\nabla f_i(x)-\nabla f_i(x^\prime)\right\| \leq L \left\| x-x^\prime\right\|$. And $f^\ast\triangleq\min_x f(x)$ exists, i.e., $f^\ast>-\infty$.
\end{assumption}

\begin{assumption}[Bounded Local Variance]
\label{bounded_local_assumption}
$\forall t, i$, LocalOPT can access an unbiased estimator $g_{t,k}^i=\nabla f_i(x_{t,k}^i, \xi_{t,k}^i)$ of true gradient $\nabla f_i(x_{t,k}^i)$, where $g_{t,k}^i$ is the stochastic gradient estimated with minibatch $\xi_{t,k}^i$. And each stochastic gradient on the $i$-th client has a bounded local variance, i.e., we have $\mathbb{E}\left[\left\| g_{t,k}^i - \nabla f_i(x_{t,k}^i) \right\|^2\right] \leq \sigma^2_l$.
\end{assumption}

\begin{assumption}[Bounded Global Variance]
\label{bounded_global_assumption}
The local loss $\{f_i(x)\}$ across all clients have bounded global variance, i.e., $\forall x$, we have $\frac{1}{n}\sum_{i=1}^{n}\left\| \nabla f_i(x)-\nabla f(x)\right\|^2\leq\sigma_g^2$.
\end{assumption}

Assumption \ref{smoothness_assumption}-\ref{bounded_global_assumption} are standard assumptions in nonconvex optimization and FL research, and have been universally adopted in most existing works \citep{reddi18adam_convergence,Li2020Fed-Non-IID,reddi2020adaptive,bao2020fast,yang2021achieving,wang22adaptive,wu2023federated,wu2023solving}. $\sigma_g^2=0$ in Assumption \ref{bounded_global_assumption} corresponds to the \textit{i.i.d.} setting. And we do not require the restrictive bounded gradient assumption \citep{reddi18adam_convergence,Avdiukhin21arbitrarycommunication,wu2023faster}.

Recall $T=\sum_{s=1}^{S}T_s$ is the number of rounds. Denote the expected gradient square as $\{\mathcal{G}_t\triangleq\mathbb{E}\left[\left\| \nabla f(x_t)\right\|^2\right]\}_{t\leq T}$. Define the average expected gradient square at $s$-th stage as $\Bar{\mathcal{G}}_s\triangleq\frac{1}{T_s}\Sigma_{t=T_1+\dots+T_{s-1}+1}^{T_1+\dots+T_s} \mathcal{G}_t$ and the average expected gradient square across $S$ stages as $\Bar{\mathcal{G}}\triangleq\frac{1}{S}\sum_{s=1}^S \Bar{\mathcal{G}}_s$. Bounding $\Bar{\mathcal{G}}$ generalizes from bounding $\frac{1}{T}\sum_{t=1}^{T}\mathbb{E}\left[\left\|\nabla f(x_t)\right\|^2\right]$ in single-stage to multistage setting. 

To reflect the common hyperparameter scheduling practice that is adopted by existing works e.g. \cite{Sutskever13Init,Smith18DontDecay,liu2020improved}, We request the stagewise hyperparameters fulfill the following constraints,
\begin{equation}
    \begin{gathered}
    \eta_{S}\leq\eta_{S-1}\leq\dots\leq\eta_{1} \quad \beta_1\leq\beta_2\leq\dots\leq\beta_{S}<1\\
    W_1 \equiv \frac{\eta_s\beta_s\nu_s}{1-\beta_s} \quad \text{and} \quad  W_2\equiv T_s\eta_s
    \end{gathered}
\label{stagewise_hyper_constraints}
\end{equation}
where $W_1$ and $W_2$ are two constants. Constraint \eqref{stagewise_hyper_constraints} essentially requires learning rate to be non-increasing and momentum factor to be non-decreasing at a similar rate, which is consistent with common practice, e.g. for SHB and NAG, \cite{Sutskever13Init,Smith18DontDecay,liu2020improved} propose a scheduler for $\beta$ to increase and close to 1 for faster convergence. And it is also natural for \eqref{stagewise_hyper_constraints} to require $T_s\eta_s$ as a constant. As the learning rate is decaying, more rounds in later stages are necessary for sufficient refined training. 

We now state the convergence guarantee of the multistage training regime in FL framework.

\subsubsection{Full Participation}
\label{subsec:full_participation}

If all clients are required to participate in each round, i.e. $\mathcal{S}_t=\{1,2,\dots,n\}$, we have,

\begin{theorem}
\label{multistage_fedgm_full_participation_convergence_theorem}
We optimize $f(x)$ with Algorithm \ref{multistage_FedGM_algorithm} (Full Participation) under Assumptions \ref{smoothness_assumption}-\ref{bounded_global_assumption}. Denote $\Bar{\eta} \triangleq \frac{1}{S}\sum_{s=1}^S \eta_s$ as the average server learning rate and $C_\eta\triangleq\frac{\eta_1}{\eta_S}$. Under the condition \footnote{The condition could be fulfilled by typical value assignment, and would recover the typical $\eta_l\leq\min\left\{\frac{1}{8KL},\frac{1}{KL\eta}\right\}$ constraint in FedAvg analysis \citep{yang2021achieving}, by setting $S=1$.} $\eta_l\leq\min\left\{\frac{1}{8KL},\frac{1}{K S C_\eta \left(L \Bar{\eta}  + 1 + L^2 W_1^2 C_\eta \right)}\right\}$, we would have:

\begin{gather*}
\Bar{\mathcal{G}}  \triangleq \frac{1}{S} \sum_{s=1}^{S} \frac{1}{T_s} \sum_{t=T_0+\dots+T_{s-1}}^{T_0+\dots+T_s-1} \mathbb{E}\left[\left\|\nabla f(x_t)\right\|^2\right] \\
 \leq \frac{64}{17}\frac{f(x_0)-f^\ast}{S W_2 \eta_l K} + \Psi_l \sigma_l^2 + \Psi_g  \sigma_g^2
\label{multistage_fedgm_full_participation_bound}
\end{gather*}
where $\Psi_l  \triangleq \frac{32}{17}\frac{L^2 W_1^2 T \Bar{\eta} \eta_l}{n W_2} +\frac{32}{17} \frac{L \Bar{\eta} \eta_l }{n} +\frac{32}{17}\frac{\eta_l}{n}+\frac{160}{17} \eta_l^2L^2K$, and $\Psi_g \triangleq \frac{960}{17} \eta_l^2L^2K^2$.

\end{theorem}

\begin{corollary}[Convergence Rate of Multistage FedAvg]
Suppose $\nu_s=0$, i.e., the FedAvg algorithm that allows learning rate vary across $S$ stages. By setting $\Bar{\eta}=\Theta\left(\sqrt{nK}\right)$ and $\eta_l=\Theta\left(\frac{1}{\sqrt{T}K}\right)$, $W_2=\Theta\left(\frac{T\sqrt{nK}}{S}\right)$, i.e. $T\Bar{\eta}$ equally divided into $S$ stages. $W_1=0$ as $\nu_s=0$. Suppose $T$ is sufficiently large, i.e. $T\ge nK$, we have a $\mathcal{O}\left(\frac{1}{\sqrt{TKn}}\right)$ convergence rate.
\label{corollary:multistage_fedavg_full_participation}
\end{corollary}

\begin{remark}[Why Multistage Helps?]
Corollary \ref{corollary:multistage_fedavg_full_participation} indicates multistage FedAvg recovers the best-known rate for general FL nonconvex optimization approaches, e.g. SCAFFOLD \citep{karimireddy2020scaffold} and FedAdam \citep{reddi2020adaptive}. Note single-stage FedAvg with two-sided learning rates also achieves the same rate \citep{yang2021achieving}. However, we do observe multistage FedAvg empirically converges much better than single-stage. We can obtain insights from Theorem \ref{multistage_fedgm_full_participation_convergence_theorem} why multistage helps. We note that $\Psi_l$ is only related to average learning rate $\Bar{\eta}$ (instead of initial learning rate $\eta_1$). At initial rounds, the first term with $f(x_0)-f^\ast$ dominates, and thus we could select a relatively large $\eta_1$ to ensure a more dramatic decay of this term. At later rounds, when $f(x_t)-f^\ast$ plateaus, we could enable smaller learning rate to control $\Bar{\eta}$. Thus, Theorem \ref{multistage_fedgm_full_participation_convergence_theorem} indicates a less stringent reliance on $\eta_1$, which enables us to flexibly select suitable $\eta$ depending on which training stage we are in, that can still guarantee convergence.
\label{remark:full_participation_why_multistage_helps}
\end{remark}

\begin{corollary}[Convergence Rate of Multistage FedGM]
Suppose $S>1$, i.e. the multistage regime, by setting $\Bar{\eta} = \Theta\left(\sqrt{nK}\right)$, $\eta_l=\Theta\left(\frac{1}{\sqrt{T}K}\right)$, $W_2=\Theta\left(\frac{T\sqrt{nK}}{S}\right)$. Let $W_1^2=\mathcal{O}\left(\frac{\sqrt{nK}}{S}\right)$ \footnote{It holds by setting an infinitesimal $\beta$ or $\nu$ at early stages when $\eta$ is large, but $\beta$ or $\nu$ can go to 1 when $\eta$ is reduced to $o\left( \frac{\sqrt[\leftroot{-3}\uproot{3}4]{nK}}{\sqrt{S}} \right)$.}. When $T\ge Kn$, we have a $\mathcal{O}\left(\frac{1}{\sqrt{TKn}}\right)$ convergence rate. 
\label{corollary:multistage_fedgm_full_participation}
\end{corollary}

\begin{remark}[Why Momentum Helps?]
We attribute the empirically superior performances of momentum to two reasons. (a) When clients are dynamically heterogeneous, historical gradient information has regularization effect to avoid the search direction from going wild. (b) Server learning rate $\eta$ acts like a multiplier to client learning rate $\eta_l$ in FedAvg, i.e. $\eta>1$ effectively enhances the reliance on current round gradient. Due to the same reason as in (a), such reliance can harm convergence. In contrast, in FedGM, $\beta$ and $\nu$ act as a buffer that could to some extent absorb the impact from a large $\eta$. We empirically observe in Appendix \ref{subsec:appendix_more_results_fedgm}, with same $\eta_l$, FedGM could sustain a much larger $\eta$, while FedAvg diverges very easily with a moderately large $\eta$.
\label{remark:full_participation_why_momentum_helps}
\end{remark}

\subsubsection{Partial Participation}
\label{subsec:partial_participation}

Full participation rarely holds in reality, thus we further analyze multistage FedGM in partial participation setting \footnote{In each round $t$, the server samples a subset of clients $\mathcal{S}_t$ (suppose $\left | \mathcal{S}_t \right | = m < n$) uniformly at random without replacement, i.e. $\mathbb{P}\left\{i\in\mathcal{S}_t\right\}=\frac{m}{n}$ and $\mathbb{P}\left\{i,j\in\mathcal{S}_t\right\}=\frac{m\left(m-1\right)}{n\left(n-1\right)}$.}.

\begin{theorem}
\label{multistage_fedgm_partial_participation_convergence_theorem}
We optimize $f(x)$ with Algorithm \ref{multistage_FedGM_algorithm} (Partial Participation) under Assumptions \ref{smoothness_assumption}-\ref{bounded_global_assumption}. Denote $\Bar{\eta}$ and $C_\eta$ as in Theorem \ref{multistage_fedgm_full_participation_convergence_theorem}. Under the condition $\eta_l\leq\frac{1}{8KL}$, and $\eta_l \left( C_\eta + L \Bar{\eta} C_\eta + L^2 W_1^2 C_\eta \right) SK \leq \min\left\{ \frac{m(n-1)}{n(m-1)}, \frac{17m}{282} \right\}$, we would have:

\begin{gather*}
\Bar{\mathcal{G}} \leq \frac{64}{17}\frac{f(z_0)-f^\ast}{S W_2 \eta_l K}  
+ \Psi_l \sigma_l^2
+ \Psi_g \sigma_g^2
\label{multistage_fedgm_partial_participation_bound}
\end{gather*}

where $\Psi_l  \triangleq \frac{\eta_l}{m}\Phi +\frac{15\left(n-m\right)K^2L^3\eta_l^3}{m\left(n-1\right)}\Phi + \frac{160}{17} \eta_l^2L^2K$, $\Psi_g \triangleq  \frac{90\left(n-m\right)K^3L^3\eta_l^3}{m\left(n-1\right)}\Phi  +  \frac{3\eta_l\left(n-m\right)K}{m\left(n-1\right)}\Phi     + \frac{960}{17} \eta_l^2L^2K^2$, and $\Phi \triangleq \frac{32 T \Bar{\eta} + 32 L T \hat{\eta}^2 + 32 L^2 W_1^2 T \Bar{\eta}}{17 W_2}$.

\end{theorem}

\begin{corollary}[Convergence Rate of Multistage FedGM]
Suppose $S>1$, i.e. the multistage regime, by setting $\Bar{\eta} = \Theta\left(\sqrt{mK}\right)$, $\hat{\eta}^2 = \Theta\left(mK\right)$, $\eta_l=\Theta\left(\frac{1}{\sqrt{T}K}\right)$, $W_2=\Theta\left(\frac{T\sqrt{mK}}{S}\right)$ and $W_1^2=\mathcal{O}\left(\sqrt{mK}\right)$, we have convergence rate as $\mathcal{O}\left(\sqrt{\frac{K}{Tm}}\right)$.
\label{corollary:fedgm_partial_participation_convergence_rate}
\end{corollary}

\begin{remark}
$\mathcal{O}\left(\sqrt{\frac{K}{Tm}}\right)$ recovers the best known convergence rate for FL nonconvex optimization \citep{yang2021achieving}. Similar to Remark \ref{remark:full_participation_why_multistage_helps}, Theorem \ref{multistage_fedgm_partial_participation_convergence_theorem} shows an reliance on average learning rate, which explains why multistage scheme helps empirically. $\mathcal{O}\left(\sqrt{\frac{K}{Tm}}\right)$ indicates a slowdown effect from more local computation, which is supported by some existing works \citep{Li2020Fed-Non-IID}, while others observe a different effect of $K$ \citep{lin2020dont}. The exact impact of $K$ on convergence warrants further investigation.
\end{remark}

%% file: subfiles/5-free.tex
\section{Momentum with System Heterogeneity}
\label{sec:autonomous}

\subsection{Autonomous Multistage FedGM}

For a simplified abstraction of real world settings, most FL algorithms make the assumption that, all clients synchronize with the same global model and they conduct identical number of local updates at any given round. Though the assumption has been adopted in most existing works \citep{McMahan2017FedAvg,Hsu2019MeasuringTE,Li20FedProx,karimireddy2020scaffold,reddi2020adaptive,bao2022doubly, wang22adaptive}, it rarely holds in reality.

In light of the limitations of existing works, we propose a general framework called Autonomous Multistage FedGM that enables the following three features, i.e. \textbf{heterogeneous local computing}, \textbf{asynchronous aggregation}, and \textbf{flexible client participation}, which is formalized in Algorithm \ref{alg:autonomous_fedgm}. 

Autonomous Multistage FedGM could effectively mitigate straggler effect and poor convergence issue in highly heterogeneous cross-device deployments. We leave a more detailed discussion of Algorithm \ref{alg:autonomous_fedgm} to Appendix \ref{sec:autonomous_multistage} due to space limit.

Specifically, in Autonomous Multistage FedGM, the client decides when to participate in the training, and idling between rounds or even completely unavailable are both allowed. Once it decides to participate at round $t$, it retrieves current global model $x_\mu$ from the server and conduct $K_{t,i}$ local steps to update to $x^i_{\mu,K_{t,i}}$. Note in vanilla FedAvg, $K_{t,i}=K$ for any $i$ and $t$. In contrast, we allow $K_{t,i}$ to be time-varying and device-dependent. The client then normalizes the model update by $K_{t,i}$ to avoid model biased towards clients with more local updates. Concurrently, the server collects the model updates from the clients. As every client may participate in training at a different round, the collected model update $\Delta_{t-\tau_{t,i}}^i$ may be from a historic timestamp, i.e. $\tau_{t,i}$ away from current time $t$. The server triggers global update whenever it collects $m$ model updates and we denote the set of $m$ responsive clients as $\mathcal{S}_t$. The global update is same as multistage FedGM (i.e. Lines 11-13). Note that server optimization is concurrent with clients, i.e., the global update happens whenever $m$ model updates are collected, regardless of whether there are still some clients conducting local computation, thus ensuring there is no straggler.

Autonomous multistage FedGM, i.e. Algorithm \ref{alg:autonomous_fedgm}, will recover multistage FedGM, i.e. Algorithm \ref{multistage_FedGM_algorithm}, if we set $K_{t,i}=K$ and $\tau_{t,i}=0$ for $\forall t, i$. Please note that varying $K_{t,i}$ and nonzero $\tau_{t,i}$ bring nontrivial extra complexity to the theoretical analysis as can be seen in our proof.

\begin{algorithm2e}[tb]
\SetAlgoVlined
\KwIn{Same as Algorithm \ref{multistage_FedGM_algorithm}}
\SetAlgoLined
\For{$s\in\{1,...,S\}$}
{
\For{$t $ \text{in stage} $s$}
{   
    \colorbox{babyblueeyes}{\textbf{At Each Client (Concurrently)}}
    
    Once decided to participate in the training, retrieve $x_\mu$ from the server and its timestamp, set $x_{\mu,0}^i=x_\mu$.

    Select a number of local steps $K_{t,i}$, which is time-varying and device-dependent.

    $\Delta_\mu^i=\textbf{LocalOPT}\left(i,\eta_l,K_{t,i},x_\mu\right)$

    Normalize and send $\Delta_\mu^i=\frac{\Delta_\mu^i}{K_{t,i}}$

    \colorbox{babyblueeyes}{\textbf{At Server (Concurrently)}}
    
    Collect $m$ local updates $\{\Delta_{t-\tau_{t,i}}^i, i\in\mathcal{S}_t\}$ returned from the clients to form set $\mathcal{S}_t$, where $\tau_{t,i}$ is the random delay of the client $i$'s local update, $i\in\mathcal{S}_t$

    Aggregate $\Delta_t=\frac{1}{\lvert\mathcal{S}_t\rvert}\sum_{i\in \mathcal{S}_t}\Delta_{t-\tau_{t,i}}^i$

    $d_{t+1}=(1-\beta_s)\Delta_{t}+\beta_s d_{t}$

    $h_{t+1}=(1-\nu_s)\Delta_{t}+\nu_s d_{t+1}$
        
    Update $x_{t+1}=x_t-\eta_s h_{t+1}$

}
}
return $x_T$
\caption{\colorbox{babyblueeyes}{Autonomous Multistage FedGM}}
\label{alg:autonomous_fedgm}
\end{algorithm2e}

\begin{figure}[h]

\centering
\subfigure{

\includegraphics[width=.21\textwidth]{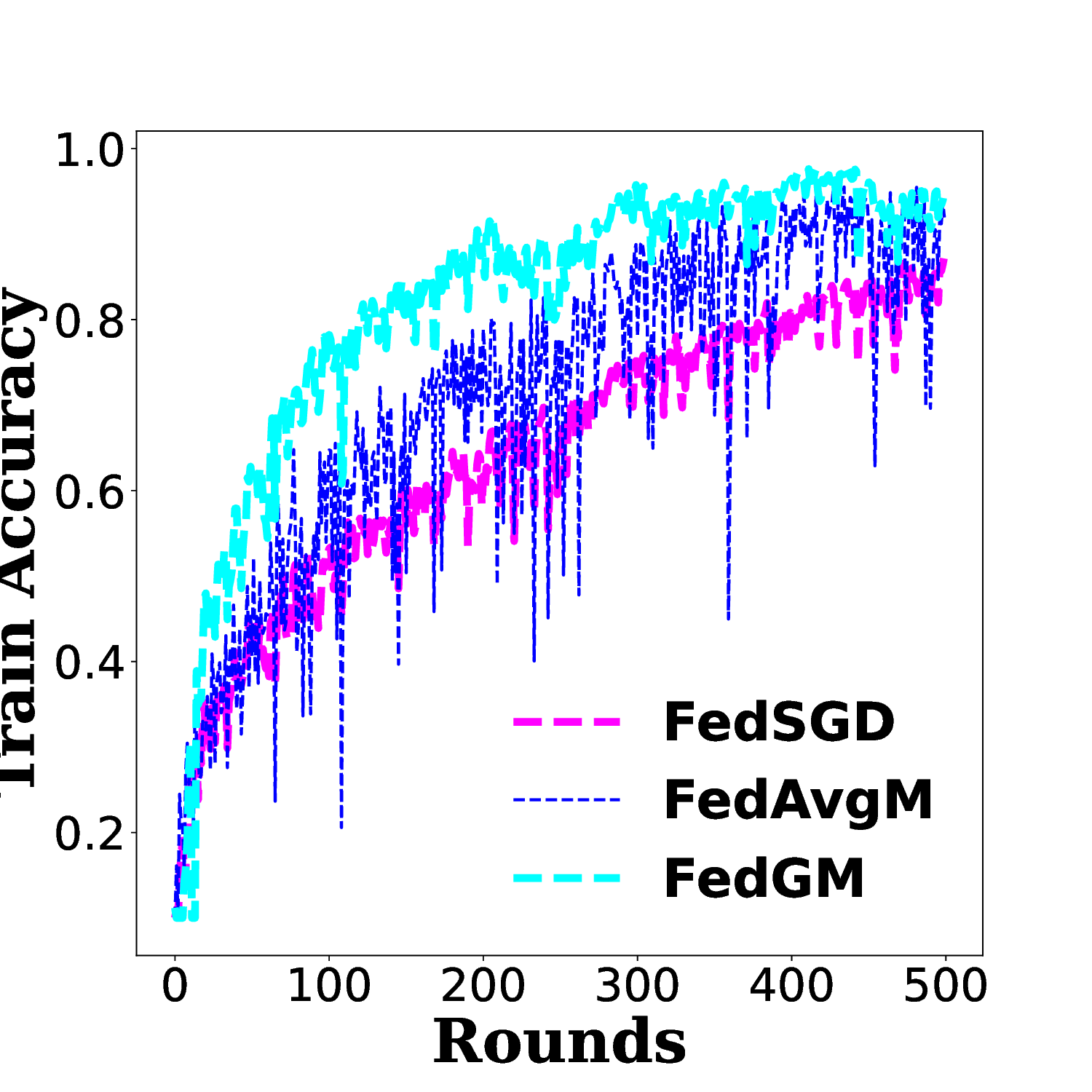}
\label{subfig:resnet_cifar10_train}
}
\hspace{-2pt}
\subfigure{
\includegraphics[width=.21\textwidth]{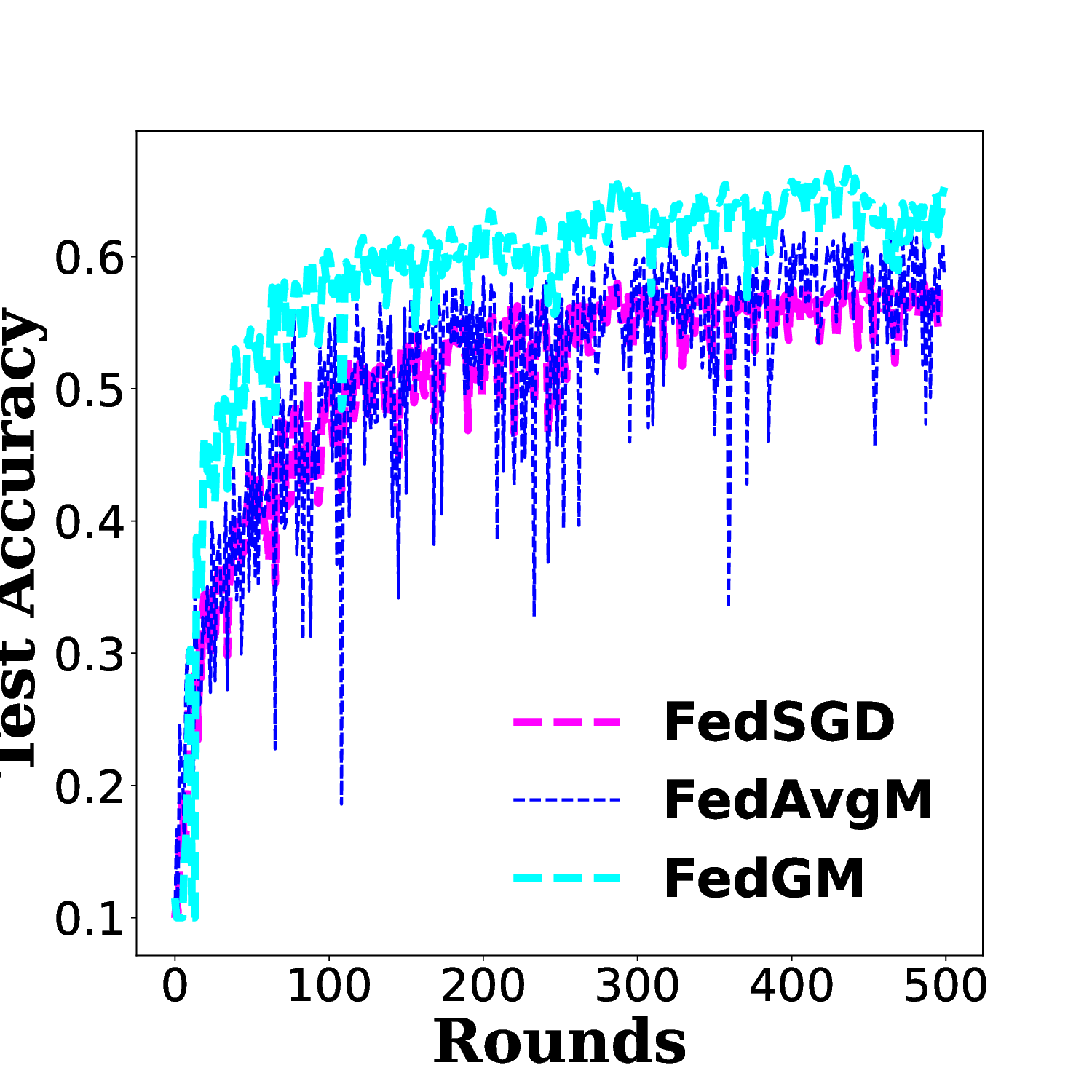}
\label{subfig:resnet_cifar10_test}
}
\hspace{-2pt}
\subfigure{
\hspace{0pt}
\includegraphics[width=.21\textwidth]{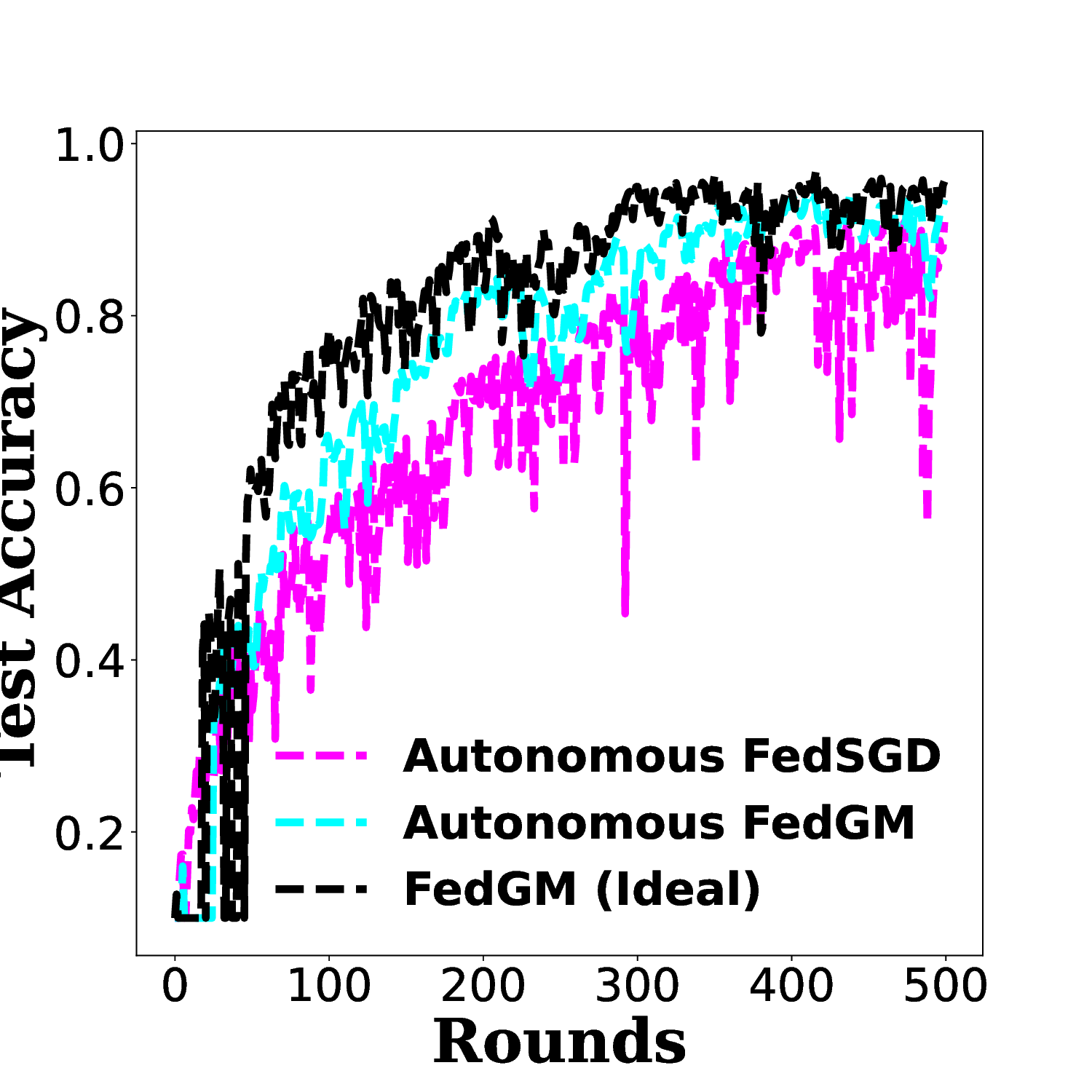}
\label{subfig:autonomous_resnet_cifar10_train}
}
\hspace{-2pt}
\subfigure{
\includegraphics[width=.21\textwidth]{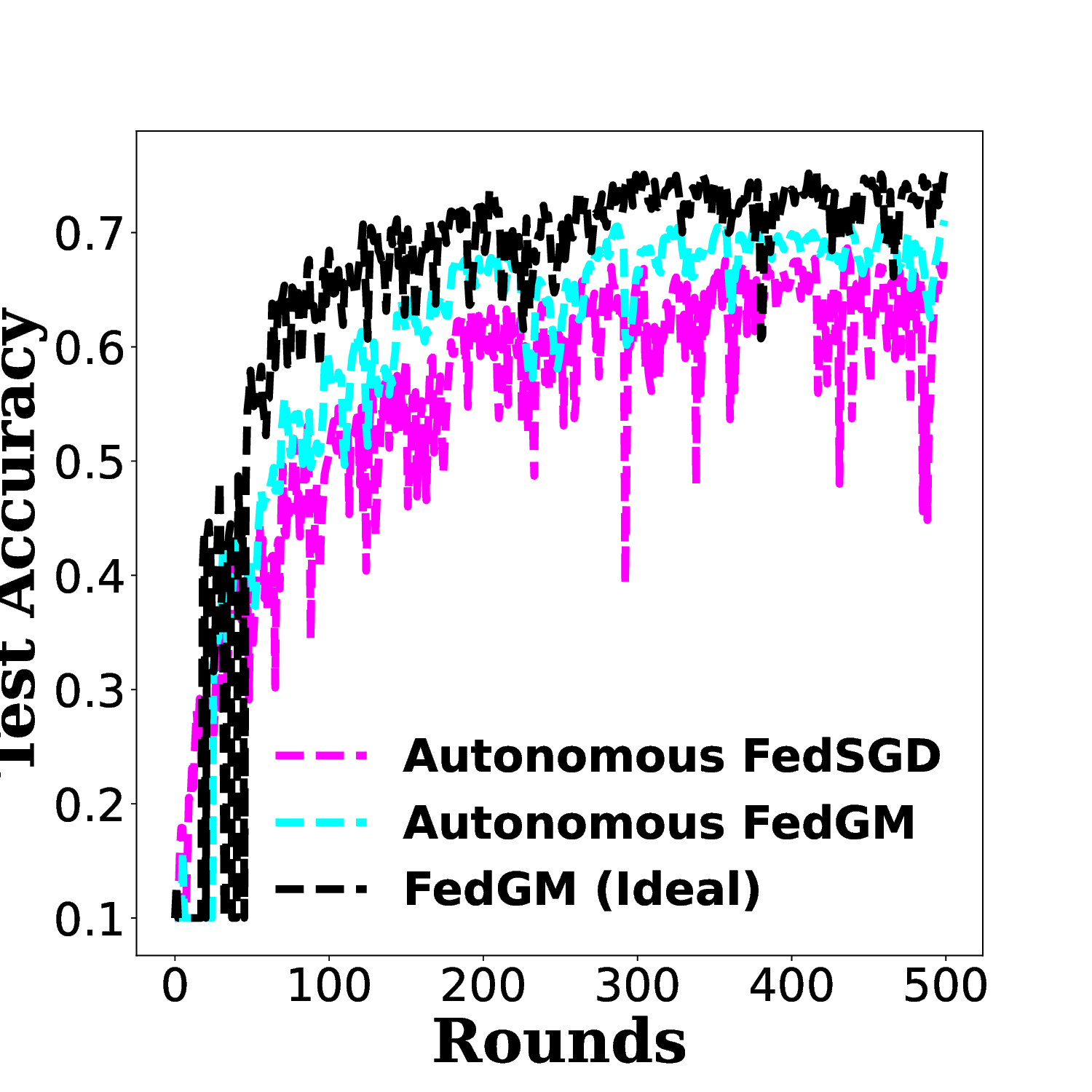}
\label{subfig:autonomous_resnet_cifar10_test}
}

\caption{\ref{subfig:resnet_cifar10_train} Training and \ref{subfig:resnet_cifar10_test} Testing Curves for FedGM (ResNet on CIFAR-10). FedGM outperforms FedAvg/FedAvgM. \ref{subfig:autonomous_resnet_cifar10_train} Training and \ref{subfig:autonomous_resnet_cifar10_test} Testing for Autonomous FedGM (ResNet on CIFAR-10).}
\label{fig:resnet_cifar10}

\end{figure}

\subsection{Convergence Analysis}

We state the convergence guarantee of autonomous multistage FedGM as follows,

\begin{theorem}
\label{multistage_fedgm_free_uniform_arrival_convergence_theorem}
We optimize $f(x)$ with Algorithm \ref{alg:autonomous_fedgm} under assumptions \ref{smoothness_assumption}-\ref{bounded_global_assumption}. Suppose the maximum delay is bounded, i.e. $\tau_{t,i}\leq\tau<\infty$ for any $i\in\mathcal{S}_t$ and $t\in\{0,1,\dots,T-1\}$. Under the condition $\eta_l\leq\min\left\{\frac{1}{8K_{t,\text{max}}L},\sqrt{\frac{1}{ 120L^2 C_\eta \tau K_{t,\text{max}}^2}} \right\}$, where $K_{t,\text{max}} =\max_{i\in\mathcal{S}_t}K_{t,i} $. And further assume each client is included in $\mathcal{S}_t$ with probability $\frac{m}{n}$ uniformly and independently. With necessary abbreviation for ease of notation \footnote{We denote $\Bar{\eta}\triangleq\frac{1}{S}\sum_{s=0}^{S-1}\eta_s$ (average server learning rate), $\hat{\eta}^2\triangleq\frac{1}{S}\sum_{s=0}^{S-1}\eta^2_s$, $\hat{\eta}^3\triangleq\frac{1}{S}\sum_{s=0}^{S-1}\eta^3_s$, $\frac{1}{K_t}=\frac{1}{m}\sum_{i\in\mathcal{S}_t}\frac{1}{K_{t,i}}$, $\bar{K}_t\triangleq \frac{1}{m}\sum_{i\in\mathcal{S}_t}K_{t,i}$, $\hat{K}_t^2 \triangleq \frac{1}{m}\sum_{i\in\mathcal{S}_t}K^2_{t,i}$, $\phi_1 \triangleq  \frac{1}{T}\sum_{t=0}^{T-1} \Bar{K}_t$, $\phi_2 \triangleq  \frac{1}{T}\sum_{t=0}^{T-1} \hat{K}_t^2$, and $\phi_3 \triangleq \frac{1}{T}\sum_{t=0}^{T-1} \frac{1}{K_t}$, for ease of notation.}, we would have:

\begin{gather*}
 \Bar{\mathcal{G}}
\leq \frac{4 \left(f(x_0) - f^\ast  \right)}{S W_2 \eta_l} 
+  \Phi_l \sigma_l^2
+ \Phi_g \sigma_g^2 
\label{multistage_fedgm_free_uniform_arrival_bound}
\end{gather*}
$\Phi_l \triangleq \frac{20  \eta^2_l L^2 T \Bar{\eta}}{W_2} \phi_1 + \frac{4 L^2\tau^2\hat{\eta}^3\eta_l^2T}{mW_2}\phi_3 + \frac{2 L^2W_1^2 \Bar{\eta} \eta_l T}{m W_2}\phi_3 + \frac{2 \Bar{\eta} \eta_l T}{m W_2}\phi_3 + \frac{2 L\hat{\eta}^2\eta_l}{m W_2}\phi_3$, and $\Phi_g \triangleq \frac{120 \eta^2_l L^2 T \Bar{\eta} \phi_2}{W_2}$. 
\end{theorem}

\begin{corollary}[Convergence Rate]
Suppose an identical $K$ for all $t$ and $i$. By appropriately setting $\Bar{\eta}$, $\eta_l$, $W_1$, $W_2$, we have the convergence rate as, $\mathcal{O}\left(\frac{1}{\sqrt{mKT}}\right)+ \mathcal{O}\left(\frac{\tau^2}{T}\right) +  \mathcal{O}\left( \frac{K^2}{T} \right)$.
\label{corollary:free_multistage_fedgm_rate}
\end{corollary}

\begin{remark}
Corollary \ref{corollary:free_multistage_fedgm_rate} indicates $\tau$ brings a slowdown in convergence. Fortunately, with a sufficiently large $T$ (e.g. $T\ge mK^5$) and a manageable $\tau$ (e.g. $\tau \leq \frac{T^\frac{1}{4}}{(mK)^\frac{1}{4}}$), autonomous multistage FedGM obtains a $\mathcal{O}\left(\frac{1}{\sqrt{mKT}}\right)$ rate. Note that we make an additional assumption that each client is included in $\mathcal{S}_t$ with probability $\frac{m}{n}$ uniformly and independently, which is necessary as the following Corollary \ref{corollary:free_multistage_fedgm_general_arrival_rate} indicates if without such assumption, the rate has a non-convergent $\mathcal{O}\left( \sigma_g^2 \right)$ term that we cannot avoid (the lower bound is $\Omega\left( \sigma_g^2 \right)$).
\end{remark}

\begin{corollary}[Convergence Rate w/o Uniform Sampling Assumption]
Suppose an identical $K$ for all $t$ and $i$. By appropriately setting $\Bar{\eta}$, $\eta_l$, $W_1$, $W_2$, we have the convergence rate as, $\mathcal{O}\left(\frac{1}{\sqrt{mKT}}\right)+ \mathcal{O}\left(\frac{\tau^2}{T}\right) +  \mathcal{O}\left( \frac{K^2}{T} \right) + \mathcal{O}\left( \sigma_g^2 \right)$, and the non-vanishing $\mathcal{O}\left( \sigma_g^2 \right)$ is unavoidable. \footnote{We informally state Corollary \ref{corollary:free_multistage_fedgm_general_arrival_rate} due to page limit, please refer to Appendix \ref{sec:proof_free_multistage_fedgm_general_arrival} for a formal statement.}
\label{corollary:free_multistage_fedgm_general_arrival_rate}
\end{corollary}

%% file: subfiles/6-experiments.tex
\section{Experimental Results}
\label{sec:experiments}

In this section, we present empirical evidence to verify our theoretical findings. We train ResNet \citep{He2016DeepResNet} and VGG \citep{Simonyan14VGG} on CIFAR10 \citep{Krizhevsky2009CIFAR}. To simulate data heterogeneity in CIFAR-10, we impose label imbalance across clients, i.e. each client is allocated a proportion of the samples of each label according to a Dirichlet distribution \citep{Hsu2019MeasuringTE, Yurochkin2019BayesianNF}. The concentration parameter $\alpha>0$ indicates the level of \textit{non-i.i.d.}, with smaller $\alpha$ implies higher heterogeneity, and $\alpha\to\infty$ implies \textit{i.i.d.} setting. Unless specified otherwise, we have 100 clients in all experiments, and the partial participation ratio is 0.05, i.e., 5 out of 100 clients are picked in each round, \textit{non-i.i.d.} is $\alpha=0.5$, and local epoch is 3. We defer many more results and details of hyperparameter settings to Appendix \ref{sec:appendix_exp}.

\subsection{Results on FedGM}
\label{subsec:exp_fedgm}

Figure \ref{fig:resnet_cifar10} shows the results for ResNet on CIFAR-10 with FedGM, FedAvgM, and FedAvg. We perform grid search over $\eta\in\{0.5,1.0,1.5,\dots,5.0\}$, $\beta\in\{0.7,0.9,0.95\}$, and $\nu\in\{0.7,0.9,0.95\}$. We report their respective best results in Figure \ref{fig:resnet_cifar10}. We observe that though FedAvgM converges faster than FedAvg, it is only marginally better in terms of testing. FedGM, in contrast, outperforms FedAvgM and FedAvg in both measures. Therefore, a general momentum, instead of only SHB, is critical empirically. We analyze possible reasons and leave more results with VGG and different heterogeneity levels $\alpha$ to Appendix \ref{subsec:appendix_more_results_fedgm}.

\subsection{Results on Multistage FedGM}
\label{subsec:exp_multistage_fedgm}

\begin{figure}[h]

\centering
\subfigure{
\hspace{0pt}
\includegraphics[width=.35\textwidth]{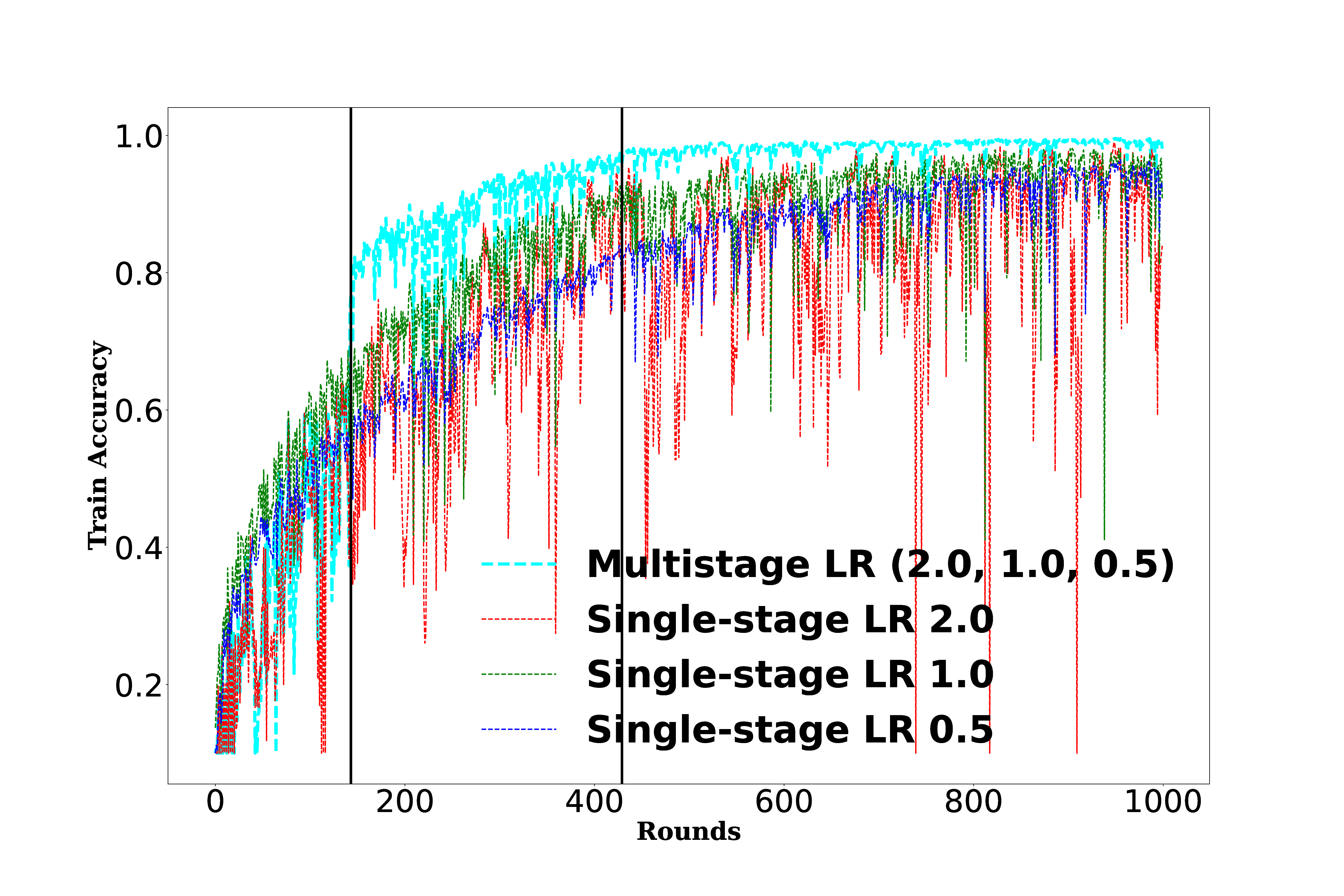}
\label{subfig:multistage_resnet_cifar10_train}
}
\subfigure{
\hspace{0pt}
\includegraphics[width=.35\textwidth]{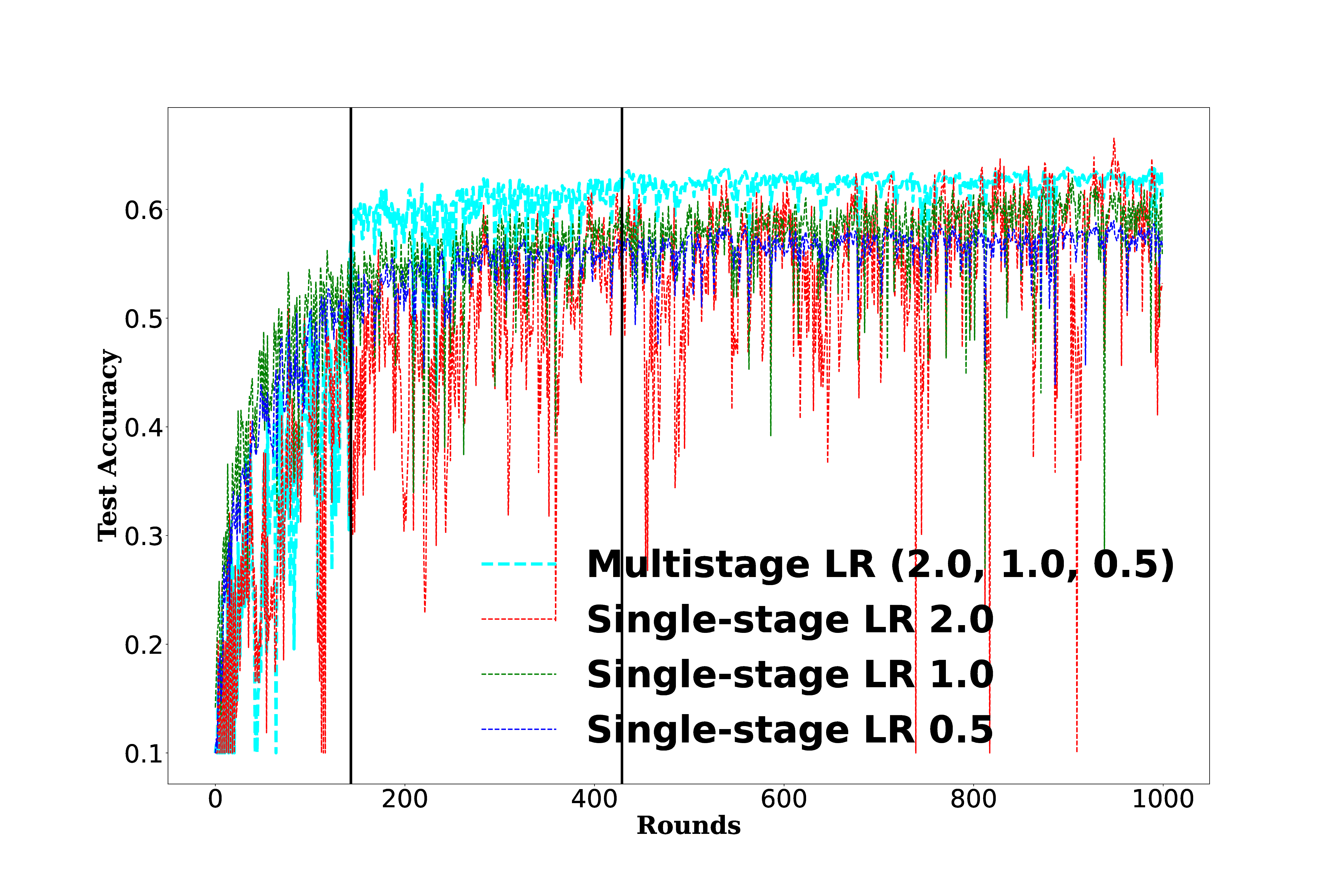}
\label{subfig:multistage_resnet_cifar10_test}
}

\caption{\ref{subfig:multistage_resnet_cifar10_train} Training and \ref{subfig:multistage_resnet_cifar10_test} Testing Curves for Multistage FedGM vs. Single-stage FedGM.}
\label{fig:resnet_cifar10_multistage}
\end{figure}

Figure \ref{fig:resnet_cifar10_multistage} shows the results for ResNet on CIFAR-10 with multistage vs. single-stage FedGM. The two black vertical lines at round 143 and 429 mark the end of 1st/2nd stage. For multistage FedGM, $(\eta_1=2.0,\eta_2=1.0,\eta_3=0.5)$, the $\beta$ also changes according to Eq. \ref{stagewise_hyper_constraints}. From Figure \ref{fig:resnet_cifar10_multistage}, we observe multistage FedGM is better than single-stage FedGM, no matter what constant $\eta$ it takes. Specifically, at first stage, $\eta_1=2.0$ makes the training curve fluctuate dramatically, but later into 2nd/3rd stage, the training stabilizes with smaller $\eta_2$ and $\eta_3$. Multistage FedGM achieves a balance between early exploration and late exploitation. Multistage is also superior to its counterpart in testing. We leave more experiments to Appendix \ref{subsec:more_exp_multistage_appendix}.

\subsection{Results on Autonomous FedGM}
\label{subsec:exp_autonomous_fedgm}

Figure \ref{fig:resnet_cifar10} shows the results for ResNet on CIFAR-10 with Autonomous FedGM (\& FedAvg). Please refer to Appendix \ref{subsec:exp_settings_appendix} for detailed settings. We perform a grid search as in Section \ref{subsec:exp_fedgm}. We report their respective best curves. We plot an ideal FedGM (i.e. synchronous and identical local epochs) as reference line. We could observe Autonomous FedGM outperforms Autonomous FedAvg with system heterogeneity. Though Autonomous FedGM suffers a slowdown compared to the ideal FedGM, it is within a small margin, which supports our theory in Corollary \ref{corollary:free_multistage_fedgm_rate} and validates the effectiveness of Autonomous FedGM. We leave more experiments to Appendix \ref{subsec:appendix_more_exp_autonomous}.

%% file: subfiles/7-conclusion.tex
\section{Conclusion}
\label{sec:conclusion}

This paper systematically studied how the server momentum could help alleviate client drift that arises from both data heterogeneity and system heterogeneity. We demonstrated the critical role of momentum schemes and proper hyperparameter schedule by providing a rigorous convergence analysis and extensive empirical evidence, which pave a way for more widely and disciplined use of server momentum in the federated learning research community. 

%% file: subfiles/8-appendix.tex
\centerline{\huge\textbf{Appendix}}

\section{Organization of Appendix}

Appendix is organized as follows. In Section \ref{sec:autonomous_multistage}, we discuss the omitted details of server momentum with system heterogeneity from Section \ref{sec:autonomous}. In Section \ref{sec:related_work}, we provide the discussion of related works. In Section \ref{sec:proof_multistage_fedgm_full_participation}, we show the proof of Theorem \ref{multistage_fedgm_full_participation_convergence_theorem}, Corollary \ref{corollary:multistage_fedavg_full_participation}, and Corollary \ref{corollary:multistage_fedgm_full_participation}. In Section \ref{sec:proof_multistage_fedgm_partial_participation}, We provide the proof of Theorem \ref{multistage_fedgm_partial_participation_convergence_theorem} and Corollary \ref{corollary:fedgm_partial_participation_convergence_rate}. In Section \ref{sec:proof_free_multistage_fedgm_uniform_arrival}, we provide the proof of Theorem \ref{multistage_fedgm_free_uniform_arrival_convergence_theorem} and Corollary \ref{corollary:free_multistage_fedgm_rate}. In Section \ref{sec:proof_free_multistage_fedgm_general_arrival}, we provide the proof of Corollary \ref{corollary:free_multistage_fedgm_general_arrival_rate}. Note that we use a 0-indexing for $T$ and $S$ in most proofs, i.e. the rounds (stages) are denoted as $\{0,\dots,T-1\}$ ($\{0,\dots,S-1\}$), which is equivalent to the 1-indexing in main text, i.e. $\{1,\dots,T\}$ ($\{1,\dots,S\}$). Finally, in Section \ref{sec:appendix_exp}, we provide experimental settings and extra experimental results that are omitted from main text.

\section{Autonomous Multistage FedGM}
\label{sec:autonomous_multistage}

In this section, we discuss the omitted details of server momentum with system heterogeneity from Section \ref{sec:autonomous}.

\subsection{Ubiquitous System Heterogeneity}

For a simplified abstraction of real world settings, most FL algorithms make the assumption that, all clients initialize with the same global model and they conduct identical number of local updates at any given round.

More formally, we could observe from LocalOPT (Algorithm \ref{alg:localopt}) that the following assumptions have been made, (a) \textit{Homogeneous Local Updates} all participating clients would do local gradient descent for $K$ steps; (b) \textit{Uniform Client Participation} each client would participate in a given communication round uniformly according to a given distribution that is independent across rounds; (c) \textit{Synchronous Local Clients} all participating clients always initialize at $x_t$, i.e., the global model at current timestamp.

Though these three assumptions have been adopted in most existing works \citep{McMahan2017FedAvg,Hsu2019MeasuringTE,Li20FedProx,karimireddy2020scaffold,reddi2020adaptive,wang22adaptive, hu2023beyond}, each of these assumptions rarely holds in reality. Due to unavoidable \textbf{heterogeneous client capability}, and \textbf{unpredictable availability}, enforcing identical local epochs and synchrony would incur straggler effect and unnecessary energy waste \citep{Kairouz21AdvancesProblems}. Therefore, realistic FL system is more economical to allow different local epochs and \textbf{asynchronous aggregation}.

When studying client heterogeneity and the resulting client drift, most works focus explicitly on data heterogeneity \citep{Li2020Fed-Non-IID,yang2021achieving}, while ignoring the equally ubiquitous system heterogeneity, which casts doubt on the applicability of the corresponding algorithms in practice.

\subsection{Autonomous Multistage FedGM}

In light of the limitations of existing works, we aim to propose a general framework that enables all three features, i.e. \textbf{heterogeneous local computing}, \textbf{asynchronous aggregation}, and \textbf{flexible client participation}, which is formalized in Algorithm \ref{alg:autonomous_fedgm}.

Specifically, in Autonomous Multistage FedGM, the client decides when to participate in the training, and idling between rounds or even completely unavailable are both allowed. Once it decides to participate at round $t$, it retrieves current global model $x_\mu$ from the server and initializes $x_{\mu,0}^i=x_\mu$ locally, and conduct $K_{t,i}$ local steps to update from $x_{\mu,0}^i$ to $x^i_{\mu,K_{t,i}}$. Note in vanilla FedAvg, $K_{t,i}=K$ for any $i$ and $t$. In contrast, we allow $K_{t,i}$ to be time-varying and device-dependent. The client then normalizes the model update by $K_{t,i}$, i.e. $\Delta_\mu^i=\frac{x_{\mu,0}^i-x_{\mu,K_{t,i}}^i}{K_{t,i}}$, to avoid model biased towards clients with more local updates. Concurrently, the server collects the model updates from the clients. As every client may participate in training at a different round, the collected model update $\Delta_{t-\tau_{t,i}}^i$ may be from a historic timestamp, i.e. $\tau_{t,i}$ away from current time $t$. If we set the random delay $\tau_{t,i}=0$, it would be ordinary synchronous aggregation. The server triggers global update whenever it collects $m$ model updates and we denote the set of $m$ responsive clients as $\mathcal{S}_t$. The global update is same as multistage FedGM (i.e. Lines 11-13). Note that server optimization is concurrent with clients, i.e., the global update happens whenever $m$ model updates are collected, regardless of whether there are still some clients conducting local computation, thus ensuring there is no straggler.

Autonomous multistage FedGM will recover multistage FedGM, i.e. Algorithm \ref{multistage_FedGM_algorithm}, if we set $K_{t,i}=K$ and $\tau_{t,i}=0$ for $\forall t, i$. Please note that varying $K_{t,i}$ and nonzero $\tau_{t,i}$ bring nontrivial extra complexity to the theoretical analysis as can be seen in our proof.

\section{Related Work}
\label{sec:related_work}

\subsection{Tackling Client Heterogeneity and Client Drift in FedAvg}
\label{subsec:fedavg_related_work}

Deep learning models have been widely applied in many different domains, e.g. \cite{Mnih2013PlayingAW,He2016DeepResNet,Devlin2019BERT,Jure2017GNN,google16deep&wide,SuoICHI19,Xun2020CorrelationNF,Wu2023DiPmarkAS}, mostly in centralized environment. Due to privacy concerns and regulatory requirements \cite{european_commission_regulation_2016,California_Consumer_Privacy_Act_CCPA}, Federated Averaging (FedAvg) \cite{McMahan2017FedAvg} has been applied to avoid data transmission in collaborative training of deep learning models in a wide range of settings \cite{Li20FedProx,rothchild20fetchsgd,Wang20FedNova,fallah2020personalized,li2021model,bao2023adaptive,wu2023federated,wu2023solving}.

Client heterogeneity and its resulting client drift is known to destabilize FedAvg convergence \citep{zhao2018federated-noniid,karimireddy2020scaffold}. FedProx \citep{Li20FedProx} proposes to regularize the difference against global model in local objective. SCAFFOLD \citep{karimireddy2020scaffold} leverages variance reduction technique to reduce client drift and achieves the best known $\mathcal{O}\left(\frac{1}{\sqrt{nKT}}\right)$ rate in full participation setting. However, SCAFFOLD is not stateless which restricts its application in cross-device FL. \cite{reddi2020adaptive,wang22adaptive} propose a family of federated adaptive optimizers e.g. FedADAM, FedADAGRAD, and FedAMS that are natural extensions from non-FL adaptive optimizers to FL settings. Recent works propose algorithms to alleviate client heterogeneity in FL bilevel optimization problems, e.g. minimax \cite{wu2023solving} or conditional stochastic optimization \cite{wu2023federated}. The most relevant line of research to this paper is on server-side momentum. Server-side momentum is first empirically studied in \citep{Hsu2019MeasuringTE}, where FedAvgM is observed to outperform FedAvg in \textit{non-i.i.d.} settings by a significant margin. Recent explorations include \citep{rothchild20fetchsgd} that studies the interplay between server momentum and compression, and \citep{khanduri2021stem} which studies a two-sided momentum scheme that allows both server momentum and client momentum. However, all existing works have the following limitations that our paper aims to address, (a) they do not provide a unified analysis for a family of momentum schemes; (b) they do not incorporate any realistic hyperparameter schedulers; (c) they ignore an important source of client heterogeneity, i.e. system heterogeneity.

\subsection{Hyperparameter Scheduling}
\label{subsec:multistage_related_work}

Adaptively adjusting hyperparameters throughout the training is key to the success of deep model training, but most of such explorations are in non-FL context. For example, previous works \cite{He19ControlBatch}, \cite{Sun22Hyperparameters}, and \cite{Sun23Diffusion} reveal the connection between hyperparameters and generalization capacity of optimizers. \cite{Krizhevsky12ImageNet} and \cite{He16Res} propose to decay learning rate $\eta$ whenever the loss saturates; \cite{GoyalDGNWKTJH17LargeMinibatch} popularizes the heuristic of warmup to increase $\eta$ from a small value to a very large value in the first few iterations; \cite{Smith17Cyclic} proposes to adopt a cyclic learning rate schedule between warmup and decay phases. Apart from learning rate, adaptively scheduling other hyperparameters (e.g. momentum factor and batch size) is also shown to be very effective in many settings. For example, \cite{Sutskever13Init} shows a slowly increasing schedule for the momentum factor is crucial; \cite{Smith18Bayesian,Smith18DontDecay} propose a procedure to enable large batch training where $\eta$ and momentum factor $\beta$ are increasing, and batch size $B$ is scaled $B \propto \frac{\eta}{1-\beta}$. From a theoretical point of view, in non-FL context, \cite{Ge2019TheSD} and \cite{wang21stepdecay} show the multistage learning rate scheduler achieves a near-optimal convergence rate of $\mathcal{O}\left(\frac{\log T}{T}\right)$ rate (faster than polynomial decay), in both convex and non-convex functions. \cite{sun21stagewise} and \cite{Sun23TKDD} show the convergence of multistage scheduler for momentum schemes. However, to our best knowledge, there is no prior work studying multistage hyperparameter scheduler (both learning rate and momentum factor) in FL settings.

\subsection{Flexible Participation and Asynchronous Aggregation}
\label{subsec:asyn_aggregation_related_work}

The existing works that are dedicated to studying system heterogeneity can be mainly categorized into the following groups,

\begin{itemize}[leftmargin=*]
    \item Heterogeneous local computing but synchronous aggregation. \cite{Wang20FedNova} is probably the first work to show heterogeneous number of local updates results in the global model converges to a mismatched optimum which can be arbitrarily away from the true optimum and proposes an effective remedy FedNova to correct the mismatch. \cite{Basu19Qsparse-Local-SGD} considers how model compression works with different number of local updates \footnote{Though the authors call the proposed model 'asynchronous', the asynchrony refers to that updates occur after different number of local iterations but the local iterations are synchronous with respect to the global clock. However, 'asynchronous' in our context refers to local iterations are asynchronous with respect to the global clock, which is more challenging to analyze.}. \cite{Avdiukhin21arbitrarycommunication} focuses on a similar setting and studies the upper bound of distances between two consecutive communications to reduce communication times as much as possible. Their theoretical analysis relies on bounded gradient assumption.
    \item Asynchronous aggregation. This line of research is most closely related to our proposed Autonomous Multistage FedGM. However, though asynchrony has been a decade-long topic in traditional distributed computing \citep{Zhang2014DeepElasticAvg,Lian15Asynchronous,Zheng17ASGD}, it has received very limited attention in federated learning. \cite{Xie2019AsynchronousFO} proposes FedAsync in which the server immediately updates the global model whenever it receives a single local model. Their theoretical analysis only applies to convex objective function which is not applicable to deep learning. Moreover, this proposal has negative implication in privacy, as it no longer hides one single update in an aggregate, which is one of the most important points to use FL in the first place. In light of this, \cite{Nguyen2021FedBuff} proposes FedBuff, in which a global update is triggered when the server receives $m$ local updates, where $m$ is a pre-specified hyperparameter. By maintaining a size $m$ buffer, FedBuff could secure the identity of each local update and is empirically faster than FedAsync. However, it does not consider the heterogeneous local computing and does not provide a convergence rate that shows the dependency on $m$. \cite{Yang2021AnarchicFL} proposes anarchic FL in which the clients are free to determine how much local computation to conduct and the asynchronous communication is in the same fashion as FedBuff. However, \cite{Yang2021AnarchicFL} only considers the case of vanilla FedAvg, while our works subsumes anarchic FL as a special case.
\end{itemize}

There are many other works that enable flexible participation scheme but still synchronous aggregation, e.g. \citep{Yan2020DistributedClient, gu2021arbitraryunavailable,wang2022arbitraryparticipation, Nishio2018ClientSelection, Chen2020ClientSampling, cho22biased_selection, Goetz2019ActiveFL,Ribero2020clientsampling}. This line of research is less related to our proposed research.

\section{Proof of Theorem \ref{multistage_fedgm_full_participation_convergence_theorem}, Corollary \ref{corollary:multistage_fedavg_full_participation}, and Corollary \ref{corollary:multistage_fedgm_full_participation}}
\label{sec:proof_multistage_fedgm_full_participation}

\begin{proof}[Proof of Multistage FedGM with Full Participation]

Recall the formulation of General Momentum:
\begin{equation}
    \begin{gathered}
    d_{t+1}=\left(1-\beta_t\right)\Delta_t+\beta_t d_t\\
    x_{t+1}=x_t-\eta_t\left[\left(1-\nu_t\right)\Delta_t+\nu_t d_{t+1}\right]
\end{gathered}\nonumber
\end{equation}
Denote the update sequence $y_t\triangleq x_{t+1}-x_t$. The updating rule is different from FedAvg in that $y_t\ne -\eta_t \Delta_t$. The proof hinges on the construction of an auxiliary sequence $\{z_t\}_{t=0}^{T}$, such that $z_{t+1}-z_t= - \eta_t \Delta_t$. This $\{z_t\}_{t=0}^{T}$ is more like vanilla FedAvg iterates and thus easier to deal with. We then study the property of $\{z_t\}_{t=0}^{T}$ and its connection to $\{x_t\}_{t=0}^{T}$. $\{z_t\}_{t=0}^{T}$ is devised as follows:
\begin{equation}
\label{auxiliary_seq}
z_t= x_t-\frac{\eta_t\beta_t\nu_t}{1-\beta_t}d_{t}
\end{equation}
where $d_0=0$.

We now verify $z_{t+1}-z_t= - \eta_t \Delta_t$,

\begin{equation}
\begin{gathered}
z_{t+1}-z_t=x_{t+1}-\frac{\eta_{t+1}\beta_{t+1}\nu_{t+1}}{1-\beta_{t+1}} d_{t+1} - x_{t}+\frac{\eta_{t}\beta_{t}\nu_{t}}{1-\beta_{t}} d_t\\
\underset{(i)}{=}-\eta_t y_t-W_1(d_{t+1}-d_t)\\
\underset{(ii)}{=}-\eta_t\left(\left(1-\nu_t\right)\Delta_t+\nu_t d_{t+1}\right)-W_1\left(\left(1-\beta_t\right)\Delta_t+\beta_t d_t-d_t\right)\\
=-\eta_t\left(1-\nu_t\right)\Delta_t-\eta_t\beta_t\nu_t\Delta_t-\eta_t\nu_t\left(d_{t+1}-\beta_t d_t\right)\\
=-\eta_t\left(1-\nu_t\right)\Delta_t-\eta_t\beta_t\nu_t\Delta_t-\eta_t\nu_t\left(1-\beta_t\right)\Delta_t=-\eta_t\Delta_t\\
\end{gathered}\nonumber
\end{equation}

where $(i)$ holds by the assumption $\frac{\eta_{t}\beta_{t}\nu_{t}}{1-\beta_{t}}$ is a constant $W_1$, $(ii)$ holds by plugging in the updating rule for $d_t$ and $x_t$.

Since $f$ is $L$-smooth, taking conditional expectation with respect to all randomness prior to step $t$, we have

\begin{equation}
\begin{gathered}
\mathbb{E}\left[f(z_{t+1})\right]\leq f(z_t)+\mathbb{E}\left[\left\langle \nabla f(z_t),z_{t+1}-z_t \right\rangle\right]+\frac{L}{2}\mathbb{E}\left[\left\| z_{t+1}-z_t\right\|^2\right]\\
\leq f(z_t)+\mathbb{E}\left[\left\langle \nabla f(z_t),-\eta_t \Delta_t \right\rangle\right]+\frac{L}{2}\eta_t^2\mathbb{E}\left[\left\| \Delta_t\right\|^2\right]\\
\le f(z_t)+ \underbrace{\mathbb{E}\left[\left\langle \sqrt{\eta_t} \left(\nabla f(z_t)-\nabla f(x_t)\right),-\sqrt{\eta_t} \Delta_t \right\rangle\right]}_{A_1} + \underbrace{\mathbb{E}\left[\left\langle \nabla f(x_t),-\eta_t \Delta_t \right\rangle\right]}_{A_2} + \underbrace{\frac{L}{2}\eta_t^2\mathbb{E}\left[\left\| \Delta_t\right\|^2\right]}_{A_3}
\end{gathered}\nonumber
\end{equation}

\textbf{Bounding} $A_1$:
\begin{equation}
\begin{gathered}
A_1 =\mathbb{E}\left[\left\langle \sqrt{\eta_t} \left(\nabla f(z_t)-\nabla f(x_t)\right),-\sqrt{\eta_t} \Delta_t \right\rangle\right]\\
\underset{(i)}{\leq}\mathbb{E}\left[\left\|\sqrt{\eta_t} \left(\nabla f(z_t)-\nabla f(x_t)\right)\right\| \cdot \left\|-\sqrt{\eta_t} \Delta_t\right\|\right]\\
\underset{(ii)}{\leq} \frac{1}{2} \eta_t^3 L^2 \left(\frac{\beta_t\nu_t}{1-\beta_t}\right)^2\mathbb{E}\left[\left\| d_t\right\|^2\right] + \frac{1}{2}\eta_t\mathbb{E}\left[\left\|\Delta_t\right\|^2\right]
\end{gathered}\nonumber
\end{equation}

where $(i)$ holds by applying Cauchy-Schwarz inequality, and $(ii)$ follows by invoking the definition of $z_t$, Young’s inequality and $f$ is $L$-smooth.

\textbf{Bounding} $A_2$:
\begin{equation}
\begin{gathered}
A_2=\mathbb{E}\left[\left\langle \nabla f(x_t),-\eta_t \Delta_t \right\rangle\right]\\
=\eta_t\mathbb{E}\left[\left\langle \nabla f(x_t),\eta_l K \nabla f(x_t) - \Delta_t - \eta_l K \nabla f(x_t)  \right\rangle\right]\\
\underset{(i)}{=}-\eta_t \eta_l K \mathbb{E}\left[\left\|\nabla f(x_t) \right\|^2\right] + \eta_t\mathbb{E}\left[\left\langle \nabla f(x_t), \eta_l K \nabla f(x_t) - \frac{1}{n}\sum_{i=1}^n\sum_{k=0}^{K-1}\eta_l g_{t,k}^i \right\rangle\right]
\end{gathered}\nonumber
\end{equation}

where $(i)$ follows from the definition $\Delta_t=\frac{1}{n}\sum_{i=1}^n\sum_{k=0}^{K-1}\eta_l g_{t,k}^i$.

where we further bound $\eta_t\mathbb{E}\left[\left\langle \nabla f(x_t), \eta_l K \nabla f(x_t) - \frac{1}{n}\sum_{i=1}^n\sum_{k=0}^{K-1}\eta_l g_{t,k}^i \right\rangle\right]$,
\begin{equation}
\begin{gathered}
\eta_t\mathbb{E}\left[\left\langle \nabla f(x_t), \eta_l K \nabla f(x_t) - \frac{1}{n}\sum_{i=1}^n\sum_{k=0}^{K-1}\eta_l g_{t,k}^i \right\rangle\right]\\
\underset{(i)}{=}\eta_t\mathbb{E}\left[\left\langle \nabla f(x_t), \eta_l K \nabla f(x_t) - \frac{1}{n}\sum_{i=1}^n\sum_{k=0}^{K-1}\eta_l \nabla f_i(x_{t,k}^i) \right\rangle\right]\\
\leq \eta_t \mathbb{E}\left[ \left\langle \sqrt{\eta_l K} \nabla f(x_t), \frac{\sqrt{\eta_l K}}{K n} \sum_{i=1}^n\sum_{k=0}^{K-1} \left( \nabla f_i(x_t)-\nabla f_i(x_{t,k}^i) \right) \right\rangle \right]\\
\underset{(ii)}{\leq} \frac{\eta_t \eta_l K}{2} \mathbb{E}\left[ \left\| \nabla f(x_t) \right\|^2\right] + \frac{\eta_t \eta_l K}{2K^2 n^2} \mathbb{E}\left[\left\| \sum_{i=1}^n\sum_{k=0}^{K-1} \left(\nabla f_i(x_t) - \nabla f_i(x_{t,k}^i) \right)\right\|^2\right]\\
-\frac{\eta_t}{2}\mathbb{E}\left[\left\|\sqrt{\eta_l K}(\nabla f(x_t)-\frac{1}{Kn}\sum_{i=1}^n\sum_{k=0}^{K-1} \left(\nabla f_i(x_t)-\nabla f_i(x_{t,k}^i))\right)\right\|^2\right] = \frac{\eta_t \eta_l K}{2}\mathbb{E}\left[\left\|\nabla f(x_t)\right\|^2\right] \\
+ \frac{\eta_t\eta_l}{2Kn^2}\mathbb{E}\left[\left\|\sum_{i=1}^n\sum_{k=0}^{K-1} \left( \nabla f_i(x_t)-\nabla f_i(x_{t,k}^i) \right)\right\|^2\right]-\frac{\eta_t\eta_l}{2Kn^2}\mathbb{E}\left[\left\| \sum_{i=1}^n\sum_{k=0}^{K-1}\nabla f_i(x_{t,k}^i)\right\|^2\right]
\end{gathered}
\label{proof_eq_1}
\end{equation}

where $(i)$ holds as we take conditional expectation with respect to all randomness prior to step $t$ and $\nabla f(x_t)=\frac{1}{n}\sum_{i=1}^n \nabla f_i(x_t)$ by definition, $(ii)$ holds as $\left\langle a, b \right\rangle = \frac{1}{2} \left\| a \right\|^2 + \frac{1}{2} \left\| b \right\|^2 - \frac{1}{2} \left\| a - b \right\|^2 $.

We further bound Equation \ref{proof_eq_1},

\begin{equation}
\begin{gathered}
\underset{(i)}{\leq} \frac{\eta_t \eta_l K}{2}\mathbb{E}\left[\left\|\nabla f(x_t)\right\|^2\right] + \frac{\eta_t\eta_l}{2 n} \sum_{i=1}^n\sum_{k=0}^{K-1} \mathbb{E}\left[\left\|  \nabla f_i(x_t)-\nabla f_i(x_{t,k}^i) \right\|^2\right] - \frac{\eta_t\eta_l}{2Kn^2}\mathbb{E}\left[\left\| \sum_{i=1}^n\sum_{k=0}^{K-1}\nabla f_i(x_{t,k}^i)\right\|^2\right]\\
\underset{(ii)}{\leq} \frac{\eta_t \eta_l K}{2}\mathbb{E}\left[\left\|\nabla f(x_t)\right\|^2\right] + \frac{\eta_t\eta_l L^2}{2 n} \sum_{i=1}^n\sum_{k=0}^{K-1} \mathbb{E}\left[\left\|  x_t - x_{t,k}^i \right\|^2\right] - \frac{\eta_t\eta_l}{2Kn^2}\mathbb{E}\left[\left\| \sum_{i=1}^n\sum_{k=0}^{K-1}\nabla f_i(x_{t,k}^i)\right\|^2\right]
\end{gathered}\nonumber
\end{equation}

where $(i)$ holds as $\left\|\sum_{i=1}^n x_i\right\|^2 \leq n \sum_{i=1}^n\left\| x_i \right\|^2$, and $(ii)$ holds due to $L$-smoothness of $f_i$.

When $\eta_l\le\frac{1}{8KL}$, for any $k$, we have the following from \citep{reddi2020adaptive},

\begin{equation}
\begin{gathered}
\frac{1}{n}\sum_{i=1}^n \mathbb{E}\left[\left\|  x_t - x_{t,k}^i \right\|^2\right] \leq 5K\eta_l^2\left(\sigma_l^2+6K\sigma_g^2\right)+30 K^2 \eta_l^2\mathbb{E}\left[\left\|\nabla f(x_t)\right\|^2\right]
\end{gathered}\nonumber
\end{equation}

Thus, we have the following,

\begin{equation}
\begin{gathered}
\frac{\eta_t \eta_l K}{2}\mathbb{E}\left[\left\|\nabla f(x_t)\right\|^2\right] + \frac{\eta_t\eta_l L^2}{2 n} \sum_{i=1}^n\sum_{k=0}^{K-1} \mathbb{E}\left[\left\|  x_t - x_{t,k}^i \right\|^2\right] - \frac{\eta_t\eta_l}{2Kn^2}\mathbb{E}\left[\left\| \sum_{i=1}^n\sum_{k=0}^{K-1}\nabla f_i(x_{t,k}^i)\right\|^2\right]\\
\leq \frac{\eta_t \eta_l K}{2}\mathbb{E}\left[\left\|\nabla f(x_t)\right\|^2\right] - \frac{\eta_t\eta_l}{2Kn^2}\mathbb{E}\left[\left\| \sum_{i=1}^n\sum_{k=0}^{K-1}\nabla f_i(x_{t,k}^i)\right\|^2\right]\\
+ \frac{\eta_t\eta_l L^2 K}{2} \left( 5K\eta_l^2\left(\sigma_l^2+6K\sigma_g^2\right)+30 K^2 \eta_l^2\mathbb{E}\left[\left\|\nabla f(x_t)\right\|^2\right] \right) \\
\leq \left(\frac{\eta_t \eta_l K}{2}+15 \eta_t \eta_l^3 K^3  L^2  \right) \mathbb{E}\left[\left\|\nabla f(x_t)\right\|^2\right] - \frac{\eta_t\eta_l}{2Kn^2}\mathbb{E}\left[\left\| \sum_{i=1}^n\sum_{k=0}^{K-1}\nabla f_i(x_{t,k}^i)\right\|^2\right] + \frac{5}{2}\eta_t\eta_l^3L^2K^2\left(\sigma_l^2+6K\sigma_g^2\right) \\
\underset{(i)}{\leq} \frac{47}{64}\eta_t \eta_l K \mathbb{E}\left[\left\|\nabla f(x_t)\right\|^2\right] - \frac{\eta_t\eta_l}{2Kn^2}\mathbb{E}\left[\left\| \sum_{i=1}^n\sum_{k=0}^{K-1}\nabla f_i(x_{t,k}^i)\right\|^2\right] + \frac{5}{2}\eta_t\eta_l^3L^2K^2\left(\sigma_l^2+6K\sigma_g^2\right)
\end{gathered}\nonumber
\end{equation}

where $(i)$ holds as $\eta_l\le\frac{1}{8KL}$.

Merging all pieces together, we have the bound for $A_2$,

\begin{equation}
\begin{gathered}
A_2 = -\eta_t \eta_l K \mathbb{E}\left[\left\|\nabla f(x_t) \right\|^2\right] + \eta_t\mathbb{E}\left[\left\langle \nabla f(x_t), \eta_l K \nabla f(x_t) - \frac{1}{n}\sum_{i=1}^n\sum_{k=0}^{K-1}\eta_l g_{t,k}^i \right\rangle\right]\\
\leq -\frac{17}{64}\eta_t \eta_l K \mathbb{E}\left[\left\|\nabla f(x_t) \right\|^2\right] + \frac{5}{2}\eta_t\eta_l^3L^2K^2\left(\sigma_l^2+6K\sigma_g^2\right)
-\frac{\eta_t\eta_l}{2Kn^2} \mathbb{E}\left[\left\| \sum_{i=1}^n\sum_{k=0}^{K-1} \nabla f_i(x_{t,k}^i)\right\|^2\right]
\end{gathered}\nonumber
\end{equation}

\textbf{Bounding} $\mathbb{E}\left[\left\| \Delta_t\right\|^2\right]$:

\begin{equation}
\begin{gathered}
\mathbb{E}\left[\left\| \Delta_t\right\|^2\right] = \mathbb{E}\left[\left\| \frac{\eta_l}{n}\sum_{i=1}^n\sum_{k=0}^{K-1} g_{t,k}^i \right\|^2\right]\\
\underset{(i)}{=} \mathbb{E}\left[\left\| \frac{\eta_l}{n}\sum_{i=1}^n\sum_{k=0}^{K-1} \left(g_{t,k}^i-\nabla f_i(x_{t,k}^i)\right) \right\|^2\right]+\mathbb{E}\left[\left\| \frac{\eta_l}{n}\sum_{i=1}^n\sum_{k=0}^{K-1}  \nabla f_i(x_{t,k}^i)  \right\|^2\right]\\
\underset{(ii)}{=} \frac{\eta_l^2 }{n^2}\sum_{i=1}^n\sum_{k=0}^{K-1} \mathbb{E}\left[\left\|  g_{t,k}^i-\nabla f_i(x_{t,k}^i) \right\|^2\right]+\mathbb{E}\left[\left\| \frac{\eta_l}{n}\sum_{i=1}^n\sum_{k=0}^{K-1}  \nabla f_i(x_{t,k}^i)  \right\|^2\right]\\
\underset{(iii)}{\leq} \frac{K\eta_l^2}{n} \sigma^2_l+\frac{\eta_l^2}{n^2}\mathbb{E}\left[\left\| \sum_{i=1}^n\sum_{k=0}^{K-1}  \nabla f_i(x_{t,k}^i) \right\|^2\right]
\end{gathered}\nonumber
\end{equation}

where $(i)$ and $(ii)$ hold as $\mathbb{E}\left[\left\|\sum_{i=1}^n x_i\right\|^2\right] = \sum_{i=1}^n \mathbb{E}\left[\left\| x_i \right\|^2\right]$ when $\mathbb{E}\left[ x_i \right]=0$, and we know $\mathbb{E}\left[g_{t,k}^i-\nabla f_i(x_{t,k}^i)\right]=0$. $(iii)$ holds due to bounded local variance assumption.

\textbf{Bounding} $\sum_{t=0}^{T-1}\mathbb{E}\left[\left\| d_t\right\|^2\right]$:

It is straightforward to verify:

\begin{equation}
\begin{gathered}
 d_t = \sum_{p=0}^t a_{t,p}\Delta_p,         \quad  \text{where} \quad  a_{t,p}=\left(1-\beta_p\right)\prod_{q=p+1}^t\beta_q
\end{gathered}\nonumber
\end{equation}

With $d_t = \sum_{p=0}^t a_{t,p}\Delta_p$, we could get,

\begin{equation}
\begin{gathered}
\mathbb{E}\left[\left\| d_t\right\|^2\right]=\mathbb{E}\left[\left\| \sum_{p=0}^t a_{t,p}\Delta_p\right\|^2\right]\\
= \sum_{e=1}^d \mathbb{E}\left[\left( \sum_{p=0}^t a_{t,p}\Delta_{p,e}\right)^2 \right]
\underset{(i)}{\leq} \sum_{e=1}^d \mathbb{E}\left[\left(\sum_{p=0}^t a_{t,p}\right)\cdot\left(\sum_{p=0}^t a_{t,p}\Delta_{p,e}^2\right)\right]\\
\underset{(ii)}{\leq} \left(1- \prod_{q=0}^t \beta_q\right)\sum_{p=0}^t a_{t,p}\mathbb{E}\left[\left\| \Delta_p \right\|^2\right]
\underset{(iii)}{\leq} \frac{K\eta_l^2}{n} \sigma^2_l + \frac{\eta^2_l}{n^2}\sum_{p=0}^t a_{t,p}\mathbb{E}\left[\left\| \sum_{i=1}^n\sum_{k=0}^{K-1}  \nabla f_i(x_{p,k}^i) \right\|^2\right]
\end{gathered}\nonumber
\end{equation}

where $\Delta_{p,e}$ denotes the $e$-th element of vector $\Delta_p$. $(i)$ holds due to Cauchy–Schwarz inequality, $(ii)$ holds as $\sum_{p=0}^t a_{t,p} = 1- \prod_{q=1}^t \beta_q$, $(iii)$ holds by plugging in the bound for $\mathbb{E}\left[\left\| \Delta_t\right\|^2\right]$ and $\beta_q < 1$.

We sum over $t\in \{0,...,T-1\}$,

\begin{equation}
\begin{gathered}
\sum_{t=0}^{T-1}\mathbb{E}\left[\left\| d_t\right\|^2\right] \leq \frac{T K\eta_l^2}{n} \sigma^2_l + \frac{\eta^2_l}{n^2} \sum_{t=0}^{T-1} \sum_{p=0}^t a_{t,p}\mathbb{E}\left[\left\| \sum_{i=1}^n\sum_{k=0}^{K-1}  \nabla f_i(x_{p,k}^i) \right\|^2\right]\\
=\frac{T K\eta_l^2}{n} \sigma^2_l + \frac{\eta^2_l}{n^2}  \sum_{p=0}^{T-1} \left( \sum_{t=p}^{T-1} a_{t,p} \right) \mathbb{E}\left[\left\| \sum_{i=1}^n\sum_{k=0}^{K-1}  \nabla f_i(x_{p,k}^i) \right\|^2\right]
\end{gathered}\nonumber
\end{equation}

Since $\{\beta_t \}_{t=0}^{T-1}$ is a non-decreasing sequence, we could verify $ \sum_{t=p}^{T-1} a_{t,p} \leq \frac{1-\beta_0}{1-\beta_S}= C_\beta$. 
\begin{equation}
\begin{gathered}
\sum_{t=0}^{T-1}\mathbb{E}\left[\left\| d_t\right\|^2\right] \leq \frac{T K\eta_l^2}{n} \sigma^2_l + \frac{\eta^2_l}{n^2}  \sum_{p=0}^{T-1} \left(\sum_{t=p}^{T-1} a_{t,p}\right)\mathbb{E}\left[\left\| \sum_{i=1}^n\sum_{k=0}^{K-1}  \nabla f_i(x_{p,k}^i) \right\|^2\right]\\
\leq \frac{T K\eta_l^2}{n} \sigma^2_l + \frac{\eta^2_l}{n^2}  C_\beta \sum_{t=0}^{T-1} \mathbb{E}\left[\left\| \sum_{i=1}^n\sum_{k=0}^{K-1}  \nabla f_i(x_{t,k}^i) \right\|^2\right]
\end{gathered}\nonumber
\end{equation}

\textbf{Bounding} $A_3$:
\begin{equation}
\begin{gathered}
A_3=\frac{L}{2}\eta_t^2\mathbb{E}\left[\left\| \Delta_t\right\|^2\right]\\
\underset{(i)}{\leq} \frac{L}{2}\eta_t^2 \left(\frac{K\eta_l^2}{n} \sigma^2_l+\frac{\eta_l^2}{n^2}\mathbb{E}\left[\left\| \sum_{i=1}^n\sum_{k=0}^{K-1}  \nabla f_i(x_{t,k}^i) \right\|^2\right]\right)\\
\leq \frac{L K \eta_t^2 \eta_l^2 }{2n}\sigma_l^2 + \frac{L\eta_t^2 \eta_l^2}{2n^2}\mathbb{E}\left[\left\| \sum_{i=1}^n\sum_{k=0}^{K-1}  \nabla f_i(x_{t,k}^i) \right\|^2\right]
\end{gathered}\nonumber
\end{equation}

where $(i)$ holds by plugging in the bound for $\mathbb{E}\left[\left\| \Delta_t\right\|^2\right]$.

Merging $A_1$, $A_2$, $A_3$ together,

\begin{equation}
\begin{gathered}
\mathbb{E}\left[f(z_{t+1})\right] - f(z_t) \leq \underbrace{\mathbb{E}\left[\left\langle \sqrt{\eta_t} \left(\nabla f(z_t)-\nabla f(x_t)\right),-\sqrt{\eta_t} \Delta_t \right\rangle\right]}_{A_1} + \underbrace{\mathbb{E}\left[\left\langle \nabla f(x_t),-\eta_t \Delta_t \right\rangle\right]}_{A_2} + \underbrace{\frac{L}{2}\eta_t^2\mathbb{E}\left[\left\| \Delta_t\right\|^2\right]}_{A_3} \\
\leq \frac{1}{2} \eta_t^3 L^2 \left( \frac{\beta_t\nu_t}{1-\beta_t} \right)^2 \mathbb{E}\left[\left\| d_t\right\|^2\right] + \frac{1}{2}\eta_t \left(\frac{K\eta_l^2}{n} \sigma^2_l+\frac{\eta_l^2}{n^2}\mathbb{E}\left[\left\| \sum_{i=1}^n\sum_{k=0}^{K-1}  \nabla f_i(x_{t,k}^i) \right\|^2\right]\right)\\
-\frac{17}{64}\eta_t \eta_l K\mathbb{E}\left[\left\|\nabla f(x_t)\right\|^2\right] + \frac{5}{2}\eta_t\eta_l^3L^2K^2\left(\sigma_l^2+6K\sigma_g^2\right) -\frac{\eta_t\eta_l}{2Kn^2} \mathbb{E}\left[\left\| \sum_{i=1}^n\sum_{k=0}^{K-1} \nabla f_i(x_{t,k}^i)\right\|^2\right] \\
+\frac{L K \eta_t^2 \eta_l^2 }{2n}\sigma_l^2 + \frac{L\eta_t^2 \eta_l^2}{2n^2}\mathbb{E}\left[\left\| \sum_{i=1}^n\sum_{k=0}^{K-1}  \nabla f_i(x_{t,k}^i) \right\|^2\right]
\end{gathered}\nonumber
\end{equation}

Reorganizing terms, we could get,

\begin{equation}
\begin{gathered}
\frac{17}{64}\eta_t \eta_l K \mathbb{E}\left[\left\|\nabla f(x_t)\right\|^2\right] \leq -\left(\mathbb{E}[f(z_{t+1})] - f(z_t)\right) + \frac{1}{2}\eta_t^3L^2\left(\frac{\beta_t\nu_t}{1-\beta_t}\right)^2\mathbb{E}\left[\left\| d_t\right\|^2\right] \\
+ \frac{1}{2}\eta_t \left(\frac{K\eta_l^2}{n} \sigma^2_l+\frac{\eta_l^2}{n^2}\mathbb{E}\left[\left\| \sum_{i=1}^n\sum_{k=0}^{K-1}  \nabla f_i(x_{t,k}^i) \right\|^2\right]\right)+\frac{5}{2}\eta_t\eta_l^3L^2K^2\left(\sigma_l^2+6K\sigma_g^2\right) \\
-\frac{\eta_t\eta_l}{2Kn^2} \mathbb{E}\left[\left\| \sum_{i=1}^n\sum_{k=0}^{K-1} \nabla f_i(x_{t,k}^i)\right\|^2\right] +\frac{L K \eta_t^2 \eta_l^2 }{2n}\sigma_l^2 + \frac{L\eta_t^2 \eta_l^2}{2n^2}\mathbb{E}\left[\left\| \sum_{i=1}^n\sum_{k=0}^{K-1}  \nabla f_i(x_{t,k}^i) \right\|^2\right]
\end{gathered}\nonumber
\end{equation}

that is,

\begin{equation}
\begin{gathered}
\mathbb{E}\left[\left\|\nabla f(x_t)\right\|^2\right] \leq -\frac{64}{17}\frac{\mathbb{E}\left[f(z_{t+1})\right] - f(z_t)}{\eta_t \eta_l K} + \frac{32}{17}\frac{L^2}{\eta_l K}W_1^2\mathbb{E}\left[\left\| d_t\right\|^2\right] \\
+ \frac{32}{17}\frac{\eta_l}{n}\sigma_l^2 +\frac{32}{17} \frac{\eta_l}{Kn^2}\mathbb{E}\left[\left\| \sum_{i=1}^n\sum_{k=0}^{K-1}  \nabla f_i(x_{t,k}^i) \right\|^2\right] +\frac{160}{17} \eta_l^2L^2K \left(\sigma_l^2+6K\sigma_g^2\right) \\
-\frac{32}{17}\frac{1}{K^2n^2} \mathbb{E}\left[\left\| \sum_{i=1}^n\sum_{k=0}^{K-1} \nabla f_i(x_{t,k}^i)\right\|^2\right] + \frac{32}{17} \frac{L \eta_t \eta_l }{n}\sigma_l^2 +\frac{32}{17} \frac{L\eta_t  \eta_l }{K n^2}\mathbb{E}\left[\left\| \sum_{i=1}^n\sum_{k=0}^{K-1}  \nabla f_i(x_{t,k}^i) \right\|^2\right]
\end{gathered}\nonumber
\end{equation}

Sum over all $S$ stages and take average, we get,

\begin{equation}
\begin{gathered}
\Bar{\mathcal{G}} \triangleq \frac{1}{S} \sum_{s=1}^{S} \frac{1}{T_s} \sum_{t=T_0+\dots+T_{s-1}}^{T_0+\dots+T_s-1} \mathbb{E}\left[\left\|\nabla f(x_t)\right\|^2\right]\\
\underset{(i)}{\leq} \frac{64}{17}\frac{f(z_0)-\mathbb{E}[f(z_{T})]}{S W_2 \eta_l K} + \frac{32}{17} \frac{L^2 W_1^2 \Bar{\eta} }{ W_2\eta_l K}\sum_{t=0}^{T-1}\mathbb{E}\left[\left\| d_t\right\|^2\right]+\frac{32}{17}\frac{\eta_l}{n}\sigma_l^2\\
+\frac{32}{17} \frac{\eta_l\Bar{\eta}}{Kn^2  W_2}\sum_{t=0}^{T-1}\mathbb{E}\left[\left\| \sum_{i=1}^n\sum_{k=0}^{K-1}  \nabla f_i(x_{t,k}^i) \right\|^2\right] +\frac{160}{17} \eta_l^2L^2K \left(\sigma_l^2+6K\sigma_g^2\right)\\
-\frac{32}{17}\frac{\eta_S}{S W_2 K^2 n^2}\sum_{t=0}^{T-1} \mathbb{E}\left[\left\| \sum_{i=1}^n\sum_{k=0}^{K-1} \nabla f_i(x_{t,k}^i)\right\|^2\right] 
+ \frac{32}{17} \frac{L \Bar{\eta} \eta_l }{n}\sigma_l^2 + \frac{32}{17} \frac{L \eta_0\Bar{\eta}  \eta_l }{ W_2 K n^2}\sum_{t=0}^{T-1}\mathbb{E}\left[\left\| \sum_{i=1}^n\sum_{k=0}^{K-1}  \nabla f_i(x_{t,k}^i) \right\|^2\right]\\
\underset{(ii)}{\leq} \frac{64}{17}\frac{f(z_0)-\mathbb{E}\left[f(z_{T})\right]}{S W_2 \eta_l K} + \frac{T K\eta_l^2}{n} \sigma^2_l \frac{32}{17} \frac{L^2 W_1^2 \Bar{\eta}}{W_2\eta_l K} 
+\frac{32}{17}\frac{\eta_l}{n}\sigma_l^2  +\frac{160}{17} \eta_l^2L^2K \left(\sigma_l^2+6K\sigma_g^2\right)+\frac{32}{17} \frac{L \Bar{\eta} \eta_l }{n}\sigma_l^2\\
+\left(\frac{\eta^2_l}{n^2} C_\beta \frac{32}{17} \frac{L^2 W_1^2 \Bar{\eta}}{W_2\eta_l K} +  \frac{32}{17} \frac{\eta_l\Bar{\eta}}{Kn^2 W_2} - \frac{32}{17}\frac{\eta_S}{S W_2K^2n^2} + \frac{32}{17} \frac{L \eta_0\Bar{\eta}  \eta_l }{W_2 K n^2} \right)\sum_{t=0}^{T-1}\mathbb{E}\left[\left\| \sum_{i=1}^n\sum_{k=0}^{K-1}  \nabla f_i(x_{t,k}^i) \right\|^2\right]
\end{gathered}\nonumber
\end{equation}

where $\Bar{\eta}=\frac{1}{S}\sum_{s=1}^S \eta_s$. Due to $\eta_t$ is stagewise, i.e. $\eta_t=\eta_s$ when $t\in \{T_0+\dots+T_{s-1},\dots,T_0+\dots+T_s-1\}$, and $\eta_s$ is decaying, i.e. $\eta_S \leq \eta_s \leq \eta_0$, for any stage $s$, thus we have the following,

\begin{equation}
\begin{gathered}
\frac{1}{S} \sum_{s=1}^{S} \frac{1}{T_s} \sum_{t=T_0+\dots+T_{s-1}}^{T_0+\dots+T_s-1}\frac{32}{17}\frac{L^2W_1^2}{\eta_l K}\mathbb{E}\left[\left\| d_t\right\|^2\right] = \frac{32}{17}\frac{L^2 W_1^2}{\eta_l K}  \frac{1}{S} \sum_{s=1}^{S} \frac{ \eta_s }{ T_s \eta_s} \sum_{t=T_0+\dots+T_{s-1}}^{T_0+\dots+T_s-1} \mathbb{E}\left[\left\| d_t\right\|^2\right]\\
= \frac{32}{17}\frac{L^2 W_1^2}{\eta_l K} \frac{1}{SW_2} \sum_{s=1}^{S} \eta_s \sum_{t=T_0+\dots+T_{s-1}}^{T_0+\dots+T_s-1} \mathbb{E}\left[\left\| d_t\right\|^2\right] \leq \frac{32}{17}\frac{L^2 W_1^2 \Bar{\eta}}{\eta_l K W_2}\sum_{s=1}^{S}\sum_{t=T_0+\dots+T_{s-1}}^{T_0+\dots+T_s-1} \mathbb{E}\left[\left\| d_t\right\|^2\right] \\
= \frac{32}{17}\frac{L^2 W_1^2 \Bar{\eta}}{\eta_l K W_2}\sum_{t=0}^{T-1} \mathbb{E}\left[\left\| d_t\right\|^2\right]
\end{gathered}\nonumber
\end{equation}

Similarly, we have,

\begin{equation}
\begin{gathered}
\frac{1}{S} \sum_{s=1}^{S} \frac{1}{T_s} \sum_{t=T_0+\dots+T_{s-1}}^{T_0+\dots+T_s-1} \frac{32}{17} \frac{L\eta_t  \eta_l }{K n^2}\mathbb{E}\left[\left\| \sum_{i=1}^n\sum_{k=0}^{K-1}  \nabla f_i(x_{t,k}^i) \right\|^2\right] \leq \frac{32}{17} \frac{L \eta_0\Bar{\eta}  \eta_l }{ W_2 K n^2}\sum_{t=0}^{T-1}\mathbb{E}\left[\left\| \sum_{i=1}^n\sum_{k=0}^{K-1}  \nabla f_i(x_{t,k}^i) \right\|^2\right]
\end{gathered}\nonumber
\end{equation}

and, we also have,

\begin{equation}
\begin{gathered}
\frac{1}{S} \sum_{s=1}^{S} \frac{1}{T_s} \sum_{t=T_0+\dots+T_{s-1}}^{T_0+\dots+T_s-1} \frac{32}{17} \frac{L \eta_t \eta_l }{n}\sigma_l^2 = \frac{32}{17}\frac{L \eta_l }{n}\sigma_l^2\frac{1}{S} \sum_{s=1}^{S} \frac{1}{T_s} \sum_{t=T_0+\dots+T_{s-1}}^{T_0+\dots+T_s-1}\eta_t = \frac{32}{17}\frac{L \eta_l \Bar{\eta}}{n}\sigma_l^2
\end{gathered}\nonumber
\end{equation}

Thus, inequality $(i)$ holds. $(ii)$ holds by plugging into the bounds for $\sum_{t=0}^{T-1}\mathbb{E}\left[\left\|d_t\right\|^2\right]$.

when the following two conditions hold,
\begin{equation}
\begin{gathered}
\eta_l\leq \frac{1}{K S C_\eta \left(L \Bar{\eta}  + 1 + L^2 W_1^2 C_\eta \right)}
\end{gathered}\nonumber
\end{equation}

where $C_\eta=\frac{\eta_0}{\eta_S}$.

we could verify the coefficient of $\sum_{t=0}^{T-1}\mathbb{E}\left[\left\| \sum_{i=1}^n\sum_{k=0}^{K-1}  \nabla f_i(x_{t,k}^i) \right\|^2\right]$ is non-positive, by plugging in the learning rate constraints and using $C_\beta\leq C_\eta$.
\begin{equation}
\begin{gathered}
\frac{\eta^2_l}{n^2} C_\beta \frac{32}{17} \frac{L^2 W_1^2 \Bar{\eta}}{W_2\eta_l K} +  \frac{32}{17} \frac{\eta_l\Bar{\eta}}{Kn^2 W_2} - \frac{32}{17}\frac{\eta_S}{S W_2K^2n^2} + \frac{32}{17} \frac{L \hat{\eta}^2  \eta_l }{W_2 K n^2} \leq 0
\end{gathered}\nonumber
\end{equation}
which results in,

\begin{equation}
\begin{gathered}
\Bar{\mathcal{G}} \triangleq \frac{1}{S} \sum_{s=1}^{S} \frac{1}{T_s} \sum_{t=T_0+\dots+T_{s-1}}^{T_0+\dots+T_s-1} \mathbb{E}\left[\left\|\nabla f(x_t)\right\|^2\right]\\
\leq \frac{64}{17}\frac{f(z_0)-\mathbb{E}\left[f(z_{T})\right]}{S W_2 \eta_l K} + \frac{T K\eta_l^2}{n} \sigma^2_l \frac{32}{17} \frac{L^2 W_1^2 \Bar{\eta}}{W_2\eta_l K} 
+\frac{32}{17}\frac{\eta_l}{n}\sigma_l^2  +\frac{160}{17} \eta_l^2L^2K \left(\sigma_l^2+6K\sigma_g^2\right)+\frac{32}{17} \frac{L \Bar{\eta} \eta_l }{n}\sigma_l^2\\
\underset{(i)}{\leq} \frac{64}{17}\frac{f(x_0)-f^\ast}{S W_2 \eta_l K} + \left( \frac{32}{17}\frac{L^2 W_1^2 T \Bar{\eta} \eta_l}{n W_2} +\frac{32}{17}\frac{\eta_l}{n}+\frac{160}{17} \eta_l^2L^2K+\frac{32}{17} \frac{L \Bar{\eta} \eta_l }{n}\right)  \sigma_l^2+
\frac{960}{17} \eta_l^2L^2K^2 \sigma_g^2
\end{gathered}\nonumber
\end{equation}

where $(i)$ holds as $f$ is assumed to have minimum $f^\ast$.

Suppose $S=1$, i.e. the typical constant hyperparameter regime, the total number of rounds are $T$, $\Bar{\eta}=\eta_0=\Theta\left(\sqrt{nK}\right)$ and $\eta_l=\Theta\left(\frac{1}{\sqrt{T}K}\right)$, $W_2=\Theta\left(T\sqrt{nK}\right)$ in this case. Suppose $W_1^2=\mathcal{O}\left(\sqrt{nK}\right)$. Considering in FedAvg, $\beta=0$ and consequently $W_1=0$, thus, the condition $W_1^2=\mathcal{O}\left(\sqrt{nK}\right)$ naturally holds. We have the bound as,

\begin{equation}
\begin{gathered}
\Bar{\mathcal{G}}\triangleq\frac{1}{S}\sum_{s=0}^{S-1} \frac{1}{T_s}\sum_{t=T_0+\dots+T_{s-1} }^{T_0+\dots+T_s-1} \mathbb{E}\left[\left\|\nabla f(x_t)\right\|^2\right]\\
\leq  \mathcal{O}\left(\frac{1}{\sqrt{TKn}}\right) \left(f(x_0)-f^\ast\right) + \left(\mathcal{O}\left(\frac{1}{\sqrt{TKn}}\right)+ \mathcal{O}\left(\frac{1}{TK}\right) + \mathcal{O}\left(\frac{1}{\sqrt{TKn}}\right) \right)  \sigma_l^2+
\mathcal{O}\left(\frac{1}{T}\right) \sigma_g^2
\end{gathered}\nonumber
\end{equation}

when $T$ is sufficiently large, i.e. $T\ge Kn$, the dominant term is $\mathcal{O}\left(\frac{1}{\sqrt{TKn}}\right)$.

Suppose $S>1$, i.e. the multistage regime, the total number of rounds are $T$, $\Bar{\eta} = \Theta\left(\sqrt{nK}\right)$, $\eta_l=\Theta\left(\frac{1}{\sqrt{T}K}\right)$, $W_2=\Theta\left(\frac{T\sqrt{nK}}{S}\right)$, i.e. $T\Bar{\eta}$ is equally divided into $S$ stages. Suppose $W_1^2=\mathcal{O}\left(\frac{\sqrt{nK}}{S}\right)$, we have the bound as,

\begin{equation}
\begin{gathered}
\Bar{\mathcal{G}}\triangleq\frac{1}{S}\sum_{s=0}^{S-1} \frac{1}{T_s}\sum_{t=T_0+\dots+T_{s-1} }^{T_0+\dots+T_s-1} \mathbb{E}\left[\left\|\nabla f(x_t)\right\|^2\right]\\
\leq  \mathcal{O}\left(\frac{1}{\sqrt{TKn}}\right) \left(f(x_0)-f^\ast\right) + \left(\mathcal{O}\left(\frac{1}{\sqrt{TKn}}\right)+ \mathcal{O}\left(\frac{1}{TK}\right) + \mathcal{O}\left(\frac{1}{\sqrt{TKn}}\right) \right)  \sigma_l^2+
\mathcal{O}\left(\frac{1}{T}\right) \sigma_g^2
\end{gathered}\nonumber
\end{equation}

The dominant term is $\mathcal{O}\left(\frac{1}{\sqrt{TKn}}\right)$.

\end{proof}

\section{Proof of Theorem \ref{multistage_fedgm_partial_participation_convergence_theorem} and Corollary \ref{corollary:fedgm_partial_participation_convergence_rate}}
\label{sec:proof_multistage_fedgm_partial_participation}

\begin{proof}[Proof of Multistage GM with Partial Participation]

Recall the formulation of General Momentum:
\begin{equation}
    \begin{gathered}
    d_{t+1}=(1-\beta_t)\Delta_t+\beta_t d_t\\
    x_{t+1}=x_t-\eta_t[(1-\nu_t)\Delta_t+\nu_t d_{t+1}]
\end{gathered}\nonumber
\end{equation}

We first show $\Delta_t$ is an unbiased estimator of a virtual average $\Delta^\prime_t \triangleq \frac{1}{n}\sum_{i=1}^n\Delta_t^i$,

\begin{equation}
\begin{gathered}
\mathbb{E}\left[\Delta_t\right] = \mathbb{E}\left[ \frac{1}{m}\sum_{i\in\mathcal{S}_t}  \Delta_t^i  \right] = \mathbb{E}\left[ \frac{1}{m} \sum_{i=1}^n \mathbf{1}\left( i\in\mathcal{S}_t \right) \Delta_t^i  \right]  \\
= \frac{1}{m} \mathbb{E}\left[ \sum_{i=1}^n \mathbb{P}\left\{i\in\mathcal{S}_t\right\} \Delta_t^i  \right] \underset{(i)}{=} \frac{1}{n}\sum_{i=1}^n\Delta_t^i=\Delta^\prime_t
\end{gathered}\nonumber
\end{equation}

where $(i)$ follows from $\mathbb{P}\left\{i\in\mathcal{S}_t\right\}=\frac{m}{n}$.

We study the following Lyapunov sequence $\{z_t\}_{t=0}^{T-1}$, which is devised as follows:

\begin{equation}
\label{auxiliary_seq}
z_t= x_t-\frac{\eta_t\beta_t\nu_t}{1-\beta_t}d_{t}
\end{equation}
where $d_{0}=0$.

We now verify $z_{t+1}-z_t= - \eta_t \Delta_t$,

\begin{align*}
z_{t+1}-z_t&=x_{t+1}-\frac{\eta_{t+1}\beta_{t+1}\nu_{t+1}}{1-\beta_{t+1}} d_{t+1} - x_{t}+\frac{\eta_{t}\beta_{t}\nu_{t}}{1-\beta_{t}} d_{t}\\
&=-\eta_t y_t-W_1(d_{t+1}-d_t)\\
&=-\eta_t((1-\nu_t)\Delta_t+\nu_t d_{t+1})-W_1((1-\beta_t)\Delta_t+\beta_t d_t-d_t)\\
&=-\eta_t(1-\nu_t)\Delta_t-\eta_t\beta_t\nu_t\Delta_t-\eta_t\nu_t(d_{t+1}-\beta_t d_t)\\
&=-\eta_t(1-\nu_t)\Delta_t-\eta_t\beta_t\nu_t\Delta_t-\eta_t\nu_t(1-\beta_t)\Delta_t=-\eta_t\Delta_t\\
\end{align*}

Since $f$ is $L$-smooth, taking conditional expectation with respect to all randomness prior to step $t$, we have

\begin{equation}
\begin{gathered}
\mathbb{E}\left[f(z_{t+1})\right] \leq f(z_t)+\mathbb{E}\left[\left\langle \nabla f(z_t),z_{t+1}-z_t \right\rangle \right] + \frac{L}{2}\mathbb{E}\left[\left\| z_{t+1}-z_t \right\|^2\right]\\
\leq f(z_t)+\mathbb{E}\left[\left\langle \nabla f(z_t),-\eta_t \Delta_t \right\rangle\right] + \frac{L}{2}\eta_t^2\mathbb{E}\left[\left\| \Delta_t \right\|^2\right]\\
\leq f(z_t)+ \underbrace{\mathbb{E}\left[\left\langle \sqrt{\eta_t} \left(\nabla f(z_t)-\nabla f(x_t)\right),-\sqrt{\eta_t} \Delta_t \right\rangle\right]}_{A_1} + \underbrace{\mathbb{E}\left[\left\langle \nabla f(x_t),-\eta_t \Delta_t \right\rangle\right]}_{A_2} + \underbrace{\frac{L}{2}\eta_t^2\mathbb{E}\left[\left\| \Delta_t \right\|^2\right]}_{A_3} \\
\end{gathered}\nonumber
\end{equation}

\textbf{Bounding} $A_1$:
\begin{equation}
\begin{gathered}
A_1 =\mathbb{E}\left[\left\langle \sqrt{\eta_t} \left(\nabla f(z_t)-\nabla f(x_t)\right),-\sqrt{\eta_t} \Delta_t \right\rangle\right]\\
\underset{(i)}{\leq}\mathbb{E}\left[\left\| \sqrt{\eta_t} \left(\nabla f(z_t)-\nabla f(x_t)\right) \right\| \cdot \left\| -\sqrt{\eta_t} \Delta_t\right\|\right]\\
\underset{(ii)}{\leq} \frac{1}{2}\eta_t^3L^2\left(\frac{\beta_t\nu_t}{1-\beta_t}\right)^2\mathbb{E}\left[\left\| d_t\right\|^2\right] + \frac{1}{2}\eta_t\mathbb{E}\left[\left\| \Delta_t \right\|^2 \right]
\end{gathered}
\end{equation}

where $(i)$ holds by applying Cauchy-Schwarz inequality, and $(ii)$ follows from Young’s inequality and $f$ is $L$-smooth.

\textbf{Bounding} $A_2$:
\begin{equation}
\begin{gathered}
A_2=\mathbb{E}\left[\left\langle \nabla f(x_t),-\eta_t \Delta_t \right\rangle\right]\\
=\eta_t\mathbb{E}\left[\left\langle \nabla f\left(x_t\right),\eta_l K \nabla f\left(x_t\right) - \Delta_t - \eta_l K \nabla f\left(x_t\right)  \right\rangle\right]\\
=-\eta_t \eta_l K \mathbb{E}\left [\left\| \nabla f\left(x_t\right) \right\|^2\right]+\eta_t\mathbb{E}\left[\left\langle \nabla f\left(x_t\right), \eta_l K \nabla f\left(x_t\right) - \Delta_t \right\rangle\right]
\end{gathered}\nonumber
\end{equation}
where we further bound $\eta_t\mathbb{E}\left[\left\langle \nabla f\left(x_t\right), \eta_l K \nabla f\left(x_t\right) - \Delta_t \right\rangle\right]$,
\begin{equation}
\begin{gathered}
\eta_t\mathbb{E}\left[\left\langle \nabla f\left(x_t\right), \eta_l K \nabla f\left(x_t\right) - \Delta_t \right\rangle\right]\\
\underset{(i)}{=} \eta_t\mathbb{E}\left[\left\langle \nabla f\left(x_t\right), \eta_l K \nabla f\left(x_t\right) - \Delta^\prime_t \right\rangle\right]=
\eta_t\mathbb{E}\left[\left\langle \nabla f\left(x_t\right), \eta_l K \nabla f\left(x_t\right) - \frac{1}{n}\sum_{i=1}^n\sum_{k=0}^{K-1}\eta_l g_{t,k}^i \right\rangle\right]\\
\underset{(ii)}{=}\eta_t\mathbb{E}\left[\left\langle \nabla f\left(x_t\right), \eta_l K \nabla f\left(x_t\right) - \frac{1}{n}\sum_{i=1}^n\sum_{k=0}^{K-1}\eta_l \nabla f_i(x^i_{t,k}) \right\rangle\right]\\
\underset{(iii)}{=} \eta_t \mathbb{E}\left[\left\langle \nabla f\left(x_t\right), \frac{\eta_l}{n}\sum_{i=1}^n\sum_{k=0}^{K-1} \left(\nabla f_i(x_t)-\nabla f_i(x^i_{t,k})\right) \right\rangle\right]\\
= \eta_t \left\langle \sqrt{\eta_l K} \nabla f(x_t), \frac{\sqrt{\eta_l K}}{K n} \mathbb{E}\left[\sum_{i=1}^n\sum_{k=0}^{K-1} \left(\nabla f_i(x_t)-\nabla f_i(x^i_{t,k})\right)\right] \right\rangle\\
\underset{(iv)}{=} \frac{\eta_t \eta_l K}{2} \mathbb{E} \left[ \left\|\nabla f(x_t)\right\|^2 \right]+\frac{\eta_t \eta_l K}{2 K^2 n^2}\mathbb{E}\left[\left\|\sum_{i=1}^n\sum_{k=0}^{K-1} \left(\nabla f_i(x_t)-\nabla f_i(x^i_{t,k})\right)\right\|^2\right]\\
-\frac{\eta_t}{2}\mathbb{E}\left[\left\|\sqrt{\eta_l K}\left(\nabla f(x_t)-\frac{1}{Kn}\sum_{i=1}^n\sum_{k=0}^{K-1} \left(\nabla f_i(x_t)-\nabla f_i(x^i_{t,k})\right)\right)\right\|^2\right]
\end{gathered}\nonumber
\end{equation}

where $(i)$ holds as $\Delta_t$ is an unbiased estimator of $\Delta^\prime_t$, $(ii)$ holds as we take conditional expectation with respect to all randomness prior to step $t$. $(iii)$ holds due to the following equality and the definition of $\nabla f(x_t)=\frac{1}{n}\sum_{i=1}^n \nabla f_i(x_t)$. $(iv)$ holds as $\left\langle a, b \right\rangle = \frac{1}{2} \left\| a \right\|^2 + \frac{1}{2} \left\| b \right\|^2 - \frac{1}{2} \left\| a - b \right\|^2 $.

We further bound the above terms as,
\begin{equation}
\begin{gathered}
=\frac{\eta_t \eta_l K}{2}\mathbb{E} \left[ \left\|\nabla f(x_t)\right\|^2 \right] + \frac{\eta_t\eta_l}{2Kn^2}\mathbb{E}\left[\left\|\sum_{i=1}^n\sum_{k=0}^{K-1} \left(\nabla f_i(x_t)-\nabla f_i(x^i_{t,k})\right)\right\|^2\right]-\frac{\eta_t\eta_l}{2Kn^2}\mathbb{E}\left[\left\| \sum_{i=1}^n\sum_{k=0}^{K-1}\nabla f_i(x^i_{t,k})\right\|^2\right]\\
\underset{(i)}{\leq} \frac{\eta_t \eta_l K}{2}\mathbb{E} \left[ \left\|\nabla f(x_t)\right\|^2 \right] - \frac{\eta_t\eta_l}{2Kn^2}\mathbb{E}\left[\left\| \sum_{i=1}^n\sum_{k=0}^{K-1}\nabla f_i(x^i_{t,k})\right\|^2\right]+\frac{\eta_t\eta_l L^2}{2Kn^2}\mathbb{E}\left[\left\|\sum_{i=1}^n\sum_{k=0}^{K-1} \left(x_t-x_{t,k}^i\right)\right\|^2\right]\\
\underset{(ii)}{\leq} \frac{\eta_t \eta_l K}{2} \mathbb{E} \left[ \left\|\nabla f(x_t)\right\|^2 \right] - \frac{\eta_t\eta_l}{2Kn^2}\mathbb{E}\left[\left\| \sum_{i=1}^n\sum_{k=0}^{K-1}\nabla f_i(x^i_{t,k})\right\|^2\right]+\frac{\eta_t\eta_l L^2}{2 n }\sum_{i=1}^n\sum_{k=0}^{K-1}\mathbb{E}\left[\left\| x_t-x_{t,k}^i\right\|^2\right]
\end{gathered}\nonumber
\end{equation}

where $(i)$ holds due to $L$-smoothness of $f_i$, $(ii)$ holds as $\left\|\sum_{i=1}^n x_i\right\|^2 \leq n \sum_{i=1}^n\left\| x_i \right\|^2$.

when $\eta_l\le\frac{1}{8KL}$, we have the following bound,

\begin{equation}
\begin{gathered}
\mathbb{E}\left[\left\| x_t-x_{t,k}^i\right\|^2\right]\leq 5K\eta_l^2\left(\sigma_l^2+6K\sigma_g^2\right)+30K^2\eta_l^2\left\|\nabla f(x_t)\right\|^2
\end{gathered}\nonumber
\end{equation}

Plug in the above bound, we would have,

\begin{equation}
\begin{gathered}
\frac{\eta_t \eta_l K}{2}\mathbb{E} \left[ \left\|\nabla f(x_t)\right\|^2 \right]-\frac{\eta_t\eta_l}{2Kn^2}\mathbb{E}\left[\left\| \sum_{i=1}^n\sum_{k=0}^{K-1}\nabla f_i(x^i_{t,k})\right\|^2\right]+\frac{\eta_t\eta_l L^2}{2 n }\sum_{i=1}^n\sum_{k=0}^{K-1}\mathbb{E}\left[\left\| x_t-x_{t,k}^i\right\|^2\right]\\
\leq \frac{\eta_t \eta_l K}{2}\mathbb{E} \left[ \left\|\nabla f(x_t)\right\|^2 \right]-\frac{\eta_t\eta_l}{2Kn^2}\mathbb{E}\left[\left\| \sum_{i=1}^n\sum_{k=0}^{K-1}\nabla f_i(x^i_{t,k})\right\|^2\right]\\
+\frac{\eta_t\eta_l L^2 K}{2 } \left(5K\eta_l^2\left(\sigma_l^2+6K\sigma_g^2\right)+30K^2\eta_l^2\left\|\nabla f(x_t)\right\|^2\right) \\
\le \left(\frac{\eta_t \eta_l K}{2}+30K^2\eta_l^2 \frac{\eta_t\eta_l L^2 K }{2}\right)\mathbb{E} \left[ \left\|\nabla f(x_t)\right\|^2 \right]-\frac{\eta_t\eta_l}{2Kn^2}\mathbb{E}\left[\left\| \sum_{i=1}^n\sum_{k=0}^{K-1}\nabla f_i(x^i_{t,k})\right\|^2\right]\\
+\frac{5}{2}\eta_t\eta_l^3L^2K^2\left(\sigma_l^2+6K\sigma_g^2\right)\\
\underset{(i)}{\leq} \frac{47}{64}\eta_t \eta_l K\mathbb{E} \left[ \left\|\nabla f(x_t)\right\|^2 \right]-\frac{\eta_t\eta_l}{2Kn^2}\mathbb{E}\left[\left\| \sum_{i=1}^n\sum_{k=0}^{K-1}\nabla f_i(x^i_{t,k})\right\|^2\right]+\frac{5}{2}\eta_t\eta_l^3L^2K^2\left(\sigma_l^2+6K\sigma_g^2\right)
\end{gathered}\nonumber
\end{equation}

where $(i)$ holds by using the constraint $\eta_l\le\frac{1}{8KL}$.

Therefore, merging all pieces together, we have,

\begin{equation}
\begin{gathered}
A_2 =-\eta_t \eta_l K \mathbb{E}\left [\left\| \nabla f\left(x_t\right) \right\|^2\right]+\eta_t\mathbb{E}\left[\left\langle \nabla f\left(x_t\right), \eta_l K \nabla f\left(x_t\right) - \Delta_t \right\rangle\right]\\
\le -\frac{17}{64}\eta_t \eta_l K \mathbb{E}\left [\left\| \nabla f\left(x_t\right) \right\|^2\right] - \frac{\eta_t\eta_l}{2Kn^2}\mathbb{E}\left[\left\| \sum_{i=1}^n\sum_{k=0}^{K-1}\nabla f_i(x^i_{t,k})\right\|^2\right]+\frac{5}{2}\eta_t\eta_l^3L^2K^2\left(\sigma_l^2+6K\sigma_g^2\right)
\end{gathered}\nonumber
\end{equation}

\textbf{Bounding} $\mathbb{E}\left[\left\| \Delta_t\right\|^2\right]$:

\begin{equation}
\begin{gathered}
\mathbb{E}\left[\left\| \Delta_t\right\|^2\right] \leq \frac{K\eta_l^2}{m}\sigma_l^2  + \frac{\eta_l^2 \left(m-1\right)}{nm\left(n-1\right)}\mathbb{E}\left[ \left\| \sum_{i=1}^n\sum_{k=0}^{K-1}\nabla f_i(x^i_{t,k}) \right\|^2 \right]\\
+ \frac{\eta_l^2\left(n-m\right)}{n m \left( n - 1 \right)}\left[ 15nK^3L^3\eta_l^2\left( \sigma_l^2+6K\sigma_g^2 \right) +\left(90nK^4L^2\eta_l^2 + 3nK^2\right)\left\|\nabla f(x_t)\right\|^2 +3nK^2\sigma_g^2 \right]
\end{gathered}\nonumber
\end{equation}

\textbf{Bounding} $\sum_{t=0}^{T-1}\mathbb{E}\left[\left\| d_t\right\|^2\right]$:

\begin{equation}
\begin{gathered}
\sum_{t=0}^{T-1}\mathbb{E}\left[\left\| d_t\right\|^2\right] \leq \frac{KT\eta_l^2}{m}\sigma_l^2 + \frac{\eta_l^2}{m^2} C_\beta \sum_{t=0}^{T-1}\mathbb{E}\left[ \left\| \sum_{i\in\mathcal{S}_t}\sum_{k=0}^{K-1}\nabla f_i(x^i_{t,k}) \right\|^2 \right]\\
\leq \frac{KT\eta_l^2}{m}\sigma_l^2 + \frac{\eta_l^2}{m^2}C_\beta\sum_{t=0}^{T-1}\mathbb{E}\left[ \left\| \sum_{i=1}^n \mathbb{P}\left\{i\in\mathcal{S}_t\right\}\sum_{k=0}^{K-1}\nabla f_i(x^i_{t,k}) \right\|^2 \right]\\
\leq \frac{KT\eta_l^2}{m}\sigma_l^2 + \frac{\eta_l^2}{m^2}\frac{m\left(n-m\right)}{n\left(n-1\right)}C_\beta\sum_{t=0}^{T-1}\sum_{i=1}^n\left\|\sum_{k=0}^{K-1}\nabla f_i(x^i_{t,k})\right\|^2 + \frac{\eta_l^2}{m^2}\frac{m\left(m-1\right)}{n\left(n-1\right)}C_\beta\sum_{t=0}^{T-1}\left\| \sum_{i=1}^n\sum_{k=0}^{K-1}\nabla f_i(x^i_{t,k}) \right\|^2\\
\leq \frac{KT\eta_l^2}{m}\sigma_l^2 + \frac{\eta_l^2\left(m-1\right)}{mn\left(n-1\right)} C_\beta \sum_{t=0}^{T-1}\left\| \sum_{i=1}^n\sum_{k=0}^{K-1}\nabla f_i(x^i_{t,k}) \right\|^2 +\\
\frac{\eta_l^2\left(n-m\right)}{m n\left(n-1\right)} C_\beta \sum_{t=0}^{T-1}\sum_{i=1}^n\left\|\sum_{k=0}^{K-1} \left[\nabla f_i(x^i_{t,k}) - \nabla f_i(x^i_{t})\right] + \left[\nabla f_i(x^i_{t}) - \nabla f(x_{t}) \right]+ \nabla f(x_{t})\right\|^2\\
\leq \frac{KT\eta_l^2}{m}\sigma_l^2 + \frac{\eta_l^2\left(m-1\right)}{mn\left(n-1\right)} C_\beta \sum_{t=0}^{T-1}\left\| \sum_{i=1}^n\sum_{k=0}^{K-1}\nabla f_i(x^i_{t,k}) \right\|^2 +\\
\frac{\eta_l^2\left(n-m\right)}{m n\left(n-1\right)} C_\beta \sum_{t=0}^{T-1}\left( 15nK^3L^3\eta_l^2\left(\sigma_l^2+6K\sigma_g^2\right)+\left(90nK^4L^2\eta_l^2+3nK^2\right)\left\|\nabla f(x_t)\right\|^2 + 3nK^2\sigma_g^2 \right)
\end{gathered}\nonumber
\end{equation}

Merging $A_1$, $A_2$, $A_3$ together,

\begin{equation}
\begin{gathered}
\mathbb{E}\left[f(z_{t+1})\right] - f(z_t) \leq \underbrace{\mathbb{E}\left[\left\langle \sqrt{\eta_t} \left(\nabla f(z_t)-\nabla f(x_t)\right),-\sqrt{\eta_t} \Delta_t \right\rangle\right]}_{A_1} + \underbrace{\mathbb{E}\left[\left\langle \nabla f(x_t),-\eta_t \Delta_t \right\rangle\right]}_{A_2} + \underbrace{\frac{L}{2}\eta_t^2\mathbb{E}\left[\left\| \Delta_t\right\|^2\right]}_{A_3} \\
\le \frac{1}{2}\eta_t^3L^2\left(\frac{\beta_t\nu_t}{1-\beta_t}\right)^2\mathbb{E}\left[\left\| d_t\right\|^2\right] + \frac{1}{2}\eta_t\mathbb{E}\left[\left\|\Delta_t\right\|^2\right] + \frac{L}{2}\eta_t^2\mathbb{E}\left[\left\| \Delta_t\right\|^2\right]\\
-\frac{17}{64}\eta_t \eta_l K \mathbb{E}\left [\left\| \nabla f\left(x_t\right) \right\|^2\right]-\frac{\eta_t\eta_l}{2Kn^2}\mathbb{E}\left[\left\| \sum_{i=1}^n\sum_{k=0}^{K-1}\nabla f_i(x^i_{t,k})\right\|^2\right]+\frac{5}{2}\eta_t\eta_l^3L^2K^2\left(\sigma_l^2+6K\sigma_g^2\right)
\end{gathered}\nonumber
\end{equation}

Reorganizing terms, we have the following,

\begin{equation}
\begin{gathered}
\mathbb{E}\left [\left\| \nabla f\left(x_t\right) \right\|^2\right] \le  \frac{64}{17}\frac{f(z_t) - \mathbb{E}\left[f(z_{t+1})\right]}{\eta_t\eta_lK}  + \frac{32}{17\eta_l K}L^2W_1^2\mathbb{E}\left[\left\| d_t\right\|^2\right] + \left(\frac{32}{17\eta_lK}+ \frac{32 L}{17}\frac{\eta_t}{\eta_lK}\right)\mathbb{E}\left[\left\|\Delta_t\right\|^2\right] \\
-\frac{32}{17K^2n^2}\mathbb{E}\left[\left\| \sum_{i=1}^n\sum_{k=0}^{K-1}\nabla f_i(x^i_{t,k})\right\|^2\right]+\frac{160}{17} \eta_l^2L^2K \left(\sigma_l^2+6K\sigma_g^2\right)
\end{gathered}\nonumber
\end{equation}

Sum over all $S$ stages and take average, we get,

\begin{equation}
\begin{gathered}
\Bar{\mathcal{G}}\triangleq\frac{1}{S}\sum_{s=0}^{S-1} \frac{1}{T_s}\sum_{t=T_0+\dots+T_{s-1} }^{T_0+\dots+T_s-1} \mathbb{E}\left [\left\| \nabla f\left(x_t\right) \right\|^2\right]\\
\leq \frac{64}{17}\frac{f(z_0)-\mathbb{E}\left[f(z_{T})\right]}{S W_2 \eta_l K} + \frac{32}{17} \frac{L^2 W_1^2 \Bar{\eta}}{W_2\eta_l K}\sum_{t=0}^{T-1}\mathbb{E}\left[\left\| d_t\right\|^2\right] - \frac{32}{17}\frac{\eta_S}{S W_2 K^2 n^2}\sum_{t=0}^{T-1} \mathbb{E}\left[\left\| \sum_{i=1}^n\sum_{k=0}^{K-1} \nabla f_i(x_{t,k}^i)\right\|^2\right] \\
+ \frac{160}{17} \eta_l^2L^2K \left(\sigma_l^2+6K\sigma_g^2\right) + \left(\frac{32\Bar{\eta}}{17\eta_lK W_2} + \frac{32 L}{17}\frac{\hat{\eta}^2}{\eta_l  W_2 K}\right)\sum_{t=0}^{T-1}\mathbb{E}\left[\left\|\Delta_t\right\|^2\right]\\
\leq \frac{64}{17}\frac{f(z_0)-\mathbb{E}\left[f(z_{T})\right]}{S W_2 \eta_l K} + \sum_{t=0}^{T-1} \mathbb{E}\left[\left\| \sum_{i=1}^n\sum_{k=0}^{K-1} \nabla f_i(x_{t,k}^i)\right\|^2\right] \cdot\\
\left( - \frac{32}{17}\frac{\eta_S}{S W_2 K^2 n^2} + \frac{\eta_l^2 \left(m-1\right)}{nm\left(n-1\right)} \left(\frac{32 \Bar{\eta}}{17 \eta_l K W_2} + \frac{32 L}{17}\frac{\hat{\eta}^2}{\eta_l W_2 K}\right) + \frac{\eta_l^2\left(m-1\right)}{mn\left(n-1\right)} \frac{32}{17} \frac{L^2 W_1^2 \Bar{\eta}}{W_2 \eta_l K}\right)\\
+\left( \frac{32}{17} \frac{L^2 W_1^2 \Bar{\eta}}{W_2 \eta_l K}\frac{\eta_l^2\left(n-m\right)}{m n\left(n-1\right)}\left(90nK^4L^2\eta_l^2+3nK^2\right)\right)\cdot \sum_{t=0}^{T-1}\left\|\nabla f(x_t)\right\|^2 \\
+  \left(\left(\frac{32\Bar{\eta}}{17 \eta_l K W_2} + \frac{32 L}{17}\frac{\hat{\eta}^2}{\eta_l W_2 K}\right)\frac{\eta_l^2\left(n-m\right)}{n m \left( n - 1 \right)}\left(90nK^4L^2\eta_l^2 + 3nK^2\right) \right)
\cdot \sum_{t=0}^{T-1}\left\|\nabla f(x_t)\right\|^2\\
+ \left( \frac{\eta_l}{m}\Phi +\frac{15\left(n-m\right)K^2L^3\eta_l^3}{m\left(n-1\right)}\Phi + \frac{160}{17} \eta_l^2L^2K \right)\sigma_l^2\\
+ \left(  \frac{90\left(n-m\right)K^3L^3\eta_l^3}{m\left(n-1\right)}\Phi  +  \frac{3\eta_l\left(n-m\right)K}{m\left(n-1\right)}\Phi     + \frac{960}{17} \eta_l^2L^2K^2         \right)\sigma_g^2
\end{gathered}\nonumber
\end{equation}

where we denote $\Phi$ for ease of notation,

\begin{equation}
\begin{gathered}
\Phi \triangleq \frac{32 T \Bar{\eta} + 32 L T \hat{\eta}^2 + 32 L^2 W_1^2 T \Bar{\eta}}{17 W_2}
\end{gathered}\nonumber
\end{equation}

We can verify, when the following condition holds,

\begin{equation}
\begin{gathered}
\eta_l \leq \frac{1}{\left( C_\eta + L \Bar{\eta} C_\eta + L^2 W_1^2 C_\eta \right) S K}\frac{m\left(n-1\right)}{n\left(m-1\right)}
\end{gathered}\nonumber
\end{equation}

where $C_\eta=\frac{\eta_0}{\eta_S}$.

we have the coefficient for $\sum_{t=0}^{T-1} \mathbb{E}\left[\left\| \sum_{i=1}^n\sum_{k=0}^{K-1} \nabla f_i(x_{t,k}^i)\right\|^2\right]$,

\begin{equation}
\begin{gathered}
- \frac{32}{17}\frac{\eta_S}{S W_2 K^2 n^2} + \frac{\eta_l^2 \left(m-1\right)}{nm\left(n-1\right)} \left(\frac{32 \Bar{\eta}}{17 \eta_l K W_2} + \frac{32 L}{17}\frac{\hat{\eta}^2}{\eta_l W_2 K}\right) + \frac{\eta_l^2\left(m-1\right)}{mn\left(n-1\right)} \frac{32}{17} \frac{L^2 W_1^2 \Bar{\eta}}{W_2 \eta_l K} \leq 0
\end{gathered}\nonumber
\end{equation}

With the following inequality,

\begin{equation}
\begin{gathered}
\frac{1}{S W_2}\sum_{t=0}^{T-1}\left\|\nabla f(x_t)\right\|^2 = \frac{1}{S}\sum_{s=0}^{S-1} \frac{1}{T_s \eta_s}\sum_{t=T_0+\dots+T_{s-1} }^{T_0+\dots+T_s-1} \left\|\nabla f(x_t)\right\|^2 \leq \frac{1}{\eta_S} \frac{1}{S}\sum_{s=0}^{S-1} \frac{1}{T_s}\sum_{t=T_0+\dots+T_{s-1} }^{T_0+\dots+T_s-1} \left\|\nabla f(x_t)\right\|^2
\end{gathered}\nonumber
\end{equation}

We could verify, when the following condition holds,

\begin{equation}
\begin{gathered}
\eta_l \leq \frac{17}{282} \frac{m}{\left(C_\eta + L C_\eta \Bar{\eta} + L^2 W_1^2 C_\eta\right)S K}
\end{gathered}\nonumber
\end{equation}

We have the following,

\begin{equation}
\begin{gathered}
\left( \frac{32}{17} \frac{L^2 W_1^2 \Bar{\eta}}{W_2 \eta_l K}\frac{\eta_l^2\left(n-m\right)}{m n\left(n-1\right)}\left(90nK^4L^2\eta_l^2+3nK^2\right)\right)\cdot \sum_{t=0}^{T-1}\left\|\nabla f(x_t)\right\|^2 \\
+  \left(\left(\frac{32\Bar{\eta}}{17 \eta_l K W_2} + \frac{32 L}{17}\frac{\hat{\eta}^2}{\eta_l W_2 K}\right)\frac{\eta_l^2\left(n-m\right)}{n m \left( n - 1 \right)}\left(90nK^4L^2\eta_l^2 + 3nK^2\right) \right)
\cdot \sum_{t=0}^{T-1}\left\|\nabla f(x_t)\right\|^2 \\
\underset{(i)}{\leq}\frac{\eta_l^2\left(n-m\right)}{n m \left( n - 1 \right)} \frac{141 nK^2}{32} \left(\frac{32\Bar{\eta}}{17 \eta_l K W_2} + \frac{32 L}{17}\frac{\hat{\eta}^2}{\eta_l W_2 K} + \frac{32}{17} \frac{L^2 W_1^2 \Bar{\eta}}{W_2 \eta_l K}\right)\cdot \sum_{t=0}^{T-1}\left\|\nabla f(x_t)\right\|^2\\
\leq \frac{\eta_l^2\left(n-m\right)}{n m \left( n - 1 \right)} \frac{141 nK^2}{32} \frac{32 \eta_S}{17\eta_l K W_2} \left(C_\eta+LC_\eta\Bar{\eta}+L^2W_1^2 C_\eta \right)\cdot \sum_{t=0}^{T-1}\left\|\nabla f(x_t)\right\|^2\\
\underset{(ii)}{\leq} \frac{1}{2} \cdot \frac{1}{S}\sum_{s=0}^{S-1} \frac{1}{T_s}\sum_{t=T_0+\dots+T_{s-1} }^{T_0+\dots+T_s-1} \left\|\nabla f(x_t)\right\|^2
\end{gathered}\nonumber
\end{equation}

where $(i)$ holds as $90nK^4L^2\eta_l^2+3nK^2\leq\frac{141}{32}nK^2$ when $\eta_l\leq\frac{1}{8KL}$, $(ii)$ holds by plugging in the learning rate constraint $\eta_l \leq \frac{17}{282} \frac{m}{\left(C_\eta + L C_\eta \Bar{\eta} + L^2 W_1^2 C_\eta\right)S K}$.

Merging everything together, we have the following,

\begin{equation}
\begin{gathered}
\Bar{\mathcal{G}}\triangleq\frac{1}{S}\sum_{s=0}^{S-1} \frac{1}{T_s}\sum_{t=T_0+\dots+T_{s-1} }^{T_0+\dots+T_s-1} \left\|\nabla f(x_t)\right\|^2\\
\leq \frac{64}{17}\frac{f(z_0)-\mathbb{E}\left[f(z_{T})\right]}{S W_2 \eta_l K}  
+ \left( \frac{\eta_l}{m}\Phi +\frac{15\left(n-m\right)K^2L^3\eta_l^3}{m\left(n-1\right)}\Phi + \frac{160}{17} \eta_l^2L^2K \right)\sigma_l^2\\
+ \left(  \frac{90\left(n-m\right)K^3L^3\eta_l^3}{m\left(n-1\right)}\Phi  +  \frac{3\eta_l\left(n-m\right)K}{m\left(n-1\right)}\Phi     + \frac{960}{17} \eta_l^2L^2K^2         \right)\sigma_g^2
\end{gathered}\nonumber
\end{equation}

Suppose $S=1$, i.e. the typical constant hyperparameter regime, the total number of rounds are $T$, $\Bar{\eta}=\eta_0=\Theta\left(\sqrt{m K}\right)$ and $\eta_l=\Theta\left(\frac{1}{\sqrt{T}K}\right)$, $W_2=\Theta\left(T\sqrt{m K}\right)$ in this case. Assume $W_1^2=\mathcal{O}\left(\sqrt{mK}\right)$, recall $\Phi \triangleq \frac{32 T \Bar{\eta} + 32 L T \hat{\eta}^2 + 32 L^2 W_1^2 T \Bar{\eta}}{17 W_2}$, we could verify $\Phi = \Theta\left(\sqrt{m K}\right)$.

We have the bound as,

\begin{equation}
\begin{gathered}
\Bar{\mathcal{G}}\triangleq\frac{1}{S}\sum_{s=0}^{S-1} \frac{1}{T_s}\sum_{t=T_0+\dots+T_{s-1} }^{T_0+\dots+T_s-1} \mathbb{E}\left[\left\|\nabla f(x_t)\right\|^2\right]\\
\leq  \mathcal{O}\left(\frac{1}{\sqrt{T K m}}\right) \left(f(x_0)-f^\ast\right) + \left( \mathcal{O}\left(\frac{1}{\sqrt{T K m}}\right) + \mathcal{O}\left(\frac{1}{\sqrt{T^3 K m}}\right) + \mathcal{O}\left(\frac{1}{TK}\right) \right)  \sigma_l^2\\
+ \left( \mathcal{O}\left( \sqrt{\frac{K}{T^3 m}}  \right) + \mathcal{O}\left( \sqrt{\frac{K}{T m}}  \right) + \mathcal{O}\left(\frac{1}{T}\right) \right) \sigma_g^2
\end{gathered}\nonumber
\end{equation}

Only keeping the dominant terms,

\begin{equation}
\begin{gathered}
\Bar{\mathcal{G}}\triangleq\frac{1}{S}\sum_{s=0}^{S-1} \frac{1}{T_s}\sum_{t=T_0+\dots+T_{s-1} }^{T_0+\dots+T_s-1} \mathbb{E}\left[\left\|\nabla f(x_t)\right\|^2\right]\\
\leq  \mathcal{O}\left(\frac{1}{\sqrt{T K m}}\right) \left(f(x_0)-f^\ast\right) +   \mathcal{O}\left(\frac{1}{\sqrt{T K m}}\right)     \sigma_l^2 
+  \mathcal{O}\left( \sqrt{\frac{K}{T m}}  \right)   \sigma_g^2
\end{gathered}\nonumber
\end{equation}

Suppose $S>1$ but $S=\Theta(1)$, i.e. the multistage regime, the total number of rounds are $T$, $\Bar{\eta} = \Theta\left(\sqrt{mK}\right)$, and $\hat{\eta}^2 = \Theta\left(mK\right)$, $\eta_l=\Theta\left(\frac{1}{\sqrt{T}K}\right)$, $W_2=\Theta\left(\frac{T\sqrt{mK}}{S}\right)$, i.e. $T\Bar{\eta}$ is equally divided into $S$ stages. Assume $W_1^2=\mathcal{O}\left(\sqrt{mK}\right)$, we could verify $\Phi=\Theta(\sqrt{mK})$. Thus, we have the bound,

\begin{equation}
\begin{gathered}
\Bar{\mathcal{G}}\triangleq\frac{1}{S}\sum_{s=0}^{S-1} \frac{1}{T_s}\sum_{t=T_0+\dots+T_{s-1} }^{T_0+\dots+T_s-1} \mathbb{E}\left[\left\|\nabla f(x_t)\right\|^2\right]\\
\leq  \mathcal{O}\left(\frac{1}{\sqrt{T K m}}\right) \left(f(x_0)-f^\ast\right) +   \mathcal{O}\left(\frac{1}{\sqrt{T K m}}\right)     \sigma_l^2 
+  \mathcal{O}\left( \sqrt{\frac{K}{T m}}  \right)   \sigma_g^2
\end{gathered}\nonumber
\end{equation}

In both cases, the dominant term is $\mathcal{O}\left( \sqrt{\frac{K}{T m}}  \right)$.

\end{proof}

\section{Proof of Theorem \ref{multistage_fedgm_free_uniform_arrival_convergence_theorem} and Corollary \ref{corollary:free_multistage_fedgm_rate}}
\label{sec:proof_free_multistage_fedgm_uniform_arrival}

\begin{proof}[Proof of Autonomous Multistage GM with Uniform Arrival]
We introduce a Lyapunov sequence $\{z_t\}_{t=0}^{T-1}$ which is devised as follows:

\begin{equation}
\label{auxiliary_seq}
z_t= x_t-\frac{\eta_t\beta_t\nu_t}{1-\beta_t}d_{t}
\end{equation}
where $d_0=0$.

We could easily verify $z_{t+1}-z_t= -\eta_t \Delta_t$. Let $-\eta y_t=x_{t+1}-x_t$, we first bound $\mathbb{E}\left[\left\| \Delta_t\right\|^2\right]$, $\sum_{t=0}^{T-1}\mathbb{E}\left[\left\| d_t\right\|^2\right]$, and $\sum_{t=0}^{T-1}\mathbb{E}\left[\left\| y_t\right\|^2\right]$.

\textbf{Bounding} $\mathbb{E}\left[\left\| \Delta_t\right\|^2\right]$:

\begin{equation}
\begin{gathered}
\mathbb{E}\left[\left\| \Delta_t\right\|^2\right] = \mathbb{E}\left[\left\| \frac{1}{m}\sum_{i\in\mathcal{S}_t}\Delta^i_{t-\tau_{t,i}} \right\|^2\right]\\
\underset{(i)}{=}\mathbb{E}\left[\left\| \frac{1}{m}\sum_{i\in\mathcal{S}_t}  \frac{\eta_l}{K_{t,i}} \sum_{k=0}^{K_{t,i}-1} g_{t-\tau_{t,i},k}^i \right\|^2\right] \\
= \mathbb{E}\left[\left\| \frac{1}{m}\sum_{i\in\mathcal{S}_t} \frac{\eta_l}{K_{t,i}} \sum_{k=0}^{K_{t,i}-1} \left( g_{t-\tau_{t,i},k}^i - \nabla f_i(x_{t-\tau_{t,i},k}^i) + \nabla f_i(x_{t-\tau_{t,i},k}^i)\right) \right\|^2\right] \\
\underset{(ii)}{=} \mathbb{E}\left[\left\| \frac{1}{m}\sum_{i\in\mathcal{S}_t} \frac{\eta_l}{K_{t,i}} \sum_{k=0}^{K_{t,i}-1} \left\{ g_{t-\tau_{t,i},k}^i - \nabla f_i(x_{t-\tau_{t,i},k}^i) \right\}  \right\|^2\right] + \mathbb{E}\left[\left\| \frac{1}{m}\sum_{i\in\mathcal{S}_t} \frac{\eta_l}{K_{t,i}} \sum_{k=0}^{K_{t,i}-1}  \nabla f_i(x_{t-\tau_{t,i},k}^i) \right\|^2\right]\\
\underset{(iii)}{\leq} \frac{1}{m^2}\sum_{i\in\mathcal{S}_t}\frac{\eta_l^2}{K^2_{t,i}}\sum_{k=0}^{K_{t,i}-1}\sigma_l^2 + \mathbb{E}\left[\left\| \frac{1}{m}\sum_{i\in\mathcal{S}_t} \frac{\eta_l}{K_{t,i}} \sum_{k=0}^{K_{t,i}-1}  \nabla f_i(x_{t-\tau_{t,i},k}^i) \right\|^2\right]\\
\leq \frac{\eta_l^2}{m}\frac{1}{K_t}\sigma^2_l + \frac{\eta_l^2}{m^2}  \mathbb{E}\left[\left\| \sum_{i=1}^n\mathbf{1}\left\{i\in\mathcal{S}_t\right\} \frac{1}{K_{t,i}} \sum_{k=0}^{K_{t,i}-1}  \nabla f_i(x_{t-\tau_{t,i},k}^i) \right\|^2\right]
\end{gathered}\nonumber
\end{equation}
where $\frac{1}{K_t}=\frac{1}{m}\sum_{i\in\mathcal{S}_t}\frac{1}{K_{t,i}}$. $(i)$ follows from the definition of $\Delta^i_{t-\tau_{t,i}}$, $(ii)$ and $(iii)$ hold as $\mathbb{E}\left[\left\|\sum_{i=1}^n x_i\right\|^2\right] = \sum_{i=1}^n \mathbb{E}\left[\left\| x_i \right\|^2\right]$ when $\mathbb{E}\left[ x_i \right]=0$, and we know $\mathbb{E}\left[g_{t-\tau_{t,i},k}^i - \nabla f_i(x_{t-\tau_{t,i},k}^i)\right]=0$.

\textbf{Bounding} $\sum_{t=0}^{T-1}\mathbb{E}\left[\left\| d_t\right\|^2\right]$:

We could verify:
\begin{equation}
\begin{gathered}
 d_t = \sum_{p=0}^t a_{t,p}\Delta_p,         \quad  \text{where} \quad  a_{t,p}=\left(1-\beta_p\right)\prod_{q=p+1}^t\beta_q
\end{gathered}\nonumber
\end{equation}
We further get,
\begin{equation}
\begin{gathered}
\mathbb{E}\left[\left\| d_t\right\|^2\right]=\mathbb{E}\left[\left\| \sum_{p=0}^t a_{t,p}\Delta_p\right\|^2\right]\\
\leq \sum_{e=1}^d \mathbb{E}\left[\sum_{p=0}^t a_{t,p}\Delta_{p,e}\right]^2 
\leq \sum_{e=1}^d \mathbb{E}\left[ \left(\sum_{p=0}^t a_{t,p}\right) \left(\sum_{p=0}^t a_{t,p}\Delta_{p,e}^2\right) \right] \leq \left(1- \prod_{q=0}^t \beta_q\right)\sum_{p=0}^t a_{t,p}\mathbb{E}\left[\left\| \Delta_p \right\|^2\right]\\
\le \left(1- \prod_{q=0}^t \beta_q\right)\sum_{p=0}^t a_{t,p}\left\{ \frac{\eta_l^2}{m}\frac{1}{K_t}\sigma^2_l + \frac{\eta_l^2}{m^2}  \mathbb{E}\left[\left\| \sum_{i=1}^n\mathbf{1}\left\{i\in\mathcal{S}_p\right\} \frac{1}{K_{p,i}} \sum_{k=0}^{K_{p,i}-1}  \nabla f_i(x_{p-\tau_{p,i},k}^i) \right\|^2\right] \right\}\\
\leq \frac{\eta_l^2}{m}\frac{1}{K_t}\sigma^2_l + \frac{\eta_l^2}{m^2} \sum_{p=0}^t a_{t,p} \cdot \mathbb{E}\left[\left\| \sum_{i=1}^n\mathbf{1}\left\{i\in\mathcal{S}_p\right\} \frac{1}{K_{p,i}} \sum_{k=0}^{K_{p,i}-1}  \nabla f_i(x_{p-\tau_{p,i},k}^i) \right\|^2\right] 
\end{gathered}\nonumber
\end{equation}

Summing over $t\in\{0,1,\dots,T-1\}$,
\begin{equation}
\begin{gathered}
\sum_{t=0}^{T-1}\mathbb{E}\left[\left\| d_t\right\|^2\right] \leq
\frac{\eta_l^2}{m}\sum_{t=0}^{T-1}\frac{1}{K_t}\sigma^2_l + \frac{\eta_l^2}{m^2}\sum_{t=0}^{T-1} \sum_{p=0}^t a_{t,p} \cdot \mathbb{E}\left[\left\| \sum_{i=1}^n\mathbf{1}\left\{i\in\mathcal{S}_p\right\} \frac{1}{K_{p,i}} \sum_{k=0}^{K_{p,i}-1}  \nabla f_i(x_{p-\tau_{p,i},k}^i) \right\|^2\right] \\
\leq \frac{\eta_l^2}{m}\sum_{t=0}^{T-1}\frac{1}{K_t}\sigma^2_l + \frac{\eta_l^2}{m^2} \sum_{p=0}^{T-1}\left(\sum_{t=p}^{T-1}a_{t,p}\right) \mathbb{E}\left[\left\| \sum_{i=1}^n\mathbf{1}\left\{i\in\mathcal{S}_p\right\} \frac{1}{K_{p,i}} \sum_{k=0}^{K_{p,i}-1}  \nabla f_i(x_{p-\tau_{p,i},k}^i) \right\|^2\right]\\
\leq \frac{\eta_l^2}{m}\sum_{t=0}^{T-1}\frac{1}{K_t}\sigma^2_l + \frac{\eta_l^2}{m^2} C_\beta \sum_{t=0}^{T-1} \mathbb{E}\left[\left\| \sum_{i=1}^n\mathbf{1}\left\{i\in\mathcal{S}_t\right\} \frac{1}{K_{t,i}} \sum_{k=0}^{K_{t,i}-1}  \nabla f_i(x_{t-\tau_{t,i},k}^i) \right\|^2\right]\\
\end{gathered}\nonumber
\end{equation}

\textbf{Bounding} $\sum_{t=0}^{T-1}\mathbb{E}\left[\left\| y_t\right\|^2\right]$:

We could verify:
\begin{equation}
\begin{gathered}
 y_t = \sum_{p=0}^t b_{t,p}\Delta_p,  
\end{gathered}\nonumber
\end{equation}
where $b_{t,p}$ is defined as follows,
\begin{equation}
\begin{gathered}
b_{t,p}= \begin{cases} 
    1-\beta_t\nu_t  & p = t \\
    \nu_t(1-\beta_p)\prod_{q=p+1}^t\beta_q   & p < t 
   \end{cases}
\end{gathered}\nonumber   
\end{equation}

We further get,
\begin{equation}
\begin{gathered}
\mathbb{E}\left[\left\| y_t\right\|^2\right]=\mathbb{E}\left[\left\| \sum_{p=0}^t b_{t,p}\Delta_p\right\|^2\right]\\
\leq \sum_{e=1}^d \mathbb{E}\left[\sum_{p=0}^t b_{t,p}\Delta_{p,e}\right]^2 
\leq \sum_{e=1}^d \mathbb{E}\left[ \left(\sum_{p=0}^t b_{t,p}\right) \left(\sum_{p=0}^t b_{t,p}\Delta_{p,e}^2\right) \right] \leq \left(1- \nu_t\prod_{q=0}^t \beta_q\right)\sum_{p=0}^t b_{t,p}\mathbb{E}\left[\left\| \Delta_p \right\|^2\right]\\
\le \left(1- \nu_t \prod_{q=0}^t \beta_q\right)\sum_{p=0}^t b_{t,p}\left\{ \frac{\eta_l^2}{m}\frac{1}{K_t}\sigma^2_l + \frac{\eta_l^2}{m^2}  \mathbb{E}\left[\left\| \sum_{i=1}^n\mathbf{1}\left\{i\in\mathcal{S}_p\right\} \frac{1}{K_{p,i}} \sum_{k=0}^{K_{p,i}-1}  \nabla f_i(x_{p-\tau_{p,i},k}^i) \right\|^2\right] \right\}\\
\leq \frac{\eta_l^2}{m}\frac{1}{K_t}\sigma^2_l + \frac{\eta_l^2}{m^2} \sum_{p=0}^t b_{t,p} \cdot \mathbb{E}\left[\left\| \sum_{i=1}^n\mathbf{1}\left\{i\in\mathcal{S}_p\right\} \frac{1}{K_{p,i}} \sum_{k=0}^{K_{p,i}-1}  \nabla f_i(x_{p-\tau_{p,i},k}^i) \right\|^2\right] 
\end{gathered}\nonumber
\end{equation}

Summing over $t\in\{0,1,\dots,T-1\}$,
\begin{equation}
\begin{gathered}
\sum_{t=0}^{T-1}\mathbb{E}\left[\left\| y_t\right\|^2\right] \leq
\frac{\eta_l^2}{m}\sum_{t=0}^{T-1}\frac{1}{K_t}\sigma^2_l + \frac{\eta_l^2}{m^2}\sum_{t=0}^{T-1} \sum_{p=0}^t b_{t,p} \cdot \mathbb{E}\left[\left\| \sum_{i=1}^n\mathbf{1}\left\{i\in\mathcal{S}_p\right\} \frac{1}{K_{p,i}} \sum_{k=0}^{K_{p,i}-1}  \nabla f_i(x_{p-\tau_{p,i},k}^i) \right\|^2\right] \\
\leq \frac{\eta_l^2}{m}\sum_{t=0}^{T-1}\frac{1}{K_t}\sigma^2_l + \frac{\eta_l^2}{m^2} \sum_{p=0}^{T-1}\left(\sum_{t=p}^{T-1}b_{t,p}\right) \mathbb{E}\left[\left\| \sum_{i=1}^n\mathbf{1}\left\{i\in\mathcal{S}_p\right\} \frac{1}{K_{p,i}} \sum_{k=0}^{K_{p,i}-1}  \nabla f_i(x_{p-\tau_{p,i},k}^i) \right\|^2\right]\\
\leq \frac{\eta_l^2}{m}\sum_{t=0}^{T-1}\frac{1}{K_t}\sigma^2_l + \frac{\eta_l^2}{m^2} C_\beta \sum_{t=0}^{T-1} \mathbb{E}\left[\left\| \sum_{i=1}^n\mathbf{1}\left\{i\in\mathcal{S}_t\right\} \frac{1}{K_{t,i}} \sum_{k=0}^{K_{t,i}-1}  \nabla f_i(x_{t-\tau_{t,i},k}^i) \right\|^2\right]\\
\end{gathered}\nonumber
\end{equation}

Since $f$ is $L$-smooth, taking conditional expectation with respect to all randomness prior to step $t$, we have

\begin{equation}
\begin{gathered}
\mathbb{E}\left[f(z_{t+1})\right]\leq
f(z_t)+\mathbb{E}\left[\left\langle \nabla f(z_t),z_{t+1}-z_t \right\rangle\right]+\frac{L}{2}\mathbb{E}\left[\left\| z_{t+1}-z_t\right\|^2\right]\\
\leq f(z_t)+\mathbb{E}\left[\left\langle \nabla f(z_t),-\eta_t  \Delta_t \right\rangle\right]+\frac{L}{2}\eta_t^2\mathbb{E}\left[\left\| \Delta_t\right\|^2\right]\\
\leq f(z_t)+ \underbrace{\mathbb{E}\left[\left\langle \sqrt{\eta_t} \left(\nabla f(z_t)-\nabla f(x_t)\right),-\sqrt{\eta_t } \Delta_t \right\rangle\right]}_{A_1} + \underbrace{\mathbb{E}\left[\left\langle \nabla f(x_t),-\eta_t \Delta_t \right\rangle\right]}_{A_2} + \underbrace{\frac{L}{2}\eta_t^2\mathbb{E}\left[\left\| \Delta_t\right\|^2\right]}_{A_3} \\
\end{gathered}\nonumber
\end{equation}

\textbf{Bounding} $A_1$:
\begin{equation}
\begin{gathered}
A_1 =\mathbb{E}\left[\left\langle \sqrt{\eta_t} \left(\nabla f(z_t)-\nabla f(x_t)\right),-\sqrt{\eta_t} \Delta_t \right\rangle\right]\\
\underset{(i)}{\leq}\mathbb{E}\left[\left\|\sqrt{\eta_t} \left(\nabla f(z_t)-\nabla f(x_t)\right)\right\| \cdot \left\|-\sqrt{\eta_t} \Delta_t\right\|\right]\\
\underset{(ii)}{\leq}\frac{1}{2}\eta_t^3 L^2\left(\frac{\beta_t\nu_t}{1-\beta_t}\right)^2\mathbb{E}\left[\left\| d_t\right\|^2\right] + \frac{1}{2}\eta_t\mathbb{E}\left[\left\|\Delta_t\right\|^2\right]
\end{gathered}\nonumber
\end{equation}

where $(i)$ holds by applying Cauchy-Schwarz inequality, and $(ii)$ follows from Young’s inequality and $f$ is $L$-smooth.

\textbf{Bounding} $A_2$:
\begin{equation}
\begin{gathered}
A_2=\mathbb{E}\left[\left\langle \nabla f(x_t),-\eta_t \Delta_t \right\rangle\right]\\
=\eta_t\mathbb{E}\left[\left\langle \nabla f\left(x_t\right),\eta_l  \nabla f\left(x_t\right) - \Delta_t - \eta_l  \nabla f\left(x_t\right)  \right\rangle\right]\\
=-\eta_t \eta_l   \mathbb{E}\left [\left\| \nabla f\left(x_t\right) \right\|^2\right]+\eta_t\mathbb{E}\left[\left\langle \nabla f\left(x_t\right), \eta_l   \nabla f\left(x_t\right) - \Delta_t \right\rangle\right]
\end{gathered}\nonumber
\end{equation}
where we further bound $\eta_t\mathbb{E}\left[\langle \nabla f\left(x_t\right), \eta_l   \nabla f\left(x_t\right) - \Delta_t \rangle\right]$,

\begin{equation}
\begin{gathered}
\eta_t\mathbb{E}\left[\langle \nabla f\left(x_t\right), \eta_l   \nabla f\left(x_t\right) - \Delta_t \rangle\right]
=\eta_t \mathbb{E}\left[\left\langle \sqrt{\eta_l} \nabla f(x_t),   \frac{\sqrt{\eta_l}}{m}\sum_{i\in\mathcal{S}_t} \frac{1}{K_{t,i}} \sum_{k=0}^{K_{t,i}-1}\left(\nabla f(x_t) -  g_{t-\tau_{t,i},k}^i\right) \right\rangle\right]\\
\underset{(i)}{=}\eta_t \mathbb{E}\left[\left\langle \sqrt{\eta_l} \nabla f(x_t),   \frac{\sqrt{\eta_l}}{m}\sum_{i\in\mathcal{S}_t} \frac{1}{K_{t,i}} \sum_{k=0}^{K_{t,i}-1}\left(\nabla f(x_t) -  \nabla f_i(x_{t-\tau_{t,i},k}^i)\right) \right\rangle\right]\\
\underset{(ii)}{=}\eta_t \mathbb{E}\left[\left\langle \sqrt{\eta_l} \nabla f(x_t),   \frac{\sqrt{\eta_l}}{n}\sum_{i=1}^n \frac{1}{K_{t,i}} \sum_{k=0}^{K_{t,i}-1}\left(\nabla f_i(x_t) -  \nabla f_i(x_{t-\tau_{t,i},k}^i)\right) \right\rangle\right]\\
\underset{(iii)}{=} \frac{\eta_t\eta_l}{2} \mathbb{E}\left[\left\| \nabla f(x_t) \right\|^2\right] - \frac{\eta_t\eta_l}{2} \mathbb{E}\left[\left\| \frac{1}{n}\sum_{i=1}^n \frac{1}{K_{t,i}} \sum_{k=0}^{K_{t,i}-1} \nabla f_i(x_{t-\tau_{t,i},k}^i) \right\|^2\right] \\
+ \frac{\eta_t\eta_l}{2} \mathbb{E}\left[\left\| \nabla f(x_t) - \frac{1}{n}\sum_{i=1}^n \frac{1}{K_{t,i}} \sum_{k=0}^{K_{t,i}-1}  \nabla f_i(x_{t-\tau_{t,i},k}^i) \right\|^2\right]
\end{gathered}\nonumber
\end{equation}

where $(i)$ holds as we take conditional expectation with respect to all randomness prior to step $t$. $(ii)$ holds due to the following equality and the definition of $\nabla f(x_t)=\frac{1}{n}\sum_{i=1}^n \nabla f_i(x_t)$. $(iii)$ holds as $\left\langle a, b \right\rangle = \frac{1}{2} \left\| a \right\|^2 + \frac{1}{2} \left\| b \right\|^2 - \frac{1}{2} \left\| a - b \right\|^2 $.

\begin{equation}
\begin{gathered}
\mathbb{E}\left[\frac{1}{m} \sum_{i\in\mathcal{S}_t} \frac{1}{K_{t,i}} \sum_{k=0}^{K_{t,i}-1}\nabla f_i(x_{t-\tau_{t,i},k}^i)\right] = \mathbb{E}\left[ \frac{1}{m}\sum_{i=1}^n \mathbf{1}\left(i\in\mathcal{S}_t\right) \frac{1}{K_{t,i}} \sum_{k=0}^{K_{t,i}-1}\nabla f_i(x_{t-\tau_{t,i},k}^i)  \right] \\
= \frac{1}{m} \mathbb{E}\left[ \sum_{i=1}^n \mathbb{P}\left\{i\in\mathcal{S}_t\right\} \frac{1}{K_{t,i}} \sum_{k=0}^{K_{t,i}-1}\nabla f_i(x_{t-\tau_{t,i},k}^i) \right] \underset{(i)}{=} \frac{1}{n}\sum_{i=1}^n \mathbb{E}\left[ \frac{1}{K_{t,i}} \sum_{k=0}^{K_{t,i}-1}\nabla f_i(x_{t-\tau_{t,i},k}^i) \right]
\end{gathered}\nonumber
\end{equation}

where $(i)$ holds due to uniform arrival assumption, in which $\mathbb{P}\left\{i\in\mathcal{S}_t\right\}=\frac{m}{n}$.

We further have,

\begin{equation}
\begin{gathered}
\frac{\eta_t\eta_l}{2} \mathbb{E}\left[\left\| \nabla f(x_t) - \frac{1}{n}\sum_{i=1}^n \frac{1}{K_{t,i}} \sum_{k=0}^{K_{t,i}-1}  \nabla f_i(x_{t-\tau_{t,i},k}^i) \right\|^2\right] \\
= \frac{\eta_t\eta_l}{2} \mathbb{E}\left[\left\| \frac{1}{n}\sum_{i=1}^n \frac{1}{K_{t,i}} \sum_{k=0}^{K_{t,i}-1} \left( \nabla f_i(x_t)  -\nabla f_i(x_{t-\tau_{t,i},k}^i) \right) \right\|^2\right]\\
= \frac{\eta_t\eta_l}{2} \mathbb{E}\left[\left\| \frac{1}{n}\sum_{i=1}^n \frac{1}{K_{t,i}} \sum_{k=0}^{K_{t,i}-1} \left( \nabla f_i(x_t)  -\nabla f_i(x_{t-\tau_{t,i}}) + \nabla f_i(x_{t-\tau_{t,i}}) - \nabla f_i(x_{t-\tau_{t,i},k}^i) \right) \right\|^2\right]\\
\underset{(i)}{\leq} \eta_t\eta_l \mathbb{E}\left[\left\| \frac{1}{n}\sum_{i=1}^n \left( \nabla f_i(x_t)  -\nabla f_i(x_{t-\tau_{t,i}}) \right) \right\|^2\right] + \eta_t\eta_l \mathbb{E}\left[\left\| \frac{1}{n}\sum_{i=1}^n \frac{1}{K_{t,i}} \sum_{k=0}^{K_{t,i}-1} \left( \nabla f_i(x_{t-\tau_{t,i}}) - \nabla f_i(x_{t-\tau_{t,i},k}^i) \right) \right\|^2\right]\\
\underset{(ii)}{\leq}  \frac{\eta_t\eta_l}{n}\sum_{i=1}^n \mathbb{E}\left[\left\|  \nabla f_i(x_t)  -\nabla f_i(x_{t-\tau_{t,i}}) \right\|^2\right] + \frac{\eta_t\eta_l}{n}\sum_{i=1}^n \frac{1}{K_{t,i}} \sum_{k=0}^{K_{t,i}-1} \mathbb{E}\left[\left\|   \nabla f_i(x_{t-\tau_{t,i}}) - \nabla f_i(x_{t-\tau_{t,i},k}^i)  \right\|^2\right]\\
\underset{(iii)}{\leq}  \frac{\eta_t\eta_l L^2}{n}\sum_{i=1}^n \mathbb{E}\left[\left\|  x_t  - x_{t-\tau_{t,i}} \right\|^2\right] + \frac{\eta_t\eta_l L^2}{n}\sum_{i=1}^n \frac{1}{K_{t,i}} \sum_{k=0}^{K_{t,i}-1} \mathbb{E}\left[\left\|    x_{t-\tau_{t,i}}  -  x_{t-\tau_{t,i},k}^i \right\|^2\right]
\end{gathered}\nonumber
\end{equation}

where $(i)$ and $(ii)$ hold as $\left\|\sum_{i=1}^n x_i\right\|^2 \leq n \sum_{i=1}^n\left\| x_i \right\|^2$, $(iii)$ holds as $f_i$ is $L$-smooth.

Thus, we have, 

\begin{equation}
\begin{gathered}
A_2
\leq - \frac{\eta_t\eta_l}{2} \mathbb{E}\left[\left\| \nabla f(x_t) \right\|^2\right] - \frac{\eta_t\eta_l}{2} \mathbb{E}\left[\left\| \frac{1}{n}\sum_{i=1}^n \frac{1}{K_{t,i}} \sum_{k=0}^{K_{t,i}-1} \nabla f_i(x_{t-\tau_{t,i},k}^i) \right\|^2\right] \\
+  \frac{\eta_t\eta_l L^2}{n}\sum_{i=1}^n \mathbb{E}\left[\left\|  x_t  - x_{t-\tau_{t,i}} \right\|^2\right] + \frac{\eta_t\eta_l L^2}{n}\sum_{i=1}^n \frac{1}{K_{t,i}} \sum_{k=0}^{K_{t,i}-1} \mathbb{E}\left[\left\|    x_{t-\tau_{t,i}}  -  x_{t-\tau_{t,i},k}^i \right\|^2\right]
\end{gathered}\nonumber
\end{equation}

When $\eta_l\leq \frac{1}{8K_{t,i}L}$, we have,
\begin{equation}
\begin{gathered}
\mathbb{E}\left[\left\| x_{t-\tau_{t,i}}-x_{t-\tau_{t,i},k}^i  \right\|^2\right] \leq 5 K_{t,i}\eta_l^2\left(\sigma_l^2+6K_{t,i}\sigma_g^2\right)+30K_{t,i}^2\eta_l^2 \mathbb{E}\left[\left\| \nabla f(x_{t-\tau_{t,i}})\right\|^2\right]
\end{gathered}\nonumber
\end{equation}

We can further bound $\frac{1}{n}\sum_{i=1}^n \mathbb{E}\left[\left\| x_t - x_{t-\tau_{t,i}} \right\|^2\right]$
\begin{equation}
\begin{gathered}
\frac{1}{n}\sum_{i=1}^n \mathbb{E}\left[\left\| x_t - x_{t-\tau_{t,i}} \right\|^2\right] 
\underset{(i)}{\leq}  \mathbb{E}\left[\left\| x_t - x_{t-\tau_{t,u}} \right\|^2\right] = \mathbb{E}\left[\left\| \sum_{k=t-\tau_{t,u}}^{t-1} \left(x_{k+1}-x_k\right) \right\|^2\right] \\
\underset{(ii)}{\leq} \mathbb{E}\left[\left\| \sum_{k=t-\tau_{t,u}}^{t-1} \eta_k y_k \right\|^2\right] \underset{(iii)}{\leq} \tau \eta_0^2 \sum_{k=t-\tau_{t,u}}^{t-1}\mathbb{E}\left[\left\|  y_k \right\|^2\right]
\end{gathered}\nonumber
\end{equation}

where $(i)$ holds as we define $u=\argmax_{i\in\{1,2,\dots,n\}}\mathbb{E}\left[\left\| x_t - x_{t-\tau_{t,i}} \right\|^2\right] $, $(ii)$ follows from the definition of $y_k$, $(iii)$ holds as bounded maximum delay assumption, i.e. $\tau_{t,i}\leq \tau$ for any $t$ and $i$, and learning rate is decaying, i.e. $\eta_t\leq\eta_0$.

Merging all pieces together,

\begin{equation}
\begin{gathered}
A_2 \leq - \frac{\eta_t\eta_l}{2} \mathbb{E}\left[\left\| \nabla f(x_t) \right\|^2\right] - \frac{\eta_t\eta_l}{2} \mathbb{E}\left[\left\| \frac{1}{n}\sum_{i=1}^n \frac{1}{K_{t,i}} \sum_{k=0}^{K_{t,i}-1} \nabla f_i(x_{t-\tau_{t,i},k}^i) \right\|^2\right] \\
+  \eta_t\eta_l L^2 \tau \eta_0^2 \sum_{k=t-\tau_{t,u}}^{t-1}\mathbb{E}\left[\left\|  y_k \right\|^2\right]  + \frac{\eta_t\eta_l L^2}{n}\sum_{i=1}^n \left\{ 5 K_{t,i}\eta_l^2\left(\sigma_l^2+6K_{t,i}\sigma_g^2\right)+30K_{t,i}^2\eta_l^2 \mathbb{E}\left[\left\| \nabla f(x_{t-\tau_{t,i}})\right\|^2\right]\right\}\\
=- \frac{\eta_t\eta_l}{2} \mathbb{E}\left[\left\| \nabla f(x_t) \right\|^2\right] - \frac{\eta_t\eta_l}{2} \mathbb{E}\left[\left\| \frac{1}{n}\sum_{i=1}^n \frac{1}{K_{t,i}} \sum_{k=0}^{K_{t,i}-1} \nabla f_i(x_{t-\tau_{t,i},k}^i) \right\|^2\right]+  \eta_t\eta_l L^2 \tau \eta_0^2 \sum_{k=t-\tau_{t,u}}^{t-1}\mathbb{E}\left[\left\|  y_k \right\|^2\right] \\
+ \frac{30 \eta_t\eta_l^3 L^2}{n}\sum_{i=1}^n K_{t,i}^2\mathbb{E}\left[\left\| \nabla f(x_{t-\tau_{t,i}})\right\|^2\right] + 5 \Bar{K}_t \eta_t\eta^3_l L^2\sigma_l^2 + 30 \hat{K}_t^2 \eta_t \eta^3_l L^2 \sigma_g^2
\end{gathered}\nonumber
\end{equation}
where $\bar{K}_t\triangleq \frac{1}{n}\sum_{i=1}^n K_{t,i}$ and $\hat{K}_t^2 \triangleq \frac{1}{n}\sum_{i=1}^n K^2_{t,i}$.

Plug all pieces back in $\mathbb{E}\left[f(z_{t+1})\right] \leq f(z_t) + A_1 + A_2 + A_3$,

\begin{equation}
\begin{gathered}
\mathbb{E}\left[f(z_{t+1})\right] - f(z_t) \leq 
- \frac{\eta_t\eta_l}{2} \mathbb{E}\left[\left\| \nabla f(x_t) \right\|^2\right] - \frac{\eta_t\eta_l}{2} \mathbb{E}\left[\left\| \frac{1}{n}\sum_{i=1}^n \frac{1}{K_{t,i}} \sum_{k=0}^{K_{t,i}-1} \nabla f_i(x_{t-\tau_{t,i},k}^i) \right\|^2\right]\\
+  \eta_t\eta_l L^2 \tau \eta_0^2 \sum_{k=t-\tau_{t,u}}^{t-1}\mathbb{E}\left[\left\|  y_k \right\|^2\right] 
+ \frac{30 \eta_t\eta_l^3 L^2}{n}\sum_{i=1}^n K_{t,i}^2\mathbb{E}\left[\left\| \nabla f(x_{t-\tau_{t,i}})\right\|^2\right] + 5 \Bar{K}_t \eta_t\eta^3_l L^2\sigma_l^2 + 30 \hat{K}_t^2 \eta_t \eta^3_l L^2 \sigma_g^2\\
+ \frac{1}{2}\eta_t^3 L^2\left(\frac{\beta_t\nu_t}{1-\beta_t}\right)^2\mathbb{E}\left[\left\| d_t\right\|^2\right] + \frac{1}{2}\eta_t\mathbb{E}\left[\left\|\Delta_t\right\|^2\right]  + \frac{L}{2}\eta_t^2\mathbb{E}\left[\left\| \Delta_t\right\|^2\right]
\end{gathered}\nonumber
\end{equation}

Reorganizing terms and we have,

\begin{equation}
\begin{gathered}
 \mathbb{E}\left[\left\| \nabla f(x_t) \right\|^2\right]  \leq \frac{2\left(f(z_t) - \mathbb{E}\left[f(z_{t+1})\right]  \right)}{\eta_t\eta_l} - \mathbb{E}\left[\left\| \frac{1}{n}\sum_{i=1}^n \frac{1}{K_{t,i}} \sum_{k=0}^{K_{t,i}-1} \nabla f_i(x_{t-\tau_{t,i},k}^i) \right\|^2\right]\\
+ 2 L^2 \tau \eta_0^2 \sum_{k=t-\tau_{t,u}}^{t-1}\mathbb{E}\left[\left\|  y_k \right\|^2\right] 
+ \frac{60 \eta_l^2 L^2}{n}\sum_{i=1}^n K_{t,i}^2\mathbb{E}\left[\left\| \nabla f(x_{t-\tau_{t,i}})\right\|^2\right] + 10 \Bar{K}_t  \eta^2_l L^2 \sigma_l^2 + 60 \hat{K}_t^2 \eta^2_l L^2 \sigma_g^2\\
+ \frac{L^2 W_1^2}{\eta_l} \mathbb{E}\left[\left\| d_t\right\|^2\right] + \frac{1}{\eta_l}\mathbb{E}\left[\left\|\Delta_t\right\|^2\right]  + \frac{L\eta_t}{\eta_l}  \mathbb{E}\left[\left\| \Delta_t\right\|^2\right]
\end{gathered}\nonumber
\end{equation}

Sum over all $S$ stages and take average, we get,

\begin{equation}
\begin{gathered}
\Bar{\mathcal{G}}\triangleq\frac{1}{S}\sum_{s=0}^{S-1} \frac{1}{T_s}\sum_{t=T_0+\dots+T_{s-1} }^{T_0+\dots+T_s-1} \left\|\nabla f(x_t)\right\|^2
\leq \frac{2\left(f(z_0) - \mathbb{E}\left[f(z_{T})\right]  \right)}{S W_2 \eta_l} \\
- \frac{\eta_S}{S W_2} \mathbb{E}\left[\left\| \frac{1}{n}\sum_{i=1}^n \frac{1}{K_{t,i}} \sum_{k=0}^{K_{t,i}-1} \nabla f_i(x_{t-\tau_{t,i},k}^i) \right\|^2\right]
+  \frac{2 L^2 \tau \hat{\eta}^3}{W_2} \sum_{t=0}^{T-1} \sum_{k=t-\tau_{t,u}}^{t-1}\mathbb{E}\left[\left\|  y_k \right\|^2\right] \\
+ \frac{60 \eta_l^2 L^2}{n} \frac{1}{S}\sum_{s=0}^{S-1} \frac{1}{T_s}\sum_{t=T_0+\dots+T_{s-1} }^{T_0+\dots+T_s-1} \sum_{i=1}^n K_{t,i}^2\mathbb{E}\left[\left\| \nabla f(x_{t-\tau_{t,i}})\right\|^2\right] \\
+ \frac{L^2 W_1^2 \Bar{\eta}}{ W_2 \eta_l} \sum_{t=0}^{T-1}\mathbb{E}\left[\left\| d_t\right\|^2\right] + \frac{\Bar{\eta}}{ W_2 \eta_l} \sum_{t=0}^{T-1} \mathbb{E}\left[\left\|\Delta_t\right\|^2\right]  + \frac{L \hat{\eta}^2}{W_2 \eta_l} \sum_{t=0}^{T-1} \mathbb{E}\left[\left\| \Delta_t\right\|^2\right]\\
+ 10  \eta^2_l L^2 \left\{\frac{1}{S}\sum_{s=0}^{S-1} \frac{1}{T_s}\sum_{t=T_0+\dots+T_{s-1} }^{T_0+\dots+T_s-1} \Bar{K}_t\right\} \sigma_l^2 + 60 \eta^2_l L^2 \left\{\frac{1}{S}\sum_{s=0}^{S-1} \frac{1}{T_s}\sum_{t=T_0+\dots+T_{s-1} }^{T_0+\dots+T_s-1} \hat{K}_t^2 \right\} \sigma_g^2
\end{gathered}\nonumber
\end{equation}
where $\Bar{\eta}=\frac{1}{S}\sum_{s=0}^{S-1}\eta_s$, $\hat{\eta}^2=\frac{1}{S}\sum_{s=0}^{S-1}\eta^2_s$, and $\hat{\eta}^3=\frac{1}{S}\sum_{s=0}^{S-1}\eta^3_s$, respectively.

When the following holds, 

\begin{equation}
\begin{gathered}
\eta_l \leq \sqrt{\frac{1}{ 120L^2 C_\eta \tau K_{t,\text{max}}^2}}, \quad  \forall t \in \left\{0,\dots,T-1\right\}
\end{gathered}\nonumber
\end{equation}

we could verify the following inequality,

\begin{equation}
\begin{gathered}
\frac{60 \eta_l^2 L^2}{n} \frac{1}{S}\sum_{s=0}^{S-1} \frac{1}{T_s}\sum_{t=T_0+\dots+T_{s-1} }^{T_0+\dots+T_s-1} \sum_{i=1}^n K_{t,i}^2\mathbb{E}\left[\left\| \nabla f(x_{t-\tau_{t,i}})\right\|^2\right] \\ 
\underset{(i)}{\leq} 60 \eta_l^2 L^2  \frac{\eta_0}{SW_2} \sum_{t=0}^{T-1} K_{t,\text{max}}^2 \mathbb{E}\left[\left\| \nabla f(x_{t-\tau_{t,i}})\right\|^2\right]\\
\underset{(ii)}{\leq} 60 \eta_l^2 L^2 \tau  \frac{\eta_0}{SW_2} \sum_{t=0}^{T-1} K_{t,\text{max}}^2 \mathbb{E}\left[\left\| \nabla f(x_{t})\right\|^2\right] \underset{(iii)}{\leq} \frac{\eta_S}{2 S W_2} \sum_{t=0}^{T-1} \mathbb{E}\left[\left\| \nabla f(x_{t})\right\|^2\right]\\
\underset{(iv)}{\leq} \frac{1}{2}\frac{1}{S}\sum_{s=0}^{S-1} \frac{1}{T_s}\sum_{t=T_0+\dots+T_{s-1} }^{T_0+\dots+T_s-1} \left\|\nabla f(x_t)\right\|^2
\end{gathered}\nonumber
\end{equation}

where $(i)$ follows from the definition of $K_{t,\text{max}}^2=\max_{i\in\{1,2,\dots,n\}}K_{t,i}^2$, and $W_2=\eta_s T_s$ for all $s\in\{1,\dots,S\}$ and $\eta_S \leq \eta_s \leq \eta_0$. $(ii)$ follows from the maximum delay assumption. $(iii)$ holds by plugging in the assumption $\eta_l \leq \sqrt{\frac{\eta_S}{ 120L^2 \eta_0 \tau K_{t,\text{max}}^2}}, \quad  \forall t \in \left\{0,\dots,T-1\right\}$. $(iv)$ holds as $\frac{\eta_S}{2 S W_2} \sum_{t=0}^{T-1} \mathbb{E}\left[\left\| \nabla f(x_{t})\right\|^2\right] \leq \frac{\eta_S}{2}\frac{1}{S}\sum_{s=0}^{S-1} \frac{1}{T_s \eta_s}\sum_{t=T_0+\dots+T_{s-1} }^{T_0+\dots+T_s-1} \left\|\nabla f(x_t)\right\|^2$ and $\frac{1}{\eta_s}\leq\frac{1}{\eta_S}$ for all $s$.

With the maximum delay assumption, we have $\frac{2 L^2 \tau \hat{\eta}^3}{W_2} \sum_{t=0}^{T-1} \sum_{k=t-\tau_{t,u}}^{t-1}\mathbb{E}\left[\left\|  y_k \right\|^2\right]\leq \frac{2 L^2 \tau^2 \hat{\eta}^3}{W_2} \sum_{t=0}^{T-1}\mathbb{E}\left[\left\|  y_t \right\|^2\right]$. Merging all pieces, we have,

\begin{equation}
\begin{gathered}
\frac{1}{2} \cdot \frac{1}{S}\sum_{s=0}^{S-1} \frac{1}{T_s}\sum_{t=T_0+\dots+T_{s-1} }^{T_0+\dots+T_s-1} \left\|\nabla f(x_t)\right\|^2
\leq \frac{2\left(f(z_0) - \mathbb{E}\left[f(z_{T})\right]  \right)}{S W_2 \eta_l} \\
- \frac{\eta_S}{S W_2} \mathbb{E}\left[\left\| \frac{1}{n}\sum_{i=1}^n \frac{1}{K_{t,i}} \sum_{k=0}^{K_{t,i}-1} \nabla f_i(x_{t-\tau_{t,i},k}^i) \right\|^2\right]
+  \frac{2 L^2 \tau^2 \hat{\eta}^3}{W_2} \sum_{t=0}^{T-1}\mathbb{E}\left[\left\|  y_t \right\|^2\right] \\
+ \frac{L^2 W_1^2 \Bar{\eta}}{ W_2 \eta_l} \sum_{t=0}^{T-1}\mathbb{E}\left[\left\| d_t\right\|^2\right] + \frac{\Bar{\eta}}{ W_2 \eta_l} \sum_{t=0}^{T-1} \mathbb{E}\left[\left\|\Delta_t\right\|^2\right]  + \frac{L \hat{\eta}^2}{W_2 \eta_l} \sum_{t=0}^{T-1} \mathbb{E}\left[\left\| \Delta_t\right\|^2\right]\\
+ 10  \eta^2_l L^2 \left\{\frac{1}{S}\sum_{s=0}^{S-1} \frac{1}{T_s}\sum_{t=T_0+\dots+T_{s-1} }^{T_0+\dots+T_s-1} \Bar{K}_t\right\} \sigma_l^2 + 60 \eta^2_l L^2 \left\{\frac{1}{S}\sum_{s=0}^{S-1} \frac{1}{T_s}\sum_{t=T_0+\dots+T_{s-1} }^{T_0+\dots+T_s-1} \hat{K}_t^2 \right\} \sigma_g^2
\end{gathered}\nonumber
\end{equation}
We define $\phi_1$, $\phi_2$, and $\phi_3$ for ease of notation.

\begin{equation}
\begin{gathered}
\phi_1 \triangleq  \frac{1}{T}\sum_{t=0}^{T-1} \Bar{K}_t,  \quad \text{and} \quad  \phi_2 \triangleq  \frac{1}{T}\sum_{t=0}^{T-1} \hat{K}_t^2 ,  \quad \text{and} \quad 
\phi_3 \triangleq \frac{1}{T}\sum_{t=0}^{T-1} \frac{1}{K_t}
\end{gathered}\nonumber
\end{equation}

We could verify,

\begin{equation}
\begin{gathered}
\frac{1}{S}\sum_{s=0}^{S-1} \frac{1}{T_s}\sum_{t=T_0+\dots+T_{s-1} }^{T_0+\dots+T_s-1} \Bar{K}_t \underset{(i)}{\leq} \frac{1}{W_2} \frac{1}{S}\sum_{s=0}^{S-1} \eta_s \sum_{t=T_0+\dots+T_{s-1} }^{T_0+\dots+T_s-1} \Bar{K}_t \underset{(ii)}{\leq} \frac{\Bar{\eta}}{W_2} \sum_{t=0}^{T-1}\Bar{K}_t = \frac{T \Bar{\eta}}{W_2} \phi_1
\end{gathered}\nonumber
\end{equation}

$(i)$ holds due to $T_s\eta_s=W_2$ by assumption, $(ii)$ holds due to $\frac{1}{S}\sum_{s=0}^{S-1} \eta_s \sum_{t=T_0+\dots+T_{s-1} }^{T_0+\dots+T_s-1} \Bar{K}_t\leq\left(\frac{1}{S}\sum_{s=0}^{S-1} \eta_s\right)\cdot\left(\sum_{s=0}^{S-1}\sum_{t=T_0+\dots+T_{s-1} }^{T_0+\dots+T_s-1} \Bar{K}_t\right)=\Bar{\eta}\sum_{t=0}^{T-1}\Bar{K}_t$.

Similarly, we have,

\begin{equation}
\begin{gathered}
\frac{1}{S}\sum_{s=0}^{S-1} \frac{1}{T_s}\sum_{t=T_0+\dots+T_{s-1} }^{T_0+\dots+T_s-1} \hat{K}^2_t \leq \frac{T \Bar{\eta}}{W_2} \phi_2,  \quad \text{and} \quad
\frac{1}{S}\sum_{s=0}^{S-1} \frac{1}{T_s}\sum_{t=T_0+\dots+T_{s-1} }^{T_0+\dots+T_s-1} \frac{1}{K_t} \leq \frac{T \Bar{\eta}}{W_2} \phi_3
\end{gathered}\nonumber
\end{equation}

Plugging in the bounds for $\sum_{t=0}^{T-1}\mathbb{E}\left[\left\| \Delta_t\right\|^2\right]$, $\sum_{t=0}^{T-1}\mathbb{E}\left[\left\| y_t\right\|^2\right]$, and $\sum_{t=0}^{T-1}\mathbb{E}\left[\left\| d_t\right\|^2\right]$,

\begin{equation}
\begin{gathered}
\frac{1}{2} \cdot \frac{1}{S}\sum_{s=0}^{S-1} \frac{1}{T_s}\sum_{t=T_0+\dots+T_{s-1} }^{T_0+\dots+T_s-1} \left\|\nabla f(x_t)\right\|^2
\leq \frac{2\left(f(z_0) - \mathbb{E}\left[f(z_{T})\right]  \right)}{S W_2 \eta_l} \\
+ \frac{60 \eta^2_l L^2 T \Bar{\eta} \phi_2}{W_2} \sigma_g^2 - \frac{\eta_S}{S W_2} \mathbb{E}\left[\left\| \frac{1}{n}\sum_{i=1}^n \frac{1}{K_{t,i}} \sum_{k=0}^{K_{t,i}-1} \nabla f_i(x_{t-\tau_{t,i},k}^i) \right\|^2\right]\\
+ \left( \frac{10  \eta^2_l L^2 T \Bar{\eta}}{W_2} \phi_1 + \frac{2L^2\tau^2\hat{\eta}^3\eta_l^2T}{mW_2}\phi_3 + \frac{L^2W_1^2 \Bar{\eta} \eta_l T}{m W_2}\phi_3 + \frac{\Bar{\eta} \eta_l T}{m W_2}\phi_3 + \frac{L\hat{\eta}^2\eta_l}{m W_2}\phi_3 \right)  \sigma_l^2 \\
+ \frac{\eta_l^2}{m^2}\left(\frac{\Bar{\eta}}{W_2\eta_l}+\frac{L\hat{\eta}^2}{W_2\eta_l}+\frac{L^2W_1^2\Bar{\eta}C_\beta}{W_2\eta_l}+\frac{2L^2\tau^2\hat{\eta}^3C_\beta}{W_2}\right) \\
\cdot \mathbb{E}\left[\left\| \sum_{i=1}^n\mathbf{1}\left\{i\in\mathcal{S}_t\right\} \frac{1}{K_{t,i}} \sum_{k=0}^{K_{t,i}-1}  \nabla f_i(x_{t-\tau_{t,i},k}^i) \right\|^2\right] 
\end{gathered}\nonumber
\end{equation}

We now bound $\mathbb{E}\left[\left\| \sum_{i=1}^n\mathbf{1}\left\{i\in\mathcal{S}_t\right\} \frac{1}{K_{t,i}} \sum_{k=0}^{K_{t,i}-1}  \nabla f_i(x_{t-\tau_{t,i},k}^i) \right\|^2\right] $,

\begin{equation}
\begin{gathered}
\mathbb{E}\left[\left\| \sum_{i=1}^n\mathbf{1}\left\{i\in\mathcal{S}_t\right\} \frac{1}{K_{t,i}} \sum_{k=0}^{K_{t,i}-1} \nabla f_i(x_{t-\tau_{t,i},k}^i) \right\|^2\right]
= \sum_{i=1}^n \mathbb{P}\left\{i\in\mathcal{S}_t\right\} \left\| \frac{1}{K_{t,i}} \sum_{k=0}^{K_{t,i}-1} \nabla f_i(x_{t-\tau_{t,i},k}^i)\right\|^2 \\+ \sum_{i\neq j}\mathbb{P}\left\{i,j\in\mathcal{S}_t\right\}\left\langle \frac{1}{K_{t,i}} \sum_{k=0}^{K_{t,i}-1} \nabla f_i(x_{t-\tau_{t,i},k}^i),\frac{1}{K_{t,j}} \sum_{k=0}^{K_{t,j}-1} \nabla f_j(x_{t-\tau_{t,j},k}^j)\right\rangle\\
\underset{(i)}{=} \frac{m}{n}\sum_{i=1}^n  \left\| \frac{1}{K_{t,i}} \sum_{k=0}^{K_{t,i}-1} \nabla f_i(x_{t-\tau_{t,i},k}^i)\right\|^2\\
+ \frac{m\left(m-1\right)}{n\left(n-1\right)} \sum_{i\neq j} \left\langle \frac{1}{K_{t,i}} \sum_{k=0}^{K_{t,i}-1} \nabla f_i(x_{t-\tau_{t,i},k}^i),\frac{1}{K_{t,j}} \sum_{k=0}^{K_{t,j}-1} \nabla f_j(x_{t-\tau_{t,j},k}^j)\right\rangle\\
\underset{(ii)}{=} \frac{m^2}{n}\sum_{i=1}^n  \left\| \frac{1}{K_{t,i}} \sum_{k=0}^{K_{t,i}-1} \nabla f_i(x_{t-\tau_{t,i},k}^i)\right\|^2 \\
- \frac{m\left(m-1\right)}{2 n\left(n-1\right)} \sum_{i\neq j} \left\| \frac{1}{K_{t,i}} \sum_{k=0}^{K_{t,i}-1} \nabla f_i(x_{t-\tau_{t,i},k}^i) - \frac{1}{K_{t,j}} \sum_{k=0}^{K_{t,j}-1} \nabla f_j(x_{t-\tau_{t,j},k}^j) \right\|^2
\end{gathered}\nonumber
\end{equation}

where $(i)$ follows from uniform arrival assumption, i.e. $\mathbb{P}\left\{i,j\in\mathcal{S}_t\right\}=\frac{m(m-1)}{n(n-1)}$ and $\mathbb{P}\left\{i\in\mathcal{S}_t\right\}=\frac{m}{n}$. $(ii)$ follows from $\left\langle a, b \right\rangle = \frac{1}{2} \left\| a \right\|^2 + \frac{1}{2} \left\| b \right\|^2 - \frac{1}{2} \left\| a - b \right\|^2 $.

The following equality with respect to $\mathbb{E}\left[\left\| \sum_{i=1}^n \frac{1}{K_{t,i}} \sum_{k=0}^{K_{t,i}-1} \nabla f_i(x_{t-\tau_{t,i},k}^i) \right\|^2\right]$ is straightforward to verify,

\begin{equation}
\begin{gathered}
\mathbb{E}\left[\left\| \sum_{i=1}^n \frac{1}{K_{t,i}} \sum_{k=0}^{K_{t,i}-1} \nabla f_i(x_{t-\tau_{t,i},k}^i) \right\|^2\right]=n\sum_{i=1}^n\mathbb{E}\left[\left\| \frac{1}{K_{t,i}} \sum_{k=0}^{K_{t,i}-1} \nabla f_i(x_{t-\tau_{t,i},k}^i) \right\|^2\right] \\- \frac{1}{2}\sum_{i\neq j} \left\| \frac{1}{K_{t,i}} \sum_{k=0}^{K_{t,i}-1} \nabla f_i(x_{t-\tau_{t,i},k}^i) - \frac{1}{K_{t,j}} \sum_{k=0}^{K_{t,j}-1} \nabla f_j(x_{t-\tau_{t,j},k}^j) \right\|^2
\end{gathered}\nonumber
\end{equation}

If the following condition holds,

\begin{equation}
\begin{gathered}
2L^2\tau^2\hat{\eta}^2C_\eta^2 S \eta_l^2 + \left( L^2 W_1^2 C_\eta^2 + L \Bar{\eta} C_\eta + C_\eta \right) S \eta_l \leq \frac{m}{n} 
\end{gathered}\nonumber
\end{equation}

we have,

\begin{equation}
\begin{gathered}
- \frac{\eta_S}{S W_2} \mathbb{E}\left[\left\| \frac{1}{n}\sum_{i=1}^n \frac{1}{K_{t,i}} \sum_{k=0}^{K_{t,i}-1} \nabla f_i(x_{t-\tau_{t,i},k}^i) \right\|^2\right]\\
+ \frac{\eta_l^2}{m^2}\left(\frac{\eta_0}{SW_2\eta_l}+\frac{L\eta_0^2}{SW_2\eta_l}+\frac{L^2W_1^2\eta_0C_\beta}{SW_2\eta_l}+\frac{2L^2\tau^2\eta_0^3C_\beta}{SW_2}\right) \cdot \mathbb{E}\left[\left\| \sum_{i=1}^n\mathbf{1}\left\{i\in\mathcal{S}_t\right\} \frac{1}{K_{t,i}} \sum_{k=0}^{K_{t,i}-1}  \nabla f_i(x_{t-\tau_{t,i},k}^i) \right\|^2\right] \\
\underset{(i)}{\leq} - \frac{\eta_S}{S W_2 n^2} \mathbb{E}\left[\left\| \sum_{i=1}^n \frac{1}{K_{t,i}} \sum_{k=0}^{K_{t,i}-1} \nabla f_i(x_{t-\tau_{t,i},k}^i) \right\|^2\right]\\
+ \frac{ \eta_S }{ mn SW_2}\mathbb{E}\left[\left\| \sum_{i=1}^n\mathbf{1}\left\{i\in\mathcal{S}_t\right\} \frac{1}{K_{t,i}} \sum_{k=0}^{K_{t,i}-1}  \nabla f_i(x_{t-\tau_{t,i},k}^i) \right\|^2\right] \\
\underset{(ii)}{=}  -\frac{\eta_S}{S W_2 n}\sum_{i=1}^n  \left\| \frac{1}{K_{t,i}} \sum_{k=0}^{K_{t,i}-1} \nabla f_i(x_{t-\tau_{t,i},k}^i)\right\|^2 + \frac{m  \eta_S}{ n^2 S W_2}   \sum_{i=1}^n  \left\| \frac{1}{K_{t,i}} \sum_{k=0}^{K_{t,i}-1} \nabla f_i(x_{t-\tau_{t,i},k}^i)\right\|^2 \\
+ \frac{(n-m)\eta_S}{2n^2SW_2(n-1)} \sum_{i\neq j} \left\| \frac{1}{K_{t,i}} \sum_{k=0}^{K_{t,i}-1} \nabla f_i(x_{t-\tau_{t,i},k}^i) - \frac{1}{K_{t,j}} \sum_{k=0}^{K_{t,j}-1} \nabla f_j(x_{t-\tau_{t,j},k}^j) \right\|^2 \\
\underset{(iii)}{\leq} \left(-\frac{\eta_S}{S W_2 n}+\frac{m  \eta_S}{ n^2 S W_2}+ \frac{(n-m)\eta_S}{n^2SW_2 } \right) \sum_{i=1}^n \left\| \frac{1}{K_{t,i}} \sum_{k=0}^{K_{t,i}-1} \nabla f_i(x_{t-\tau_{t,i},k}^i)\right\|^2 \underset{(iv)}{=} 0
\end{gathered}\nonumber
\end{equation}

where $(i)$ holds by plugging in the assumption, $2L^2\tau^2\hat{\eta}^2C_\eta^2 S \eta_l^2 + \left( L^2 W_1^2 C_\eta^2 + L \Bar{\eta} C_\eta + C_\eta \right) S \eta_l \leq \frac{m}{n}$. $(ii)$ holds by plugging in the equality for $\mathbb{E}\left[\left\| \sum_{i=1}^n \frac{1}{K_{t,i}} \sum_{k=0}^{K_{t,i}-1} \nabla f_i(x_{t-\tau_{t,i},k}^i) \right\|^2\right]$ and $\mathbb{E}\left[\left\| \sum_{i=1}^n\mathbf{1}\left\{i\in\mathcal{S}_t\right\} \frac{1}{K_{t,i}} \sum_{k=0}^{K_{t,i}-1}  \nabla f_i(x_{t-\tau_{t,i},k}^i) \right\|^2\right]$. $(iii)$ holds as $\sum_{i\neq j} \left\| \frac{1}{K_{t,i}} \sum_{k=0}^{K_{t,i}-1} \nabla f_i(x_{t-\tau_{t,i},k}^i) - \frac{1}{K_{t,j}} \sum_{k=0}^{K_{t,j}-1} \nabla f_j(x_{t-\tau_{t,j},k}^j) \right\|^2 \leq 2 (n-1) \sum_{i=1}^n \left\| \frac{1}{K_{t,i}} \sum_{k=0}^{K_{t,i}-1} \nabla f_i(x_{t-\tau_{t,i},k}^i)\right\|^2$. $(iv)$ holds as $-\frac{\eta_S}{S W_2 n}+\frac{m  \eta_S}{ n^2 S W_2}+ \frac{(n-m)\eta_S}{n^2SW_2 } = 0$.

Merging all pieces together,

\begin{equation}
\begin{gathered}
 \frac{1}{S}\sum_{s=0}^{S-1} \frac{1}{T_s}\sum_{t=T_0+\dots+T_{s-1} }^{T_0+\dots+T_s-1} \left\|\nabla f(x_t)\right\|^2
\leq \frac{4 \left(f(x_0) - f^\ast  \right)}{S W_2 \eta_l} 
+ \frac{120 \eta^2_l L^2 T \Bar{\eta} \phi_2}{W_2} \sigma_g^2 \\
+ \left( \frac{20  \eta^2_l L^2 T \Bar{\eta}}{W_2} \phi_1 + \frac{4 L^2\tau^2\hat{\eta}^3\eta_l^2T}{mW_2}\phi_3 + \frac{2 L^2W_1^2 \Bar{\eta} \eta_l T}{m W_2}\phi_3 + \frac{2 \Bar{\eta} \eta_l T}{m W_2}\phi_3 + \frac{2 L\hat{\eta}^2\eta_l}{m W_2}\phi_3 \right)  \sigma_l^2
\end{gathered}\nonumber
\end{equation}

Suppose $S=1$, i.e. the typical constant hyperparameter regime, and further suppose local updating number as $K$, the total number of rounds as $T$, $\eta_0=\Bar{\eta}=\Theta\left(\sqrt{mK}\right)$ and $\eta_l=\Theta\left(\frac{1}{\sqrt{T}}\right)$. In this case, $\phi_1=K$, $\phi_2=K^2$, $\phi_3=\frac{1}{K}$, $W_2=\Theta\left(T\sqrt{mK}\right)$. Assume $W_1^2=\mathcal{O}\left(\sqrt{mK}\right)$. We have the bound as,

We have the bounds as,

\begin{equation}
\begin{gathered}
 \frac{1}{S}\sum_{s=0}^{S-1} \frac{1}{T_s}\sum_{t=T_0+\dots+T_{s-1} }^{T_0+\dots+T_s-1} \left\|\nabla f(x_t)\right\|^2
\leq \mathcal{O}\left(\frac{1}{\sqrt{mKT}}\right) \left(f(z_0) - f^\ast  \right) + \\
+\left( \mathcal{O}\left(\frac{K}{T}\right) + \mathcal{O}\left(\frac{\tau^2}{T}\right) + \mathcal{O}\left(\frac{1}{mK\sqrt{T}}\right) + \mathcal{O}\left(\frac{1}{\sqrt{mKT}}\right) \right)\sigma_l^2 +  \mathcal{O}\left( \frac{K^2}{T} \right) \sigma_g^2
\end{gathered}\nonumber
\end{equation}

Only keep the dominant terms, we could get,

\begin{equation}
\begin{gathered}
 \frac{1}{S}\sum_{s=0}^{S-1} \frac{1}{T_s}\sum_{t=T_0+\dots+T_{s-1} }^{T_0+\dots+T_s-1} \left\|\nabla f(x_t)\right\|^2
\leq \mathcal{O}\left(\frac{1}{\sqrt{mKT}}\right) \left(f(z_0) - f^\ast  \right) + \\
+\left( \mathcal{O}\left(\frac{\tau^2}{T}\right) + \mathcal{O}\left(\frac{1}{\sqrt{mKT}} \right) \right)\sigma_l^2 +  \mathcal{O}\left( \frac{K^2}{T} \right) \sigma_g^2
\end{gathered}\nonumber
\end{equation}

Suppose $S=\Theta(1)$, i.e. the multistage regime, the total number of rounds are $T$, $\Bar{\eta} = \Theta\left(\sqrt{mK}\right)$, $\hat{\eta}^2 = \Theta\left(mK\right)$, $\hat{\eta}^3 = \Theta\left(m^{\frac{3}{2}} K^{\frac{3}{2}}\right)$, and  $\eta_l=\Theta\left(\frac{1}{\sqrt{T}}\right)$, $W_2=\Theta\left(\frac{T\sqrt{mK}}{S}\right)$, i.e. $T\Bar{\eta}$ is equally divided into $S$ stages. Assume $W_1^2=\mathcal{O}\left(\sqrt{mK}\right)$. We have the bound as,

\begin{equation}
\begin{gathered}
 \frac{1}{S}\sum_{s=0}^{S-1} \frac{1}{T_s}\sum_{t=T_0+\dots+T_{s-1} }^{T_0+\dots+T_s-1} \left\|\nabla f(x_t)\right\|^2
\leq \mathcal{O}\left(\frac{1}{\sqrt{mKT}}\right) \left(f(z_0) - f^\ast  \right) + \\
+\left( \mathcal{O}\left(\frac{\tau^2}{T}\right) + \mathcal{O}\left(\frac{1}{\sqrt{mKT}} \right) \right)\sigma_l^2 +  \mathcal{O}\left( \frac{K^2}{T} \right)  \sigma_g^2
\end{gathered}\nonumber
\end{equation}

In both cases, the rate is $\mathcal{O}\left(\frac{1}{\sqrt{mKT}}\right)+\mathcal{O}\left(\frac{\tau^2}{T}\right)+\mathcal{O}\left( \frac{K^2}{T} \right)$.

\end{proof}

\section{Proof of Corollary \ref{corollary:free_multistage_fedgm_general_arrival_rate}}
\label{sec:proof_free_multistage_fedgm_general_arrival}

\begin{corollary}[Formal Statement of Corollary \ref{corollary:free_multistage_fedgm_general_arrival_rate}]
\label{multistage_fedgm_free_general_arrival_convergence_theorem}
We optimize $f(x)$ using Algorithm \ref{alg:autonomous_fedgm} (General Arrival) and $\{f_i\}_{i=1}^n$ fulfills Assumptions \ref{smoothness_assumption}-\ref{bounded_global_assumption}. Suppose bounded maximum delay, i.e. $\tau_{t,i}\leq\tau<\infty$ for any $i\in\mathcal{S}_t$ and $t\in\{1,\dots,T\}$. Denote $C_\eta\triangleq\frac{\eta_1}{\eta_S}$. Under the condition $\eta_l\leq\min\left\{\frac{1}{8K_{t,\text{max}}L},\sqrt{\frac{1}{ 180L^2 C_\eta \tau K_{t,\text{max}}^2}}\right\}$. We would have:

\begin{gather*}
\Bar{\mathcal{G}}
\leq \frac{4 \left(f(z_0) - f^\ast  \right)}{S W_2 \eta_l} + 
+\Phi_l \sigma_l^2 
+ \Phi_g \sigma_g^2
\label{multistage_fedgm_free_general_arrival_bound}
\end{gather*}

where we define $\Bar{\eta}\triangleq\frac{1}{S}\sum_{s=0}^{S-1}\eta_s$ (average server learning rate), $\hat{\eta}^2\triangleq\frac{1}{S}\sum_{s=0}^{S-1}\eta^2_s$, $\hat{\eta}^3\triangleq\frac{1}{S}\sum_{s=0}^{S-1}\eta^3_s$, $\frac{1}{K_t}=\frac{1}{m}\sum_{i\in\mathcal{S}_t}\frac{1}{K_{t,i}}$, $\bar{K}_t\triangleq \frac{1}{m}\sum_{i\in\mathcal{S}_t}K_{t,i}$, $\hat{K}_t^2 \triangleq \frac{1}{m}\sum_{i\in\mathcal{S}_t}K^2_{t,i}$, $\phi_1 \triangleq  \frac{1}{T}\sum_{t=0}^{T-1} \Bar{K}_t$, $\phi_2 \triangleq  \frac{1}{T}\sum_{t=0}^{T-1} \hat{K}_t^2$, and $\phi_3 \triangleq \frac{1}{T}\sum_{t=0}^{T-1} \frac{1}{K_t}$, for ease of notation. And $\Phi_l \triangleq \frac{30  \eta^2_l L^2 \phi_1 T \Bar{\eta}}{W_2}  + \frac{6 L^2 \tau^2 \hat{\eta}^3 \eta_l^2 T}{m W_2} \phi_3 + \frac{ 2 L^2 W_1^2 \Bar{\eta}\eta_l T}{m W_2} \phi_3 + \frac{2 \Bar{\eta}\eta_l T}{m W_2} \phi_3 + \frac{2 L \hat{\eta}^2 \eta_l T}{m W_2} \phi_3$ and $\Phi_g \triangleq 6 + \frac{180 \eta^2_l L^2 T \Bar{\eta}\phi_2 }{W_2}$.

\begin{gather*}
\Phi_l \triangleq \frac{30  \eta^2_l L^2 \phi_1 T \Bar{\eta}}{W_2}  + \frac{6 L^2 \tau^2 \hat{\eta}^3 \eta_l^2 T}{m W_2} \phi_3 + \frac{ 2 L^2 W_1^2 \Bar{\eta}\eta_l T}{m W_2} \phi_3 \\
+ \frac{2 \Bar{\eta}\eta_l T}{m W_2} \phi_3 + \frac{2 L \hat{\eta}^2 \eta_l T}{m W_2} \phi_3 \\
\Phi_g \triangleq 6 + \frac{180 \eta^2_l L^2 T \Bar{\eta}\phi_2 }{W_2}
\nonumber
\end{gather*}

Suppose $S=\Theta(1)$, i.e. the multistage regime, the total number of rounds are $T$, $\Bar{\eta} = \Theta\left(\sqrt{mK}\right)$, $\hat{\eta}^2 = \Theta\left(mK\right)$, $\hat{\eta}^3 = \Theta\left(m^{\frac{3}{2}} K^{\frac{3}{2}}\right)$, and  $\eta_l=\Theta\left(\frac{1}{\sqrt{T}}\right)$, $W_2=\Theta\left(\frac{T\sqrt{mK}}{S}\right)$, i.e. $T\Bar{\eta}$ is equally divided into $S$ stages. Assume $W_1^2=\mathcal{O}\left(\sqrt{mK}\right)$. We have the bound as,

\begin{equation}
\begin{gathered}
 \Bar{\mathcal{G}}  
\leq \mathcal{O}\left(\frac{1}{\sqrt{mKT}}\right)+ \mathcal{O}\left(\frac{\tau^2}{T}\right) +  \mathcal{O}\left( \frac{K^2}{T} \right) + \mathcal{O}\left( \sigma_g^2 \right)
\end{gathered}\nonumber
\end{equation}

\end{corollary}

\begin{proof}[Proof of Multistage GM with General Arrival]

We introduce a Lyapunov sequence $\{z_t\}_{t=0}^{T-1}$ which is devised as follows:

\begin{equation}
\label{auxiliary_seq}
z_t= x_t-\frac{\eta_t\beta_t\nu_t}{1-\beta_t}d_{t}
\end{equation}
where $d_0=0$.

We could easily verify $z_{t+1}-z_t= -\eta_t \Delta_t$. Let $-\eta y_t=x_{t+1}-x_t$, we first bound $\mathbb{E}\left[\left\| \Delta_t\right\|^2\right]$, $\sum_{t=0}^{T-1}\mathbb{E}\left[\left\| d_t\right\|^2\right]$, and $\sum_{t=0}^{T-1}\mathbb{E}\left[\left\| y_t\right\|^2\right]$.

\textbf{Bounding} $\mathbb{E}\left[\left\| \Delta_t\right\|^2\right]$:

\begin{equation}
\begin{gathered}
\mathbb{E}\left[\left\| \Delta_t\right\|^2\right] = \mathbb{E}\left[\left\| \frac{1}{m}\sum_{i\in\mathcal{S}_t}\Delta^i_{t-\tau_{t,i}} \right\|^2\right]\\
\underset{(i)}{=}\mathbb{E}\left[\left\| \frac{1}{m}\sum_{i\in\mathcal{S}_t}  \frac{\eta_l}{K_{t,i}} \sum_{k=0}^{K_{t,i}-1} g_{t-\tau_{t,i},k}^i \right\|^2\right] \\
= \mathbb{E}\left[\left\| \frac{1}{m}\sum_{i\in\mathcal{S}_t} \frac{\eta_l}{K_{t,i}} \sum_{k=0}^{K_{t,i}-1} \left( g_{t-\tau_{t,i},k}^i - \nabla f_i(x_{t-\tau_{t,i},k}^i) + \nabla f_i(x_{t-\tau_{t,i},k}^i)\right) \right\|^2\right] \\
\underset{(ii)}{=} \mathbb{E}\left[\left\| \frac{1}{m}\sum_{i\in\mathcal{S}_t} \frac{\eta_l}{K_{t,i}} \sum_{k=0}^{K_{t,i}-1} \left\{ g_{t-\tau_{t,i},k}^i - \nabla f_i(x_{t-\tau_{t,i},k}^i) \right\}  \right\|^2\right] + \mathbb{E}\left[\left\| \frac{1}{m}\sum_{i\in\mathcal{S}_t} \frac{\eta_l}{K_{t,i}} \sum_{k=0}^{K_{t,i}-1}  \nabla f_i(x_{t-\tau_{t,i},k}^i) \right\|^2\right]\\
\underset{(iii)}{\leq} \frac{1}{m^2}\sum_{i\in\mathcal{S}_t}\frac{\eta_l^2}{K^2_{t,i}}\sum_{k=0}^{K_{t,i}-1}\sigma_l^2 + \mathbb{E}\left[\left\| \frac{1}{m}\sum_{i\in\mathcal{S}_t} \frac{\eta_l}{K_{t,i}} \sum_{k=0}^{K_{t,i}-1}  \nabla f_i(x_{t-\tau_{t,i},k}^i) \right\|^2\right]\\
\leq \frac{\eta_l^2}{m}\frac{1}{K_t}\sigma^2_l + \frac{\eta_l^2}{m^2}  \mathbb{E}\left[\left\| \sum_{i\in\mathcal{S}_t} \frac{1}{K_{t,i}} \sum_{k=0}^{K_{t,i}-1}  \nabla f_i(x_{t-\tau_{t,i},k}^i) \right\|^2\right]
\end{gathered}\nonumber
\end{equation}
where $\frac{1}{K_t}=\frac{1}{m}\sum_{i\in\mathcal{S}_t}\frac{1}{K_{t,i}}$. $(i)$ follows from the definition of $\Delta^i_{t-\tau_{t,i}}$, $(ii)$ and $(iii)$ hold as $\mathbb{E}\left[\left\|\sum_{i=1}^n x_i\right\|^2\right] = \sum_{i=1}^n \mathbb{E}\left[\left\| x_i \right\|^2\right]$ when $\mathbb{E}\left[ x_i \right]=0$, and we know $\mathbb{E}\left[g_{t-\tau_{t,i},k}^i - \nabla f_i(x_{t-\tau_{t,i},k}^i)\right]=0$.

\textbf{Bounding} $\sum_{t=0}^{T-1}\mathbb{E}\left[\left\| d_t\right\|^2\right]$:

We could verify:
\begin{equation}
\begin{gathered}
 d_t = \sum_{p=0}^t a_{t,p}\Delta_p,         \quad  \text{where} \quad  a_{t,p}=\left(1-\beta_p\right)\prod_{q=p+1}^t\beta_q
\end{gathered}\nonumber
\end{equation}
We further get,
\begin{equation}
\begin{gathered}
\mathbb{E}\left[\left\| d_t\right\|^2\right]=\mathbb{E}\left[\left\| \sum_{p=0}^t a_{t,p}\Delta_p\right\|^2\right]\\
\leq \sum_{e=1}^d \mathbb{E}\left[\sum_{p=0}^t a_{t,p}\Delta_{p,e}\right]^2 
\leq \sum_{e=1}^d \mathbb{E}\left[ \left(\sum_{p=0}^t a_{t,p}\right) \left(\sum_{p=0}^t a_{t,p}\Delta_{p,e}^2\right) \right] \leq \left(1- \prod_{q=0}^t \beta_q\right)\sum_{p=0}^t a_{t,p}\mathbb{E}\left[\left\| \Delta_p \right\|^2\right]\\
\le \left(1- \prod_{q=0}^t \beta_q\right)\sum_{p=0}^t a_{t,p}\left\{ \frac{\eta_l^2}{m}\frac{1}{K_t}\sigma^2_l + \frac{\eta_l^2}{m^2}  \mathbb{E}\left[\left\| \sum_{i\in\mathcal{S}_t} \frac{1}{K_{p,i}} \sum_{k=0}^{K_{p,i}-1}  \nabla f_i(x_{p-\tau_{p,i},k}^i) \right\|^2\right] \right\}\\
\leq \frac{\eta_l^2}{m}\frac{1}{K_t}\sigma^2_l + \frac{\eta_l^2}{m^2} \sum_{p=0}^t a_{t,p} \cdot \mathbb{E}\left[\left\| \sum_{i\in\mathcal{S}_t} \frac{1}{K_{p,i}} \sum_{k=0}^{K_{p,i}-1}  \nabla f_i(x_{p-\tau_{p,i},k}^i) \right\|^2\right] 
\end{gathered}\nonumber
\end{equation}

Summing over $t\in\{0,1,\dots,T-1\}$,
\begin{equation}
\begin{gathered}
\sum_{t=0}^{T-1}\mathbb{E}\left[\left\| d_t\right\|^2\right] \leq
\frac{\eta_l^2}{m}\sum_{t=0}^{T-1}\frac{1}{K_t}\sigma^2_l + \frac{\eta_l^2}{m^2}\sum_{t=0}^{T-1} \sum_{p=0}^t a_{t,p} \cdot \mathbb{E}\left[\left\| \sum_{i\in\mathcal{S}_t} \frac{1}{K_{p,i}} \sum_{k=0}^{K_{p,i}-1}  \nabla f_i(x_{p-\tau_{p,i},k}^i) \right\|^2\right] \\
\leq \frac{\eta_l^2}{m}\sum_{t=0}^{T-1}\frac{1}{K_t}\sigma^2_l + \frac{\eta_l^2}{m^2} \sum_{p=0}^{T-1}\left(\sum_{t=p}^{T-1}a_{t,p}\right) \mathbb{E}\left[\left\| \sum_{i\in\mathcal{S}_t} \frac{1}{K_{p,i}} \sum_{k=0}^{K_{p,i}-1}  \nabla f_i(x_{p-\tau_{p,i},k}^i) \right\|^2\right]\\
\leq \frac{\eta_l^2}{m}\sum_{t=0}^{T-1}\frac{1}{K_t}\sigma^2_l + \frac{\eta_l^2}{m^2} C_\beta \sum_{t=0}^{T-1} \mathbb{E}\left[\left\| \sum_{i\in\mathcal{S}_t} \frac{1}{K_{t,i}} \sum_{k=0}^{K_{t,i}-1}  \nabla f_i(x_{t-\tau_{t,i},k}^i) \right\|^2\right]\\
\end{gathered}\nonumber
\end{equation}

\textbf{Bounding} $\sum_{t=0}^{T-1}\mathbb{E}\left[\left\| y_t\right\|^2\right]$:

We could verify:
\begin{equation}
\begin{gathered}
 y_t = \sum_{p=0}^t b_{t,p}\Delta_p,  
\end{gathered}\nonumber
\end{equation}
where $b_{t,p}$ is defined as follows,
\begin{equation}
\begin{gathered}
b_{t,p}= \begin{cases} 
    1-\beta_t\nu_t  & p = t \\
    \nu_t(1-\beta_p)\prod_{q=p+1}^t\beta_q   & p < t 
   \end{cases}
\end{gathered}\nonumber   
\end{equation}

We further get,
\begin{equation}
\begin{gathered}
\mathbb{E}\left[\left\| y_t\right\|^2\right]=\mathbb{E}\left[\left\| \sum_{p=0}^t b_{t,p}\Delta_p\right\|^2\right]\\
\leq \sum_{e=1}^d \mathbb{E}\left[\sum_{p=0}^t b_{t,p}\Delta_{p,e}\right]^2 
\leq \sum_{e=1}^d \mathbb{E}\left[ \left(\sum_{p=0}^t b_{t,p}\right) \left(\sum_{p=0}^t b_{t,p}\Delta_{p,e}^2\right) \right] \leq \left(1- \nu_t\prod_{q=0}^t \beta_q\right)\sum_{p=0}^t b_{t,p}\mathbb{E}\left[\left\| \Delta_p \right\|^2\right]\\
\le \left(1- \nu_t \prod_{q=0}^t \beta_q\right)\sum_{p=0}^t b_{t,p}\left\{ \frac{\eta_l^2}{m}\frac{1}{K_t}\sigma^2_l + \frac{\eta_l^2}{m^2}  \mathbb{E}\left[\left\| \sum_{i\in\mathcal{S}_t} \frac{1}{K_{p,i}} \sum_{k=0}^{K_{p,i}-1}  \nabla f_i(x_{p-\tau_{p,i},k}^i) \right\|^2\right] \right\}\\
\leq \frac{\eta_l^2}{m}\frac{1}{K_t}\sigma^2_l + \frac{\eta_l^2}{m^2} \sum_{p=0}^t b_{t,p} \cdot \mathbb{E}\left[\left\| \sum_{i\in\mathcal{S}_t} \frac{1}{K_{p,i}} \sum_{k=0}^{K_{p,i}-1}  \nabla f_i(x_{p-\tau_{p,i},k}^i) \right\|^2\right] 
\end{gathered}\nonumber
\end{equation}

Summing over $t\in\{0,1,\dots,T-1\}$,
\begin{equation}
\begin{gathered}
\sum_{t=0}^{T-1}\mathbb{E}\left[\left\| y_t\right\|^2\right] \leq
\frac{\eta_l^2}{m}\sum_{t=0}^{T-1}\frac{1}{K_t}\sigma^2_l + \frac{\eta_l^2}{m^2}\sum_{t=0}^{T-1} \sum_{p=0}^t b_{t,p} \cdot \mathbb{E}\left[\left\| \sum_{i\in\mathcal{S}_t} \frac{1}{K_{p,i}} \sum_{k=0}^{K_{p,i}-1}  \nabla f_i(x_{p-\tau_{p,i},k}^i) \right\|^2\right] \\
\leq \frac{\eta_l^2}{m}\sum_{t=0}^{T-1}\frac{1}{K_t}\sigma^2_l + \frac{\eta_l^2}{m^2} \sum_{p=0}^{T-1}\left(\sum_{t=p}^{T-1}b_{t,p}\right) \mathbb{E}\left[\left\| \sum_{i\in\mathcal{S}_t} \frac{1}{K_{p,i}} \sum_{k=0}^{K_{p,i}-1}  \nabla f_i(x_{p-\tau_{p,i},k}^i) \right\|^2\right]\\
\leq \frac{\eta_l^2}{m}\sum_{t=0}^{T-1}\frac{1}{K_t}\sigma^2_l + \frac{\eta_l^2}{m^2} C_\beta \sum_{t=0}^{T-1} \mathbb{E}\left[\left\| \sum_{i\in\mathcal{S}_t} \frac{1}{K_{t,i}} \sum_{k=0}^{K_{t,i}-1}  \nabla f_i(x_{t-\tau_{t,i},k}^i) \right\|^2\right]\\
\end{gathered}\nonumber
\end{equation}

Since $f$ is $L$-smooth, taking conditional expectation with respect to all randomness prior to step $t$, we have

\begin{equation}
\begin{gathered}
\mathbb{E}\left[f(z_{t+1})\right]\leq
f(z_t)+\mathbb{E}\left[\left\langle \nabla f(z_t),z_{t+1}-z_t \right\rangle\right]+\frac{L}{2}\mathbb{E}\left[\left\| z_{t+1}-z_t\right\|^2\right]\\
\leq f(z_t)+\mathbb{E}\left[\left\langle \nabla f(z_t),-\eta_t  \Delta_t \right\rangle\right]+\frac{L}{2}\eta_t^2\mathbb{E}\left[\left\| \Delta_t\right\|^2\right]\\
\leq f(z_t)+ \underbrace{\mathbb{E}\left[\left\langle \sqrt{\eta_t} \left(\nabla f(z_t)-\nabla f(x_t)\right),-\sqrt{\eta_t } \Delta_t \right\rangle\right]}_{A_1} + \underbrace{\mathbb{E}\left[\left\langle \nabla f(x_t),-\eta_t \Delta_t \right\rangle\right]}_{A_2} + \underbrace{\frac{L}{2}\eta_t^2\mathbb{E}\left[\left\| \Delta_t\right\|^2\right]}_{A_3} \\
\end{gathered}\nonumber
\end{equation}

\textbf{Bounding} $A_1$:
\begin{equation}
\begin{gathered}
A_1 =\mathbb{E}\left[\left\langle \sqrt{\eta_t} \left(\nabla f(z_t)-\nabla f(x_t)\right),-\sqrt{\eta_t} \Delta_t \right\rangle\right]\\
\underset{(i)}{\leq}\mathbb{E}\left[\left\|\sqrt{\eta_t} \left(\nabla f(z_t)-\nabla f(x_t)\right)\right\| \cdot \left\|-\sqrt{\eta_t} \Delta_t\right\|\right]\\
\underset{(ii)}{\leq}\frac{1}{2}\eta_t^3 L^2\left(\frac{\beta_t\nu_t}{1-\beta_t}\right)^2\mathbb{E}\left[\left\| d_t\right\|^2\right] + \frac{1}{2}\eta_t\mathbb{E}\left[\left\|\Delta_t\right\|^2\right]
\end{gathered}\nonumber
\end{equation}

where $(i)$ holds by applying Cauchy-Schwarz inequality, and $(ii)$ follows from Young’s inequality and $f$ is $L$-smooth.

\textbf{Bounding} $A_2$:
\begin{equation}
\begin{gathered}
A_2=\mathbb{E}\left[\left\langle \nabla f(x_t),-\eta_t \Delta_t \right\rangle\right]\\
=\eta_t\mathbb{E}\left[\left\langle \nabla f\left(x_t\right),\eta_l  \nabla f\left(x_t\right) - \Delta_t - \eta_l  \nabla f\left(x_t\right)  \right\rangle\right]\\
=-\eta_t \eta_l   \mathbb{E}\left [\left\| \nabla f\left(x_t\right) \right\|^2\right]+\eta_t\mathbb{E}\left[\left\langle \nabla f\left(x_t\right), \eta_l   \nabla f\left(x_t\right) - \Delta_t \right\rangle\right]
\end{gathered}\nonumber
\end{equation}
where we further bound $\eta_t\mathbb{E}\left[\langle \nabla f\left(x_t\right), \eta_l   \nabla f\left(x_t\right) - \Delta_t \rangle\right]$,
 
\begin{equation}
\begin{gathered}
\eta_t\mathbb{E}\left[\langle \nabla f\left(x_t\right), \eta_l   \nabla f\left(x_t\right) - \Delta_t \rangle\right]
=\eta_t \mathbb{E}\left[\left\langle \sqrt{\eta_l} \nabla f(x_t),   \frac{\sqrt{\eta_l}}{m}\sum_{i\in\mathcal{S}_t} \frac{1}{K_{t,i}} \sum_{k=0}^{K_{t,i}-1}\left(\nabla f(x_t) -  g_{t-\tau_{t,i},k}^i\right) \right\rangle\right]\\
\underset{(i)}{=}\eta_t \mathbb{E}\left[\left\langle \sqrt{\eta_l} \nabla f(x_t),   \frac{\sqrt{\eta_l}}{m}\sum_{i\in\mathcal{S}_t} \frac{1}{K_{t,i}} \sum_{k=0}^{K_{t,i}-1}\left(\nabla f(x_t) -  \nabla f_i(x_{t-\tau_{t,i},k}^i)\right) \right\rangle\right]\\
\underset{(ii)}{=} \frac{\eta_t\eta_l}{2} \mathbb{E}\left[\left\| \nabla f(x_t) \right\|^2\right] - \frac{\eta_t\eta_l}{2} \mathbb{E}\left[\left\| \frac{1}{m}\sum_{i\in\mathcal{S}_t} \frac{1}{K_{t,i}} \sum_{k=0}^{K_{t,i}-1} \nabla f_i(x_{t-\tau_{t,i},k}^i) \right\|^2\right] \\
+ \frac{\eta_t\eta_l}{2} \mathbb{E}\left[\left\| \nabla f(x_t) - \frac{1}{m}\sum_{i\in\mathcal{S}_t} \frac{1}{K_{t,i}} \sum_{k=0}^{K_{t,i}-1}  \nabla f_i(x_{t-\tau_{t,i},k}^i) \right\|^2\right]
\end{gathered}\nonumber
\end{equation}

where $(i)$ holds as we take conditional expectation with respect to all randomness prior to step $t$. $(ii)$ holds as $\left\langle a, b \right\rangle = \frac{1}{2} \left\| a \right\|^2 + \frac{1}{2} \left\| b \right\|^2 - \frac{1}{2} \left\| a - b \right\|^2 $.

We further have,

\begin{equation}
\begin{gathered}
\frac{\eta_t\eta_l}{2} \mathbb{E}\left[\left\| \nabla f(x_t) - \frac{1}{m}\sum_{i\in\mathcal{S}_t} \frac{1}{K_{t,i}} \sum_{k=0}^{K_{t,i}-1}  \nabla f_i(x_{t-\tau_{t,i},k}^i) \right\|^2\right] \\
= \frac{\eta_t\eta_l}{2} \mathbb{E}\left[\left\| \frac{1}{m}\sum_{i\in\mathcal{S}_t} \frac{1}{K_{t,i}} \sum_{k=0}^{K_{t,i}-1} \left( \nabla f(x_t)  -\nabla f_i(x_{t-\tau_{t,i},k}^i) \right) \right\|^2\right]\\
\underset{(i)}{\leq} \frac{3}{2} \eta_t\eta_l \mathbb{E}\left[\left\| \frac{1}{m}\sum_{i\in\mathcal{S}_t} \left( \nabla f(x_t)  -\nabla f_i(x_t) \right) \right\|^2\right] + \frac{3}{2} \eta_t\eta_l \mathbb{E}\left[\left\| \frac{1}{m}\sum_{i\in\mathcal{S}_t} \left( \nabla f_i(x_t)  -\nabla f_i(x_{t-\tau_{t,i}}) \right) \right\|^2\right] \\
+ \frac{3}{2} \eta_t\eta_l \mathbb{E}\left[\left\| \frac{1}{m}\sum_{i\in\mathcal{S}_t} \frac{1}{K_{t,i}} \sum_{k=0}^{K_{t,i}-1} \left( \nabla f_i(x_{t-\tau_{t,i}}) - \nabla f_i(x_{t-\tau_{t,i},k}^i) \right) \right\|^2\right]\\
\underset{(ii)}{\leq}  \frac{3}{2} \eta_t\eta_l \frac{1}{m}\sum_{i\in\mathcal{S}_t} \mathbb{E}\left[\left\|    \nabla f(x_t)  -\nabla f_i(x_t)   \right\|^2\right] + \frac{3}{2} \eta_t\eta_l \frac{1}{m}\sum_{i\in\mathcal{S}_t} \mathbb{E}\left[\left\|    \nabla f_i(x_t)  -\nabla f_i(x_{t-\tau_{t,i}})   \right\|^2\right] \\
+ \frac{3}{2} \eta_t\eta_l \frac{1}{m}\sum_{i\in\mathcal{S}_t} \mathbb{E}\left[\left\|  \frac{1}{K_{t,i}} \sum_{k=0}^{K_{t,i}-1}   \nabla f_i(x_{t-\tau_{t,i}}) - \nabla f_i(x_{t-\tau_{t,i},k}^i)  \right\|^2\right]\\
\underset{(iii)}{\leq} \frac{3}{2} \eta_t\eta_l \sigma_g^2 +  \frac{3\eta_t\eta_l L^2}{2m} \sum_{i\in\mathcal{S}_t} \mathbb{E}\left[\left\|  x_t  - x_{t-\tau_{t,i}} \right\|^2\right] + \frac{3 \eta_t \eta_l L^2}{2 m} \sum_{i\in\mathcal{S}_t} \frac{1}{K_{t,i}} \sum_{k=0}^{K_{t,i}-1} \mathbb{E}\left[\left\|    x_{t-\tau_{t,i}}  -  x_{t-\tau_{t,i},k}^i \right\|^2\right]
\end{gathered}\nonumber
\end{equation}

where $(i)$ and $(ii)$ hold as $\left\|\sum_{i=1}^n x_i\right\|^2 \leq n \sum_{i=1}^n\left\| x_i \right\|^2$, $(iii)$ holds as $f_i$ is $L$-smooth.

Thus, we have, 

\begin{equation}
\begin{gathered}
A_2
\leq - \frac{\eta_t\eta_l}{2} \mathbb{E}\left[\left\| \nabla f(x_t) \right\|^2\right] - \frac{\eta_t\eta_l}{2} \mathbb{E}\left[\left\| \frac{1}{m}\sum_{i\in\mathcal{S}_t} \frac{1}{K_{t,i}} \sum_{k=0}^{K_{t,i}-1} \nabla f_i(x_{t-\tau_{t,i},k}^i) \right\|^2\right] \\
\frac{3}{2} \eta_t\eta_l \sigma_g^2 +  \frac{3\eta_t\eta_l L^2}{2m} \sum_{i\in\mathcal{S}_t} \mathbb{E}\left[\left\|  x_t  - x_{t-\tau_{t,i}} \right\|^2\right] + \frac{3 \eta_t \eta_l L^2}{2 m} \sum_{i\in\mathcal{S}_t} \frac{1}{K_{t,i}} \sum_{k=0}^{K_{t,i}-1} \mathbb{E}\left[\left\|    x_{t-\tau_{t,i}}  -  x_{t-\tau_{t,i},k}^i \right\|^2\right]
\end{gathered}\nonumber
\end{equation}

When $\eta_l\leq \frac{1}{8K_{t,i}L}$, we have,
\begin{equation}
\begin{gathered}
\mathbb{E}\left[\left\| x_{t-\tau_{t,i}}-x_{t-\tau_{t,i},k}^i  \right\|^2\right] \leq 5 K_{t,i}\eta_l^2\left(\sigma_l^2+6K_{t,i}\sigma_g^2\right)+30K_{t,i}^2\eta_l^2 \mathbb{E}\left[\left\| \nabla f(x_{t-\tau_{t,i}})\right\|^2\right]
\end{gathered}\nonumber
\end{equation}

We can further bound $\frac{1}{m}\sum_{i\in\mathcal{S}_t} \mathbb{E}\left[\left\| x_t - x_{t-\tau_{t,i}} \right\|^2\right]$
\begin{equation}
\begin{gathered}
\frac{1}{m}\sum_{i\in\mathcal{S}_t} \mathbb{E}\left[\left\| x_t - x_{t-\tau_{t,i}} \right\|^2\right] 
\underset{(i)}{\leq}  \mathbb{E}\left[\left\| x_t - x_{t-\tau_{t,u}} \right\|^2\right] = \mathbb{E}\left[\left\| \sum_{k=t-\tau_{t,u}}^{t-1} \left(x_{k+1}-x_k\right) \right\|^2\right] \\
\underset{(ii)}{\leq} \mathbb{E}\left[\left\| \sum_{k=t-\tau_{t,u}}^{t-1} \eta_k y_k \right\|^2\right] \underset{(iii)}{\leq} \tau \eta_{t-\tau_{t,u}}^2 \sum_{k=t-\tau_{t,u}}^{t-1}\mathbb{E}\left[\left\|  y_k \right\|^2\right]
\end{gathered}\nonumber
\end{equation}

where $(i)$ holds as we define $u=\argmax_{i\in\{1,2,\dots,n\}}\mathbb{E}\left[\left\| x_t - x_{t-\tau_{t,i}} \right\|^2\right] $, $(ii)$ follows from the definition of $y_k$, $(iii)$ holds as bounded maximum delay assumption, i.e. $\tau_{t,i}\leq \tau$ for any $t$ and $i$, and learning rate is decaying, i.e. $\eta_t \leq \eta_{t-\tau_{t,u}}$.

Merging all pieces together, we have the following,

\begin{equation}
\begin{gathered}
A_2 \leq - \frac{\eta_t\eta_l}{2} \mathbb{E}\left[\left\| \nabla f(x_t) \right\|^2\right] - \frac{\eta_t\eta_l}{2} \mathbb{E}\left[\left\| \frac{1}{m}\sum_{i\in\mathcal{S}_t} \frac{1}{K_{t,i}} \sum_{k=0}^{K_{t,i}-1} \nabla f_i(x_{t-\tau_{t,i},k}^i) \right\|^2\right] \\
\frac{3}{2} \eta_t\eta_l \sigma_g^2 +  \frac{3\eta_t\eta_l L^2}{2m} \sum_{i\in\mathcal{S}_t} \mathbb{E}\left[\left\|  x_t  - x_{t-\tau_{t,i}} \right\|^2\right] + \frac{3 \eta_t \eta_l L^2}{2 m} \sum_{i\in\mathcal{S}_t} \frac{1}{K_{t,i}} \sum_{k=0}^{K_{t,i}-1} \mathbb{E}\left[\left\|    x_{t-\tau_{t,i}}  -  x_{t-\tau_{t,i},k}^i \right\|^2\right]\\
\leq - \frac{\eta_t\eta_l}{2} \mathbb{E}\left[\left\| \nabla f(x_t) \right\|^2\right] - \frac{\eta_t\eta_l}{2} \mathbb{E}\left[\left\| \frac{1}{m}\sum_{i\in\mathcal{S}_t} \frac{1}{K_{t,i}} \sum_{k=0}^{K_{t,i}-1} \nabla f_i(x_{t-\tau_{t,i},k}^i) \right\|^2\right] 
+ \frac{3}{2} \eta_t\eta_l \sigma_g^2 \\
+ \frac{15}{2}L^2\eta_t\eta_l^3\bar{K}_t\sigma_l^2+45L^2\eta_t\eta_l^3\hat{K}_t^2\sigma_g^2+45L^2\eta_t\eta_l^3 \frac{1}{m}\sum_{i\in\mathcal{S}_t}K_{t,i}^2\mathbb{E}\left[\left\| \nabla f(x_{t-\tau_{t,i}})\right\|^2\right]\\
+\frac{3}{2}\tau L^2\eta_t\eta_{t-\tau_{t,u}}^2\eta_l \sum_{k=t-\tau_{t,u}}^{t-1}\mathbb{E}\left[\left\|  y_k \right\|^2\right]
\end{gathered}\nonumber
\end{equation}

where $\bar{K}_t\triangleq \frac{1}{m}\sum_{i\in\mathcal{S}_t}K_{t,i}$ and $\hat{K}_t^2 \triangleq \frac{1}{m}\sum_{i\in\mathcal{S}_t}K^2_{t,i}$.

Plug all pieces back in $\mathbb{E}\left[f(z_{t+1})\right] \leq f(z_t) + A_1 + A_2 + A_3$,

\begin{equation}
\begin{gathered}
\mathbb{E}\left[f(z_{t+1})\right] - f(z_t) \leq 
- \frac{\eta_t\eta_l}{2} \mathbb{E}\left[\left\| \nabla f(x_t) \right\|^2\right] - \frac{\eta_t\eta_l}{2} \mathbb{E}\left[\left\| \frac{1}{m}\sum_{i\in\mathcal{S}_t} \frac{1}{K_{t,i}} \sum_{k=0}^{K_{t,i}-1} \nabla f_i(x_{t-\tau_{t,i},k}^i) \right\|^2\right] 
+ \frac{3}{2} \eta_t\eta_l \sigma_g^2 \\
+ \frac{15}{2}L^2\eta_t\eta_l^3\bar{K}_t\sigma_l^2+45L^2\eta_t\eta_l^3\hat{K}_t^2\sigma_g^2+45L^2\eta_t\eta_l^3 \frac{1}{m}\sum_{i\in\mathcal{S}_t}K_{t,i}^2\mathbb{E}\left[\left\| \nabla f(x_{t-\tau_{t,i}})\right\|^2\right]\\
+\frac{3}{2}\tau L^2\eta_t\eta_{t-\tau_{t,u}}^2\eta_l \sum_{k=t-\tau_{t,u}}^{t-1}\mathbb{E}\left[\left\|  y_k \right\|^2\right]\\
+ \frac{1}{2}\eta_t^3 L^2\left(\frac{\beta_t\nu_t}{1-\beta_t}\right)^2\mathbb{E}\left[\left\| d_t\right\|^2\right] + \frac{1}{2}\eta_t\mathbb{E}\left[\left\|\Delta_t\right\|^2\right]  + \frac{L}{2}\eta_t^2\mathbb{E}\left[\left\| \Delta_t\right\|^2\right]
\end{gathered}\nonumber
\end{equation}

Reorganizing terms and we have,

\begin{equation}
\begin{gathered}
 \mathbb{E}\left[\left\| \nabla f(x_t) \right\|^2\right]  \leq \frac{2\left(f(z_t) - \mathbb{E}\left[f(z_{t+1})\right]  \right)}{\eta_t\eta_l} - \mathbb{E}\left[\left\| \frac{1}{m}\sum_{i\in\mathcal{S}_t} \frac{1}{K_{t,i}} \sum_{k=0}^{K_{t,i}-1} \nabla f_i(x_{t-\tau_{t,i},k}^i) \right\|^2\right]
+ 3 \sigma_g^2 \\
+ 15 L^2 \eta_l^2\bar{K}_t\sigma_l^2+ 90 L^2 \eta_l^2 \hat{K}_t^2\sigma_g^2 + 90 L^2 \eta_l^2 \frac{1}{m}\sum_{i\in\mathcal{S}_t}K_{t,i}^2\mathbb{E}\left[\left\| \nabla f(x_{t-\tau_{t,i}})\right\|^2\right] + 3 \tau L^2 \eta_{t-\tau_{t,u}}^2  \sum_{k=t-\tau_{t,u}}^{t-1}\mathbb{E}\left[\left\|  y_k \right\|^2\right]\\
+ \frac{1}{\eta_l}\eta_t^2 L^2\left(\frac{\beta_t\nu_t}{1-\beta_t}\right)^2\mathbb{E}\left[\left\| d_t\right\|^2\right] + \frac{1}{\eta_l} \mathbb{E}\left[\left\|\Delta_t\right\|^2\right]  + \frac{L}{\eta_l}\eta_t\mathbb{E}\left[\left\| \Delta_t\right\|^2\right]
\end{gathered}\nonumber
\end{equation}

Sum over all $S$ stages and take average, by some algebraic transformations, we get,

\begin{equation}
\begin{gathered}
\Bar{\mathcal{G}}\triangleq\frac{1}{S}\sum_{s=0}^{S-1} \frac{1}{T_s}\sum_{t=T_0+\dots+T_{s-1} }^{T_0+\dots+T_s-1} \left\|\nabla f(x_t)\right\|^2
\leq \frac{2\left(f(z_0) - \mathbb{E}\left[f(z_{T})\right]  \right)}{S W_2 \eta_l} \\
- \frac{\eta_S}{S W_2} \mathbb{E}\left[\left\| \frac{1}{m}\sum_{i\in\mathcal{S}_t} \frac{1}{K_{t,i}} \sum_{k=0}^{K_{t,i}-1} \nabla f_i(x_{t-\tau_{t,i},k}^i) \right\|^2\right] + 3 \sigma_g^2
+  \frac{3 L^2 \tau \hat{\eta}^3}{W_2} \sum_{t=0}^{T-1} \sum_{k=t-\tau_{t,u}}^{t-1}\mathbb{E}\left[\left\|  y_k \right\|^2\right] \\
+ \frac{90 \eta_l^2 L^2}{m} \frac{1}{S}\sum_{s=0}^{S-1} \frac{1}{T_s}\sum_{t=T_0+\dots+T_{s-1} }^{T_0+\dots+T_s-1} \sum_{i\in\mathcal{S}_t} K_{t,i}^2\mathbb{E}\left[\left\| \nabla f(x_{t-\tau_{t,i}})\right\|^2\right] \\
+ \frac{L^2 W_1^2 \Bar{\eta}}{W_2 \eta_l} \sum_{t=0}^{T-1}\mathbb{E}\left[\left\| d_t\right\|^2\right] + \frac{\Bar{\eta}}{W_2 \eta_l} \sum_{t=0}^{T-1} \mathbb{E}\left[\left\|\Delta_t\right\|^2\right]  + \frac{L\hat{\eta}^2}{S W_2 \eta_l} \sum_{t=0}^{T-1} \mathbb{E}\left[\left\| \Delta_t\right\|^2\right]\\
+ 15 \eta^2_l L^2 \left\{\frac{1}{S}\sum_{s=0}^{S-1} \frac{1}{T_s}\sum_{t=T_0+\dots+T_{s-1} }^{T_0+\dots+T_s-1} \Bar{K}_t\right\} \sigma_l^2 + 90 \eta^2_l L^2 \left\{\frac{1}{S}\sum_{s=0}^{S-1} \frac{1}{T_s}\sum_{t=T_0+\dots+T_{s-1} }^{T_0+\dots+T_s-1} \hat{K}_t^2 \right\} \sigma_g^2
\end{gathered}\nonumber
\end{equation}

where $\Bar{\eta}=\frac{1}{S}\sum_{s=0}^{S-1}\eta_s$, $\hat{\eta}^2=\frac{1}{S}\sum_{s=0}^{S-1}\eta^2_s$, and $\hat{\eta}^3=\frac{1}{S}\sum_{s=0}^{S-1}\eta^3_s$, respectively.

When the following holds, 

\begin{equation}
\begin{gathered}
\eta_l \leq \sqrt{\frac{1}{ 180L^2 C_\eta \tau K_{t,\text{max}}^2}}, \quad  \forall t \in \left\{0,\dots,T-1\right\}
\end{gathered}\nonumber
\end{equation}

where $C_\eta=\frac{\eta_0}{\eta_S}$.

we could verify the following inequality,

\begin{equation}
\begin{gathered}
\frac{90 \eta_l^2 L^2}{m} \frac{1}{S}\sum_{s=0}^{S-1} \frac{1}{T_s}\sum_{t=T_0+\dots+T_{s-1} }^{T_0+\dots+T_s-1} \sum_{i\in\mathcal{S}_t} K_{t,i}^2\mathbb{E}\left[\left\| \nabla f(x_{t-\tau_{t,i}})\right\|^2\right] \\ 
\underset{(i)}{\leq} 90 \eta_l^2 L^2  \frac{\eta_0}{S W_2} \sum_{t=0}^{T-1} K_{t,\text{max}}^2 \mathbb{E}\left[\left\| \nabla f(x_{t-\tau_{t,i}})\right\|^2\right]\\
\underset{(ii)}{\leq} 90 \eta_l^2 L^2 \tau  \frac{\eta_0}{S W_2} \sum_{t=0}^{T-1} K_{t,\text{max}}^2 \mathbb{E}\left[\left\| \nabla f(x_{t})\right\|^2\right] \underset{(iii)}{\leq} \frac{\eta_S}{2 S W_2} \sum_{t=0}^{T-1} \mathbb{E}\left[\left\| \nabla f(x_{t})\right\|^2\right]\\
\underset{(iv)}{\leq} \frac{1}{2}\frac{1}{S}\sum_{s=0}^{S-1} \frac{1}{T_s}\sum_{t=T_0+\dots+T_{s-1} }^{T_0+\dots+T_s-1} \left\|\nabla f(x_t)\right\|^2
\end{gathered}\nonumber
\end{equation}

where $(i)$ follows from the definition of $K_{t,\text{max}}^2=\max_{i\in\{1,2,\dots,n\}}K_{t,i}^2$, and $W_2=\eta_s T_s$ for all $s\in\{1,\dots,S\}$ and $\eta_S \leq \eta_s \leq \eta_0$. $(ii)$ follows from the maximum delay assumption. $(iii)$ holds by plugging in the assumption $\eta_l \leq \sqrt{\frac{\eta_S}{ 180L^2 \eta_0 \tau K_{t,\text{max}}^2}}, \quad  \forall t \in \left\{0,\dots,T-1\right\}$. $(iv)$ holds as $\frac{\eta_S}{2 S W_2} \sum_{t=0}^{T-1} \mathbb{E}\left[\left\| \nabla f(x_{t})\right\|^2\right] \leq \frac{\eta_S}{2}\frac{1}{S}\sum_{s=0}^{S-1} \frac{1}{T_s \eta_s}\sum_{t=T_0+\dots+T_{s-1} }^{T_0+\dots+T_s-1} \left\|\nabla f(x_t)\right\|^2$ and $\frac{1}{\eta_s}\leq\frac{1}{\eta_S}$ for all $s$.

With the maximum delay assumption, we have $\frac{3 L^2 \tau \hat{\eta}^3}{W_2} \sum_{t=0}^{T-1} \sum_{k=t-\tau_{t,u}}^{t-1}\mathbb{E}\left[\left\|  y_k \right\|^2\right] \leq \frac{3 L^2 \tau^2 \hat{\eta}^3}{W_2} \sum_{t=0}^{T-1} \mathbb{E}\left[\left\|  y_t \right\|^2\right]$. Merging all pieces, we have,

\begin{equation}
\begin{gathered}
\frac{1}{2} \cdot \frac{1}{S}\sum_{s=0}^{S-1} \frac{1}{T_s}\sum_{t=T_0+\dots+T_{s-1} }^{T_0+\dots+T_s-1} \left\|\nabla f(x_t)\right\|^2
\leq \frac{2\left(f(z_0) - \mathbb{E}\left[f(z_{T})\right]  \right)}{S W_2 \eta_l} \\
- \frac{\eta_S}{S W_2} \mathbb{E}\left[\left\| \frac{1}{m}\sum_{i\in\mathcal{S}_t} \frac{1}{K_{t,i}} \sum_{k=0}^{K_{t,i}-1} \nabla f_i(x_{t-\tau_{t,i},k}^i) \right\|^2\right] + 3 \sigma_g^2
+ \frac{3 L^2 \tau^2 \hat{\eta}^3}{W_2} \sum_{t=0}^{T-1}\mathbb{E}\left[\left\|  y_t \right\|^2\right] \\
+ \frac{L^2 W_1^2 \Bar{\eta}}{W_2 \eta_l} \sum_{t=0}^{T-1}\mathbb{E}\left[\left\| d_t\right\|^2\right] + \frac{\Bar{\eta}}{W_2 \eta_l} \sum_{t=0}^{T-1} \mathbb{E}\left[\left\|\Delta_t\right\|^2\right]  + \frac{L \hat{\eta}^2}{W_2 \eta_l} \sum_{t=0}^{T-1} \mathbb{E}\left[\left\| \Delta_t\right\|^2\right]\\
+ 15 \eta^2_l L^2 \left\{\frac{1}{S}\sum_{s=0}^{S-1} \frac{1}{T_s}\sum_{t=T_0+\dots+T_{s-1} }^{T_0+\dots+T_s-1} \Bar{K}_t\right\} \sigma_l^2 + 90 \eta^2_l L^2 \left\{\frac{1}{S}\sum_{s=0}^{S-1} \frac{1}{T_s}\sum_{t=T_0+\dots+T_{s-1} }^{T_0+\dots+T_s-1} \hat{K}_t^2 \right\} \sigma_g^2
\end{gathered}\nonumber
\end{equation}

We define $\phi_1$, $\phi_2$, and $\phi_3$ for ease of notation.

\begin{equation}
\begin{gathered}
\phi_1 \triangleq  \frac{1}{T}\sum_{t=0}^{T-1} \Bar{K}_t,  \quad \text{and} \quad  \phi_2 \triangleq  \frac{1}{T}\sum_{t=0}^{T-1} \hat{K}_t^2 ,  \quad \text{and} \quad 
\phi_3 \triangleq \frac{1}{T}\sum_{t=0}^{T-1} \frac{1}{K_t}
\end{gathered}\nonumber
\end{equation}

We could verify,

\begin{equation}
\begin{gathered}
\frac{1}{S}\sum_{s=0}^{S-1} \frac{1}{T_s}\sum_{t=T_0+\dots+T_{s-1} }^{T_0+\dots+T_s-1} \Bar{K}_t \underset{(i)}{\leq} \frac{1}{W_2} \frac{1}{S}\sum_{s=0}^{S-1} \eta_s \sum_{t=T_0+\dots+T_{s-1} }^{T_0+\dots+T_s-1} \Bar{K}_t \underset{(ii)}{\leq} \frac{\Bar{\eta}}{W_2} \sum_{t=0}^{T-1}\Bar{K}_t = \frac{T \Bar{\eta}}{W_2} \phi_1
\end{gathered}\nonumber
\end{equation}

$(i)$ holds due to $T_s\eta_s=W_2$ by assumption, $(ii)$ holds due to $\frac{1}{S}\sum_{s=0}^{S-1} \eta_s \sum_{t=T_0+\dots+T_{s-1} }^{T_0+\dots+T_s-1} \Bar{K}_t\leq\left(\frac{1}{S}\sum_{s=0}^{S-1} \eta_s\right)\cdot\left(\sum_{s=0}^{S-1}\sum_{t=T_0+\dots+T_{s-1} }^{T_0+\dots+T_s-1} \Bar{K}_t\right)=\Bar{\eta}\sum_{t=0}^{T-1}\Bar{K}_t$.

Similarly, we have,

\begin{equation}
\begin{gathered}
\frac{1}{S}\sum_{s=0}^{S-1} \frac{1}{T_s}\sum_{t=T_0+\dots+T_{s-1} }^{T_0+\dots+T_s-1} \hat{K}^2_t \leq \frac{T \Bar{\eta}}{W_2} \phi_2,  \quad \text{and} \quad
\frac{1}{S}\sum_{s=0}^{S-1} \frac{1}{T_s}\sum_{t=T_0+\dots+T_{s-1} }^{T_0+\dots+T_s-1} \frac{1}{K_t} \leq \frac{T \Bar{\eta}}{W_2} \phi_3
\end{gathered}\nonumber
\end{equation}

Plugging in the bounds for $\sum_{t=0}^{T-1}\mathbb{E}\left[\left\| \Delta_t\right\|^2\right]$, $\sum_{t=0}^{T-1}\mathbb{E}\left[\left\| y_t\right\|^2\right]$, and $\sum_{t=0}^{T-1}\mathbb{E}\left[\left\| d_t\right\|^2\right]$,

\begin{equation}
\begin{gathered}
\frac{1}{2} \cdot \frac{1}{S}\sum_{s=0}^{S-1} \frac{1}{T_s}\sum_{t=T_0+\dots+T_{s-1} }^{T_0+\dots+T_s-1} \left\|\nabla f(x_t)\right\|^2
\leq \frac{2\left(f(z_0) - \mathbb{E}\left[f(z_{T})\right]  \right)}{S W_2 \eta_l} + \\
\left( -\frac{\eta_S}{S W_2 m^2} + \frac{3 L^2 \tau^2 \hat{\eta}^3 \eta_l^2 C_\beta}{W_2 m^2} + \frac{L^2 W_1^2 \Bar{\eta} \eta_l C_\beta}{W_2 m^2} + \frac{\Bar{\eta}\eta_l}{W_2 m^2} + \frac{L\hat{\eta}^2\eta_l}{W_2 m^2} \right)\\
\cdot \mathbb{E}\left[\left\| \sum_{i\in\mathcal{S}_t} \frac{1}{K_{t,i}} \sum_{k=0}^{K_{t,i}-1} \nabla f_i(x_{t-\tau_{t,i},k}^i) \right\|^2\right]\\
+\left( \frac{15  \eta^2_l L^2 \phi_1 T \Bar{\eta}}{W_2}  + \frac{3 L^2 \tau^2 \hat{\eta}^3 \eta_l^2 T}{m W_2} \phi_3 + \frac{L^2W_1^2\Bar{\eta}\eta_l T}{m W_2} \phi_3 + \frac{\Bar{\eta}\eta_l T}{m W_2} \phi_3 + \frac{L \hat{\eta}^2 \eta_l T}{m W_2} \phi_3 \right)\sigma_l^2\\
\left(3 + \frac{90 \eta^2_l L^2 T \Bar{\eta}\phi_2 }{W_2}\right)\sigma_g^2
\end{gathered}\nonumber
\end{equation}

We could verify, when the following condition holds,

\begin{equation}
\begin{gathered}
3L^2 S \tau^2 \hat{\eta_0}^2 C_\eta^2 \eta_l^2 + C_\eta S \left( L^2 W_1^2 C_\eta + 1 + L \Bar{\eta} \right)  \eta_l \leq 1
\end{gathered}\nonumber
\end{equation}

we have the coefficient for $\mathbb{E}\left[\left\| \sum_{i\in\mathcal{S}_t} \frac{1}{K_{t,i}} \sum_{k=0}^{K_{t,i}-1} \nabla f_i(x_{t-\tau_{t,i},k}^i) \right\|^2\right]$, i.e.,

\begin{equation}
\begin{gathered}
 -\frac{\eta_S}{S W_2 m^2} + \frac{3 L^2 \tau^2 \hat{\eta}^3 \eta_l^2 C_\beta}{W_2 m^2} + \frac{L^2 W_1^2 \Bar{\eta} \eta_l C_\beta}{W_2 m^2} + \frac{\Bar{\eta}\eta_l}{W_2 m^2} + \frac{L\hat{\eta}^2\eta_l}{W_2 m^2} \leq 0
\end{gathered}\nonumber
\end{equation}

Therefore, we have,

\begin{equation}
\begin{gathered}
 \frac{1}{S}\sum_{s=0}^{S-1} \frac{1}{T_s}\sum_{t=T_0+\dots+T_{s-1} }^{T_0+\dots+T_s-1} \left\|\nabla f(x_t)\right\|^2
\leq \frac{4 \left(f(z_0) - f^\ast  \right)}{S W_2 \eta_l} +\\
+\left( \frac{30  \eta^2_l L^2 \phi_1 T \Bar{\eta}}{W_2}  + \frac{6 L^2 \tau^2 \hat{\eta}^3 \eta_l^2 T}{m W_2} \phi_3 + \frac{ 2 L^2 W_1^2 \Bar{\eta}\eta_l T}{m W_2} \phi_3 + \frac{2 \Bar{\eta}\eta_l T}{m W_2} \phi_3 + \frac{2 L \hat{\eta}^2 \eta_l T}{m W_2} \phi_3 \right)\sigma_l^2\\
\left(6 + \frac{180 \eta^2_l L^2 T \Bar{\eta}\phi_2 }{W_2}\right)\sigma_g^2
\end{gathered}\nonumber
\end{equation}

Suppose $S=1$, i.e. the typical constant hyperparameter regime, and further suppose local updating number as $K$, the total number of rounds as $T$, $\eta_0=\Bar{\eta}=\Theta\left(\sqrt{mK}\right)$ and $\eta_l=\Theta\left(\frac{1}{\sqrt{T}}\right)$. In this case, $\phi_1=K$, $\phi_2=K^2$, $\phi_3=\frac{1}{K}$, $W_2=\Theta\left(T\sqrt{mK}\right)$. Suppose $W_1^2=\mathcal{O}\left(\sqrt{mK}\right)$. We have the bound as,

We have the bounds as,

\begin{equation}
\begin{gathered}
 \frac{1}{S}\sum_{s=0}^{S-1} \frac{1}{T_s}\sum_{t=T_0+\dots+T_{s-1} }^{T_0+\dots+T_s-1} \left\|\nabla f(x_t)\right\|^2
\leq \mathcal{O}\left(\frac{1}{\sqrt{mKT}}\right) \left(f(z_0) - f^\ast  \right) + \\
+\left( \mathcal{O}\left(\frac{K}{T}\right) + \mathcal{O}\left(\frac{\tau^2}{T}\right) + \mathcal{O}\left(\frac{1}{mK\sqrt{T}}\right) + \mathcal{O}\left(\frac{1}{\sqrt{mKT}}\right) \right)\sigma_l^2 + \left(6 + \mathcal{O}\left( \frac{K^2}{T} \right) \right)\sigma_g^2
\end{gathered}\nonumber
\end{equation}

Only keep the dominant terms, we could get,

\begin{equation}
\begin{gathered}
 \frac{1}{S}\sum_{s=0}^{S-1} \frac{1}{T_s}\sum_{t=T_0+\dots+T_{s-1} }^{T_0+\dots+T_s-1} \left\|\nabla f(x_t)\right\|^2
\leq \mathcal{O}\left(\frac{1}{\sqrt{mKT}}\right) \left(f(z_0) - f^\ast  \right) + \\
+\left( \mathcal{O}\left(\frac{\tau^2}{T}\right) + \mathcal{O}\left(\frac{1}{\sqrt{mKT}} \right) \right)\sigma_l^2 + \left(6 + \mathcal{O}\left( \frac{K^2}{T} \right) \right)\sigma_g^2
\end{gathered}\nonumber
\end{equation}

Suppose $S=\Theta(1)$, i.e. the multistage regime, the total number of rounds are $T$, $\Bar{\eta} = \Theta\left(\sqrt{mK}\right)$, $\hat{\eta}^2 = \Theta\left(mK\right)$, $\hat{\eta}^3 = \Theta\left(m^{\frac{3}{2}} K^{\frac{3}{2}}\right)$, and  $\eta_l=\Theta\left(\frac{1}{\sqrt{T}}\right)$, $W_2=\Theta\left(\frac{T\sqrt{mK}}{S}\right)$, i.e. $T\Bar{\eta}$ is equally divided into $S$ stages, suppose $W_1^2=\mathcal{O}\left(\sqrt{mK}\right)$, we have the bound as,

\begin{equation}
\begin{gathered}
 \frac{1}{S}\sum_{s=0}^{S-1} \frac{1}{T_s}\sum_{t=T_0+\dots+T_{s-1} }^{T_0+\dots+T_s-1} \left\|\nabla f(x_t)\right\|^2
\leq \mathcal{O}\left(\frac{1}{\sqrt{mKT}}\right) \left(f(z_0) - f^\ast  \right) + \\
+\left( \mathcal{O}\left(\frac{\tau^2}{T}\right) + \mathcal{O}\left(\frac{1}{\sqrt{mKT}} \right) \right)\sigma_l^2 + \left(6 + \mathcal{O}\left( \frac{K^2}{T} \right) \right)\sigma_g^2
\end{gathered}\nonumber
\end{equation}

The bound is thus, $\mathcal{O}\left(\frac{1}{\sqrt{mKT}}\right)+ \mathcal{O}\left(\frac{\tau^2}{T}\right) +  \mathcal{O}\left( \frac{K^2}{T} \right) + \mathcal{O}\left( \sigma_g^2 \right)$.

\end{proof}

\section{Experiments}
\label{sec:appendix_exp}

\subsection{Experimental Settings}
\label{subsec:exp_settings_appendix}

We test how the performances of our proposed algorithms compared to FedAvg baseline in different settings. We train ResNet \citep{He2016DeepResNet} and VGG \citep{Simonyan14VGG} on CIFAR10 \citep{Krizhevsky2009CIFAR}. To simulate data heterogeneity in CIFAR-10, we impose label imbalance across clients, i.e. each client is allocated a proportion of the samples of each label according to a Dirichlet distribution. Same procedure has been taken by \citep{Hsu2019MeasuringTE, Yurochkin2019BayesianNF, Wang20FedNova, li2022NIIDBenchmark}. The concentration parameter $\alpha>0$ indicates the level of \textit{non-i.i.d.}, with a smaller $\alpha$ implies higher heterogeneity, and $\alpha\to\infty$ implies \textit{i.i.d.} setting. 

\textbf{How Asynchrony and Heterogeneous Local Epochs Are Implemented in Autonomous FedGM?}

To simulate the asynchrony, we allow each worker to select one global model randomly from the last recent 5 global models instead of only using the current round's model in vanilla FedAvg. To simulate the heterogeneous local epochs, we allow each worker to randomly select local epoch number from $\{1,2, \dots, 6\}$ at each round so that each worker has a time-varying, device dependent local epoch. Note in vanilla FedAvg, we fix the local epoch as 3.

Unless specified otherwise, we have the following default experimental settings,

\begin{table}[htbp]
\vskip -2pt
    \caption{Default Experimental Settings}
    \centering
    \begin{tabular}{c|c}
    \hline
    Number of Clients: 100 & Participation Ratio: 0.05\\ 
    \hline
    Concentration Parameter: $\alpha=0.5$ & Local Epoch: 3\\ 
    \hline
    Local Learning Rate: $\eta_l=0.01$ &  Total Number of Rounds: 500\\ 
    \hline
    $\eta$ Grid: $\{0.5,1.0,1.5,\dots,5.0\}$ & $\beta$ Grid: $\{0.7,0.9,0.95\}$ \\ 
    \hline
    $\nu$ Grid: $\{0.7,0.9,0.95\}$ & Local Momentum: Disabled\\ 
    \hline
    \end{tabular}
    \label{default_setting_table}
\vskip -2pt
\end{table}

\subsection{More Experiments in Section \ref{subsec:exp_fedgm}}
\label{subsec:appendix_more_results_fedgm}

\textbf{Different Model Architecture and Levels of Heterogeneity}

\begin{figure}[h]
\vspace*{-6pt}
\centering
\subfigure{
\hspace{0pt}
\includegraphics[width=.4\textwidth]{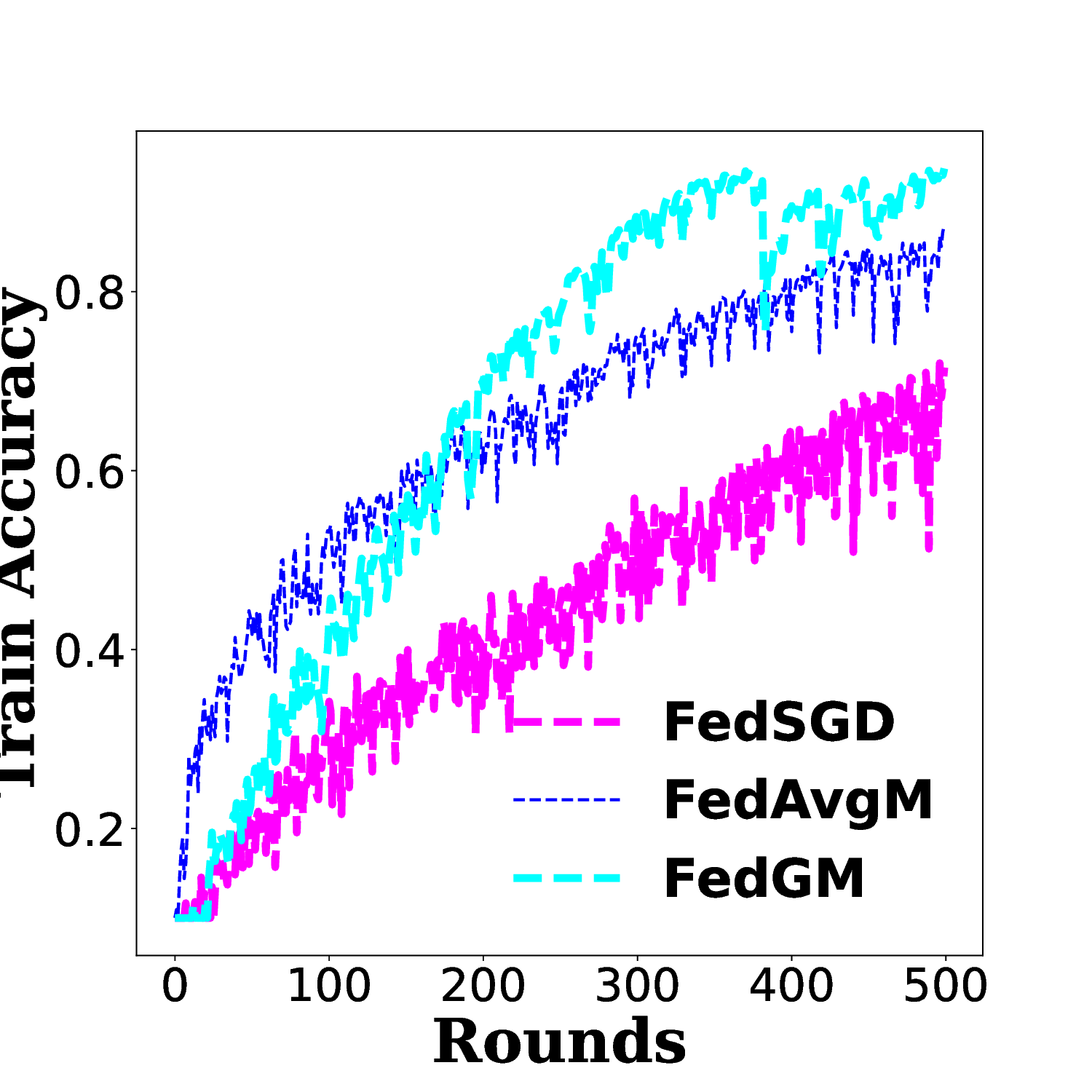}
\label{subfig:vgg_cifar10_train}
}
\subfigure{
\hspace{0pt}
\includegraphics[width=.4\textwidth]{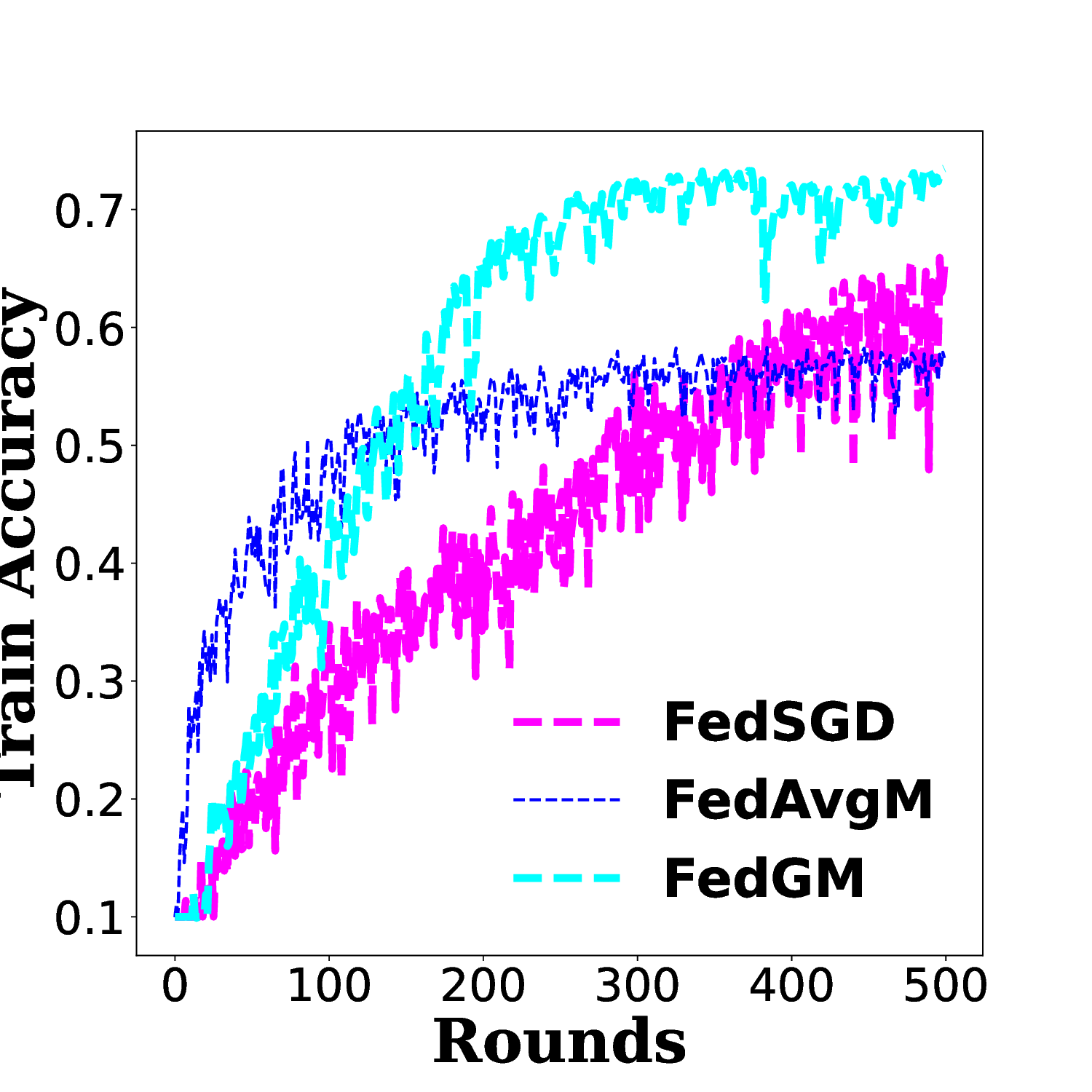}
\label{subfig:vgg_cifar10_test}
}
\vspace*{-6pt}
\caption{\ref{subfig:vgg_cifar10_train} Training and \ref{subfig:vgg_cifar10_test} Testing Curve for VGG on CIFAR-10.}
\label{fig:cifar10_vgg_result}
\end{figure}

Figure \ref{fig:cifar10_vgg_result} shows the results for VGG on CIFAR-10 with FedGM, FedAvgM, and FedAvg. We perform grid search over $\eta\in\{0.5,1.0,1.5,\dots,5.0\}$, $\beta\in\{0.7,0.9,0.95\}$, and $\nu\in\{0.7,0.9,0.95\}$. We report the curves with best final test accuracy after 500 rounds. We could observe FedGM outperforms FedAvgM and FedAvg in both training and testing, which again verifies our claim that general momentum is a more capable algorithm compared to FedAvgM.

\begin{figure}[h]
\vspace*{-6pt}
\centering
\subfigure{
\hspace{0pt}
\includegraphics[width=.4\textwidth]{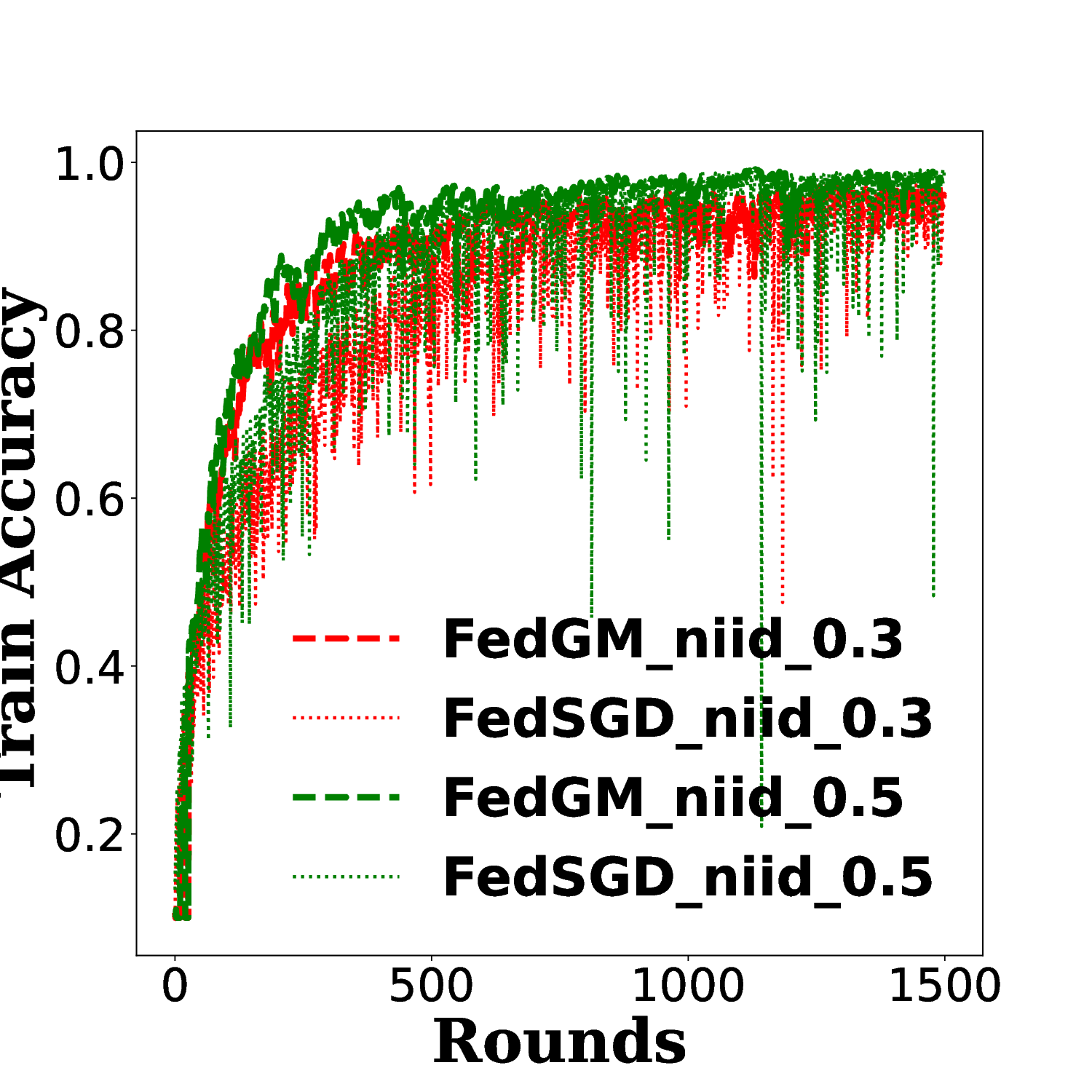}
\label{subfig:resnet_cifar10_train_various_iid}
}
\subfigure{
\hspace{0pt}
\includegraphics[width=.4\textwidth]{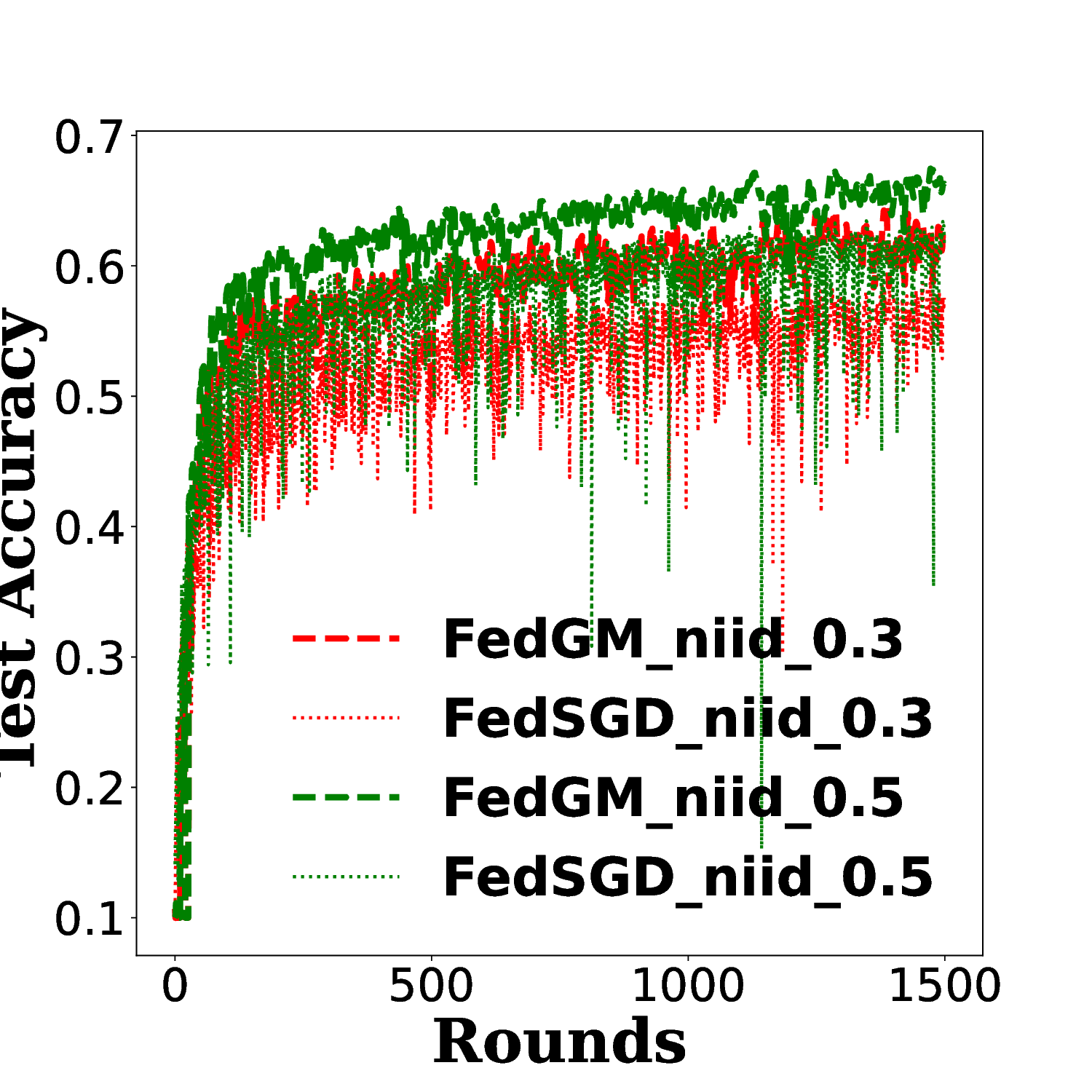}
\label{subfig:resnet_cifar10_test_various_iid}
}
\vspace*{-6pt}
\caption{\ref{subfig:resnet_cifar10_train_various_iid} Training and \ref{subfig:resnet_cifar10_test_various_iid} Testing Curve for ResNet on CIFAR-10 with Various Levels of Heterogeneity}
\label{fig:cifar10_resnet_various_niid_result}
\end{figure}

Figure \ref{fig:cifar10_resnet_various_niid_result} shows the results for ResNet on CIFAR-10 with FedGM and FedAvg with different concentration parameters $\alpha=0.3$ and $\alpha=0.5$ (i.e. \textit{non-i.i.d.}). We perform a similar grid search as in Section \ref{subsec:exp_fedgm}. We could observe the superiority of FedGM compared to FedAvg is consistent with different levels of \textit{non i.i.d.}.

\textbf{Verifying Remark \ref{remark:full_participation_why_momentum_helps}}

Remark \ref{remark:full_participation_why_momentum_helps} hypothesizes FedGM could converge with a large $\eta$ while FedAvg would diverge easily with an only moderately large server learning rate. The reason is that $\eta$ acts like a multiplier to client learning rate $\eta_l$ in FedAvg, while in FedGM, $\beta$ and $\nu$ act as a buffer that could absorb the impact from a large $\eta$. We verify this remark here.

\begin{figure}[h]
\vspace*{-6pt}
\centering
\subfigure{
\hspace{0pt}
\includegraphics[width=.35\textwidth]{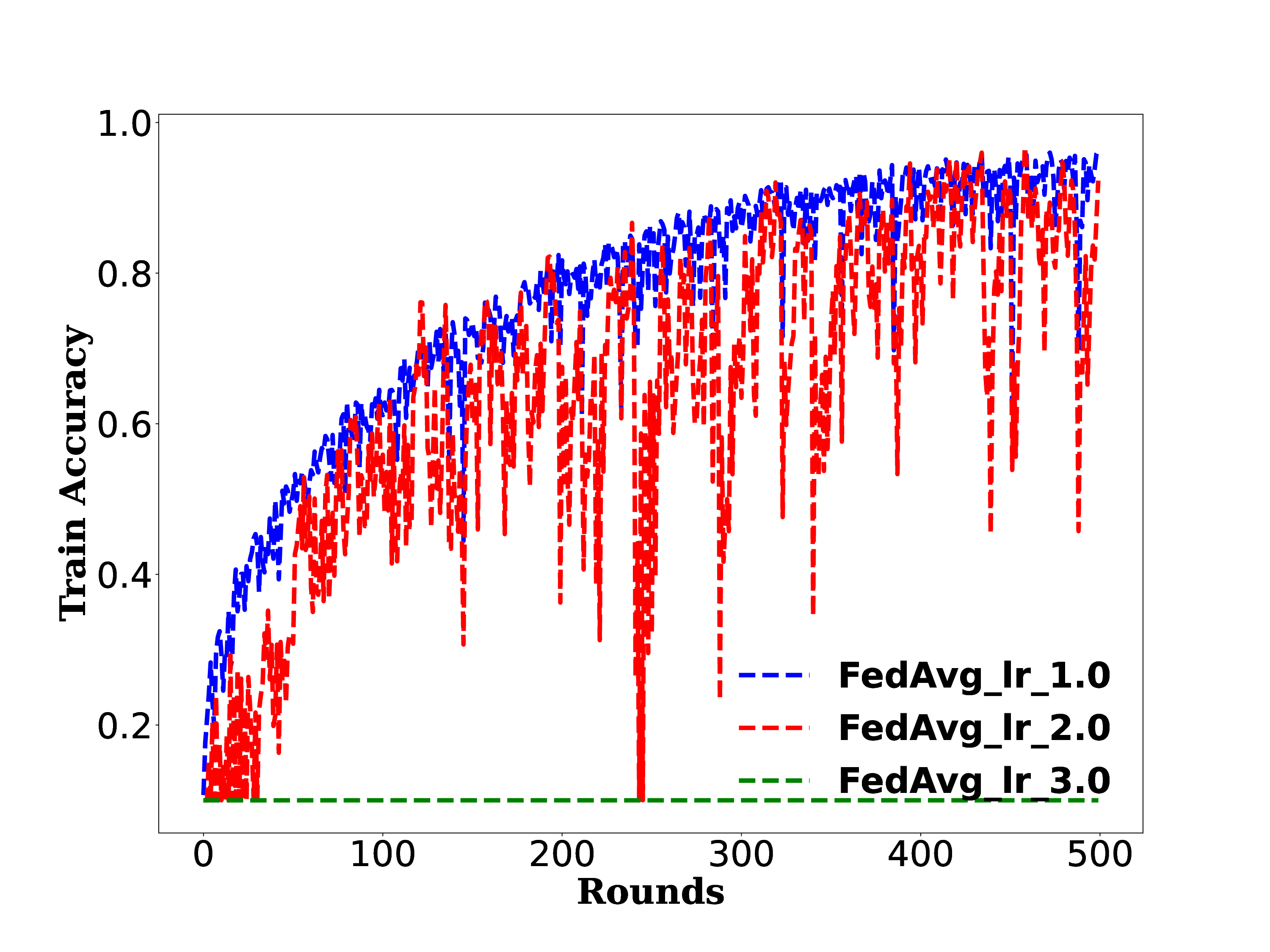}
\label{subfig:fedavg_various_lr_train}
}
\subfigure{
\hspace{0pt}
\includegraphics[width=.35\textwidth]{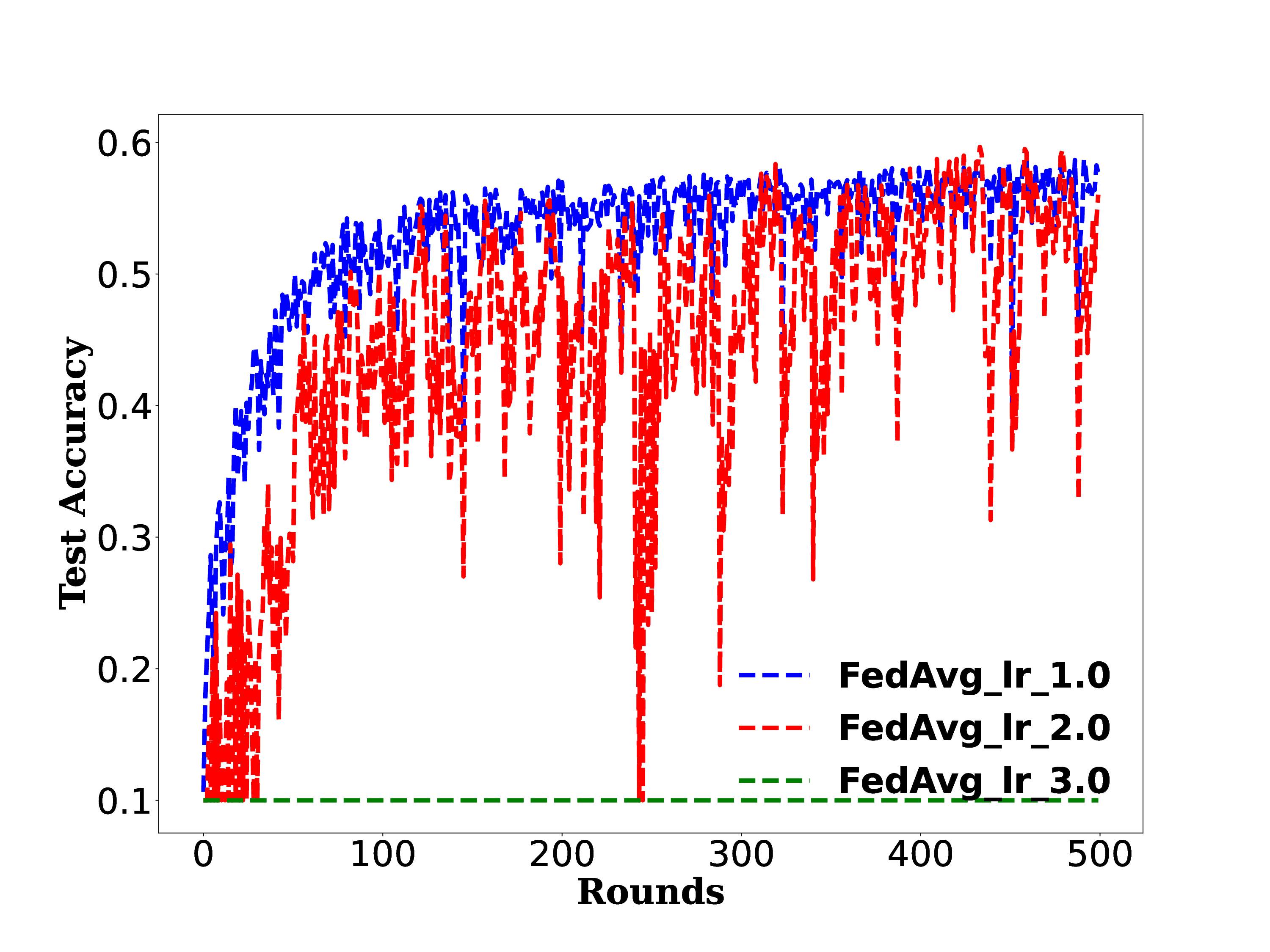}
\label{subfig:fedavg_various_lr_test}
}
\vspace*{-6pt}
\caption{\ref{subfig:fedavg_various_lr_train} Training and \ref{subfig:fedavg_various_lr_test} Testing Curve for FedAvg with various server learning rates $\eta$.}
\label{fig:fedavg_various_lr_result}
\end{figure}

Figure \ref{fig:fedavg_various_lr_result} shows the results for ResNet on CIFAR-10 with FedAvg but different learning rates $\eta=1.0$, $\eta=2.0$, and $\eta=3.0$. We could see FedAvg experiences an unstable convergence even when $\eta=2.0$ and completely divergent when $\eta=3.0$.

\begin{figure}[h]
\vspace*{-6pt}
\centering
\subfigure{
\hspace{0pt}
\includegraphics[width=.35\textwidth]{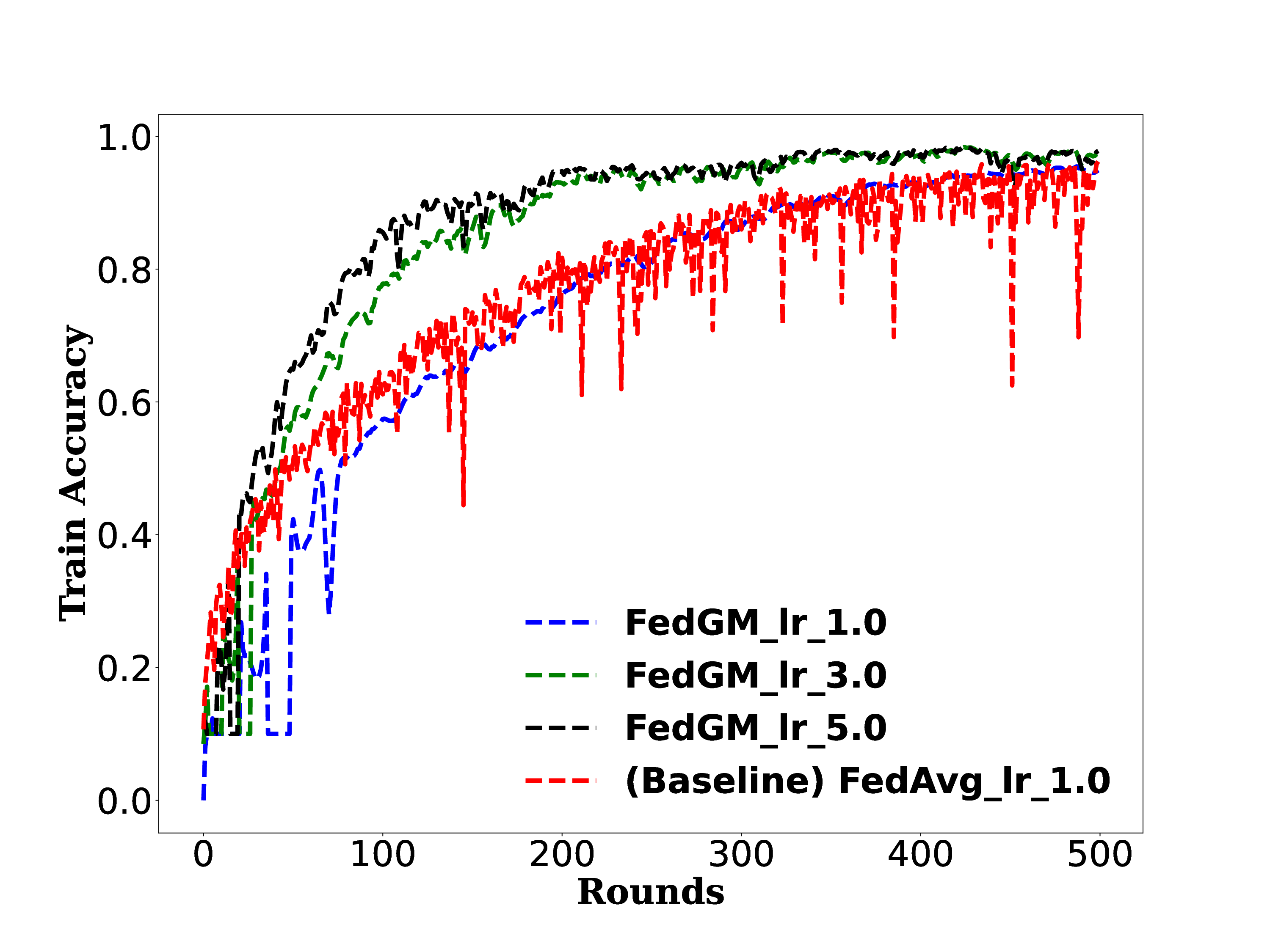}
\label{subfig:fedgm_various_lr_train}
}
\subfigure{
\hspace{0pt}
\includegraphics[width=.35\textwidth]{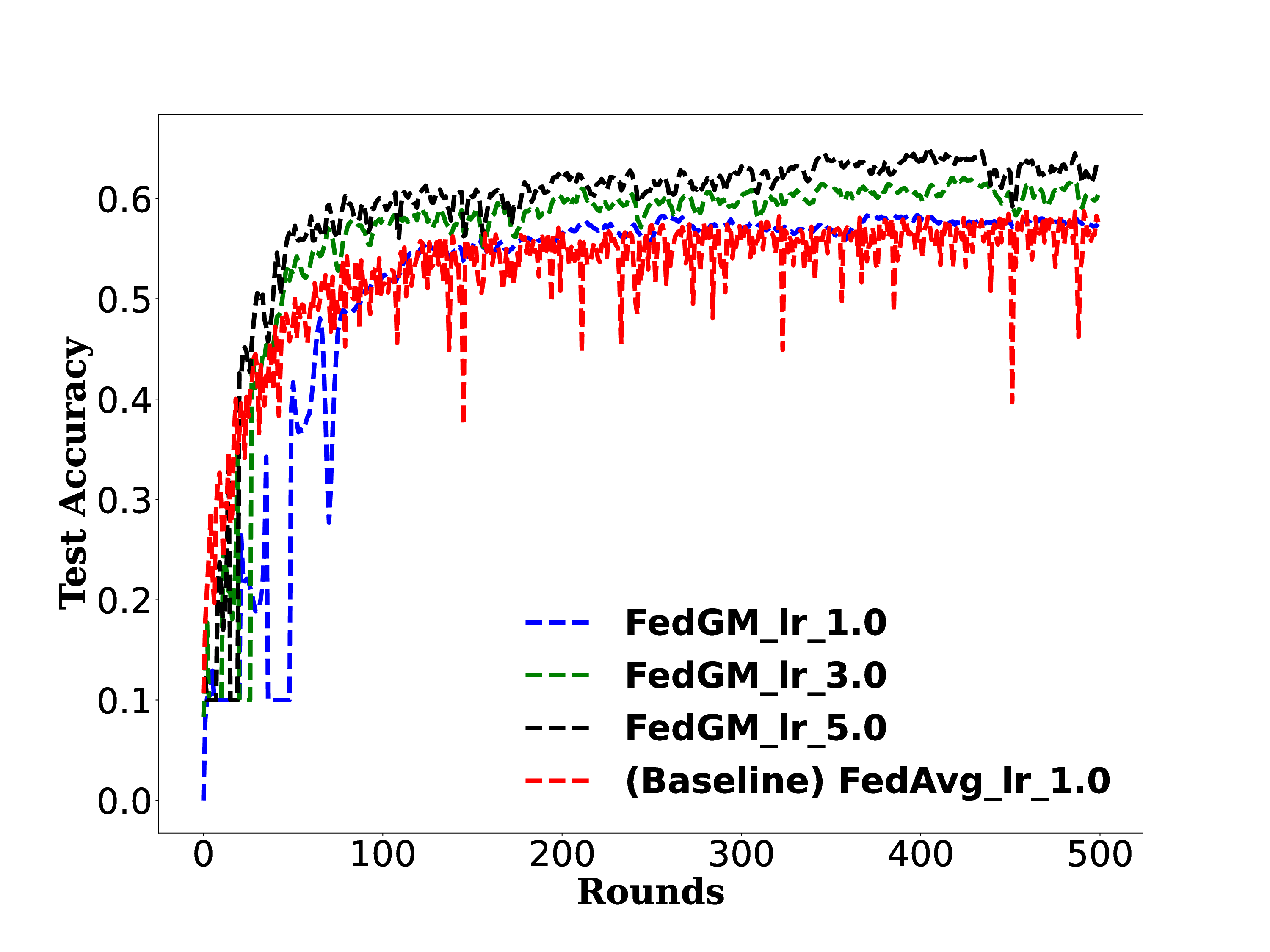}
\label{subfig:fedgm_various_lr_test}
}
\vspace*{-6pt}
\caption{\ref{subfig:fedgm_various_lr_train} Training and \ref{subfig:fedgm_various_lr_test} Testing Curve for FedGM with various server learning rates $\eta$.}
\label{fig:fedgm_various_lr_result}
\end{figure}

Figure \ref{fig:fedgm_various_lr_result} shows the results for FedGM but different learning rates $\eta=1.0$, $\eta=3.0$, and $\eta=5.0$. All experimental settings are identical to Figure \ref{fig:fedavg_various_lr_result} except for the difference between FedAvg and FedGM. We could see FedGM sustains a much larger $\eta$ compared to FedAvg. It could converge and even accelerate with $\eta=5.0$ compared to FedAvg baseline.

\subsection{More Experiments in Section \ref{subsec:exp_multistage_fedgm}}
\label{subsec:more_exp_multistage_appendix}

\begin{figure}[h]
\vspace*{-6pt}
\centering
\subfigure{
\hspace{0pt}
\includegraphics[width=.42\textwidth]{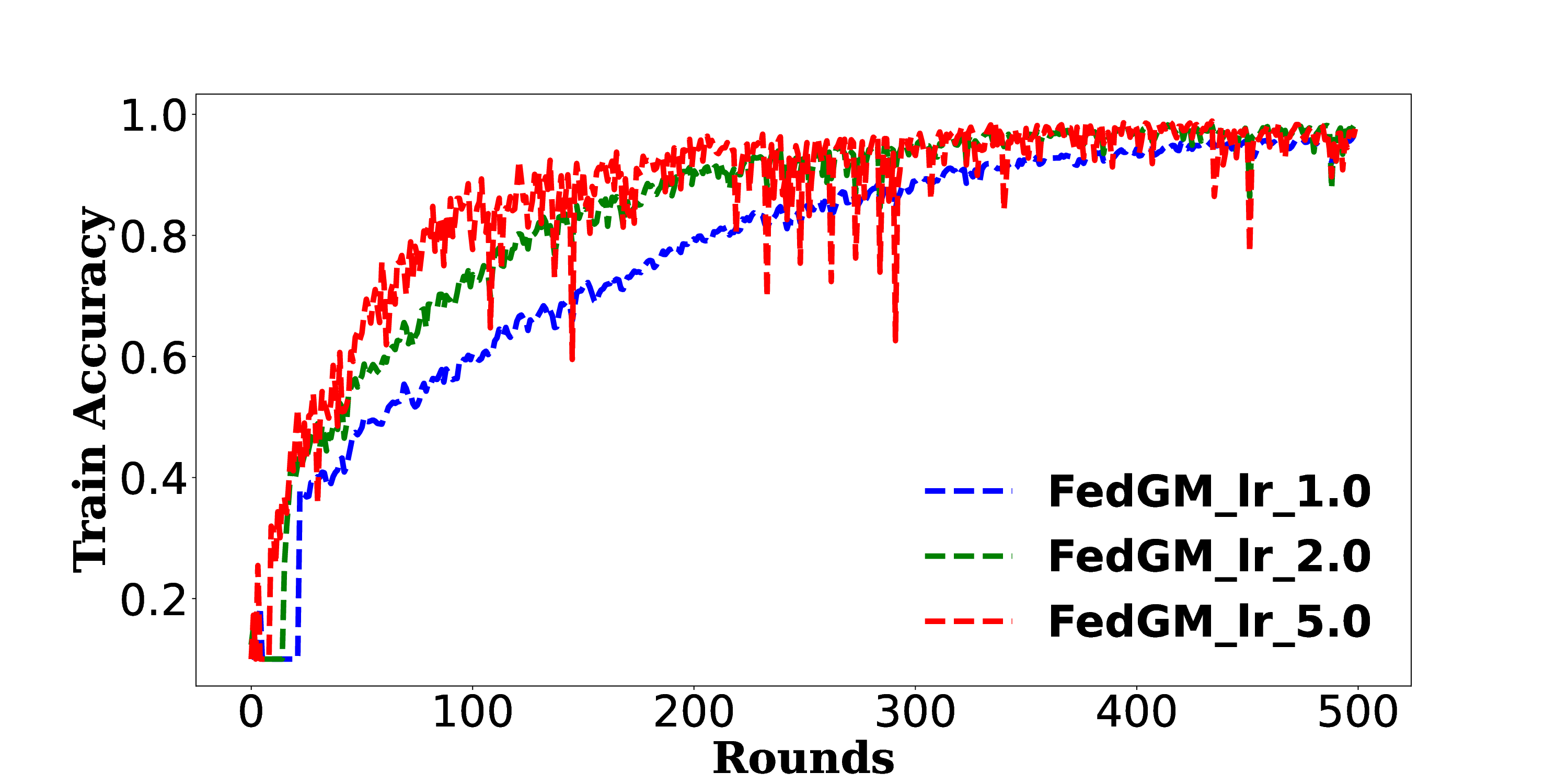}
\label{subfig:fedgm_various_lr_initial_train}
}
\subfigure{
\hspace{0pt}
\includegraphics[width=.42\textwidth]{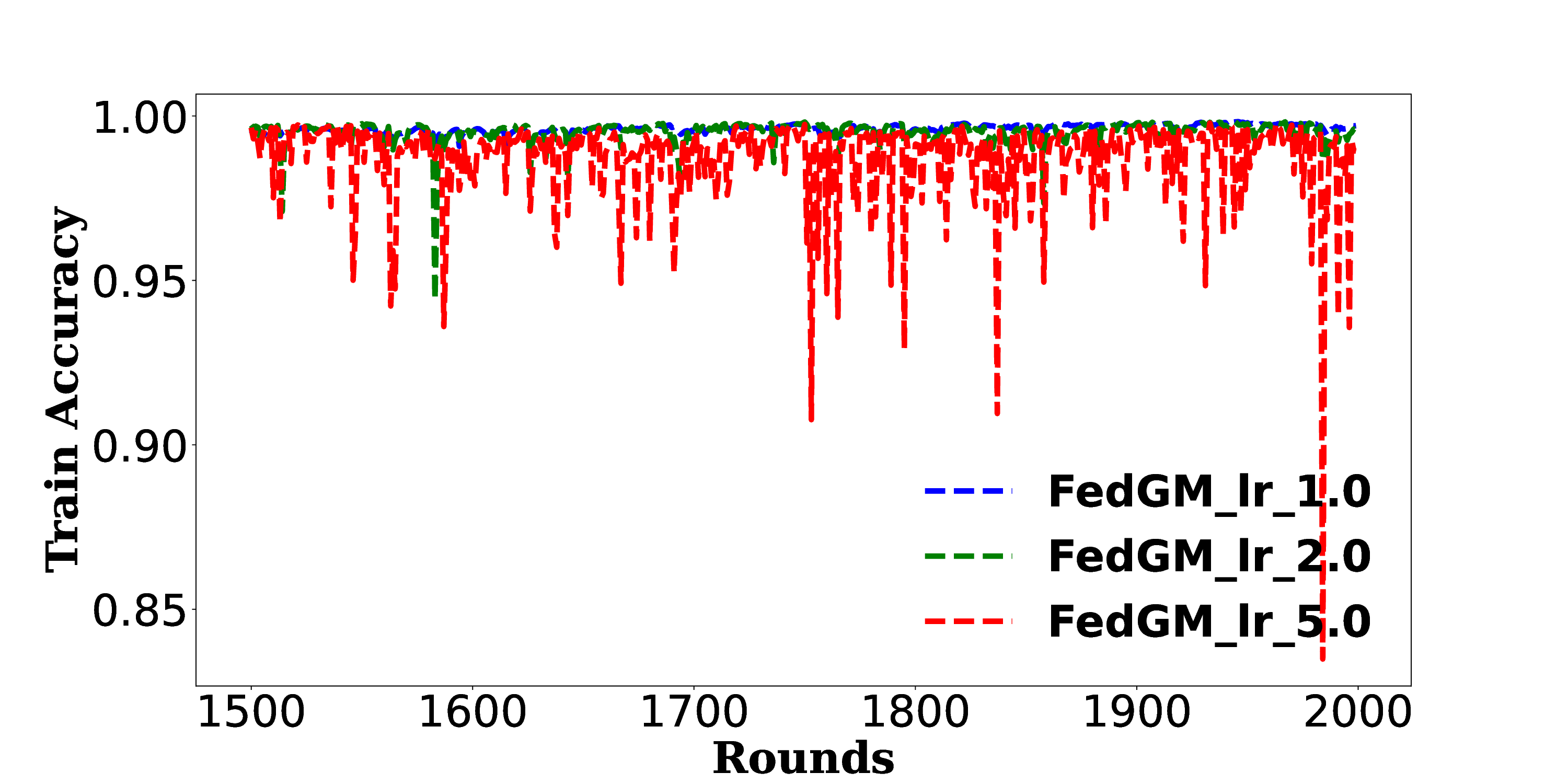}
\label{subfig:fedgm_various_lr_late_train}
}
\vspace*{-6pt}
\caption{Training Curves for FedGM with various server learning rates $\eta$. \ref{subfig:fedgm_various_lr_initial_train} the first 500 rounds; \ref{subfig:fedgm_various_lr_late_train} the last 500 rounds.}
\label{fig:fedgm_various_lr_early_late_result}
\end{figure}

Figure \ref{fig:fedgm_various_lr_early_late_result} motivates our multistage FedGM. We run ResNet on CIFAR-10 with FedGM but different learning rates $\eta=1.0$, $\eta=2.0$, and $\eta=5.0$ for 2000 rounds. We fix $\beta=\nu=0.95$ for expository purpose. We could see in early rounds (i.e. the first 500 rounds), $\eta=5.0$ has advantages that it converges faster than small $\eta=1.0$. However, $\eta=1.0$ is much more stable than  $\eta=5.0$ in the last 500 rounds when they all get nearly perfect training accuracy. This is consistent with the motivation of multistage FedGM, i.e. large initial $\eta$ benefits exploration, while small later $\eta$ benefits exploitation, and multistage scheduler obtains a balance.

\begin{figure}[h]
\vspace*{-6pt}
\centering
\subfigure{
\hspace{0pt}
\includegraphics[width=.45\textwidth]{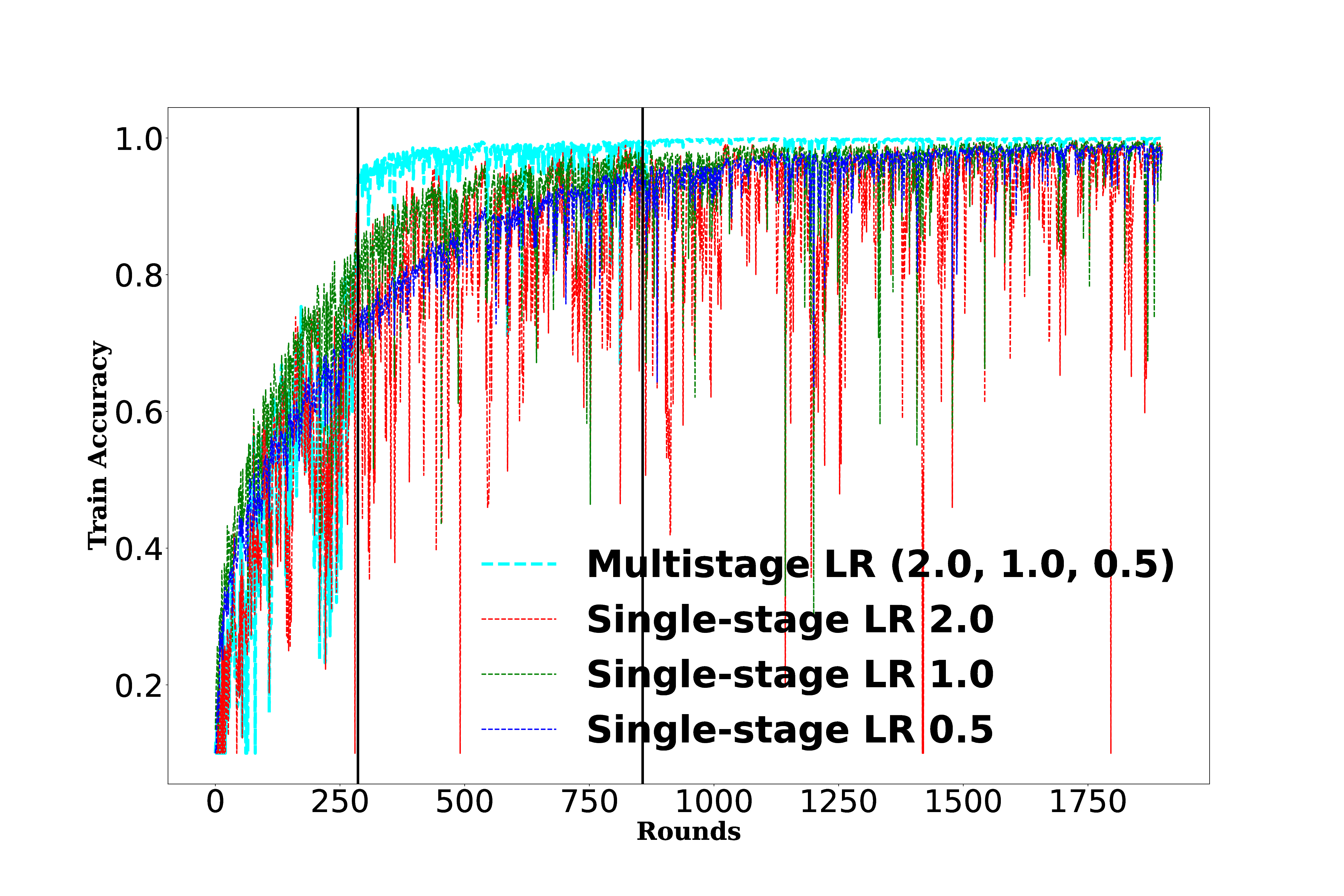}
\label{subfig:2000r_multistage_resnet_cifar10_train}
}
\subfigure{
\hspace{0pt}
\includegraphics[width=.45\textwidth]{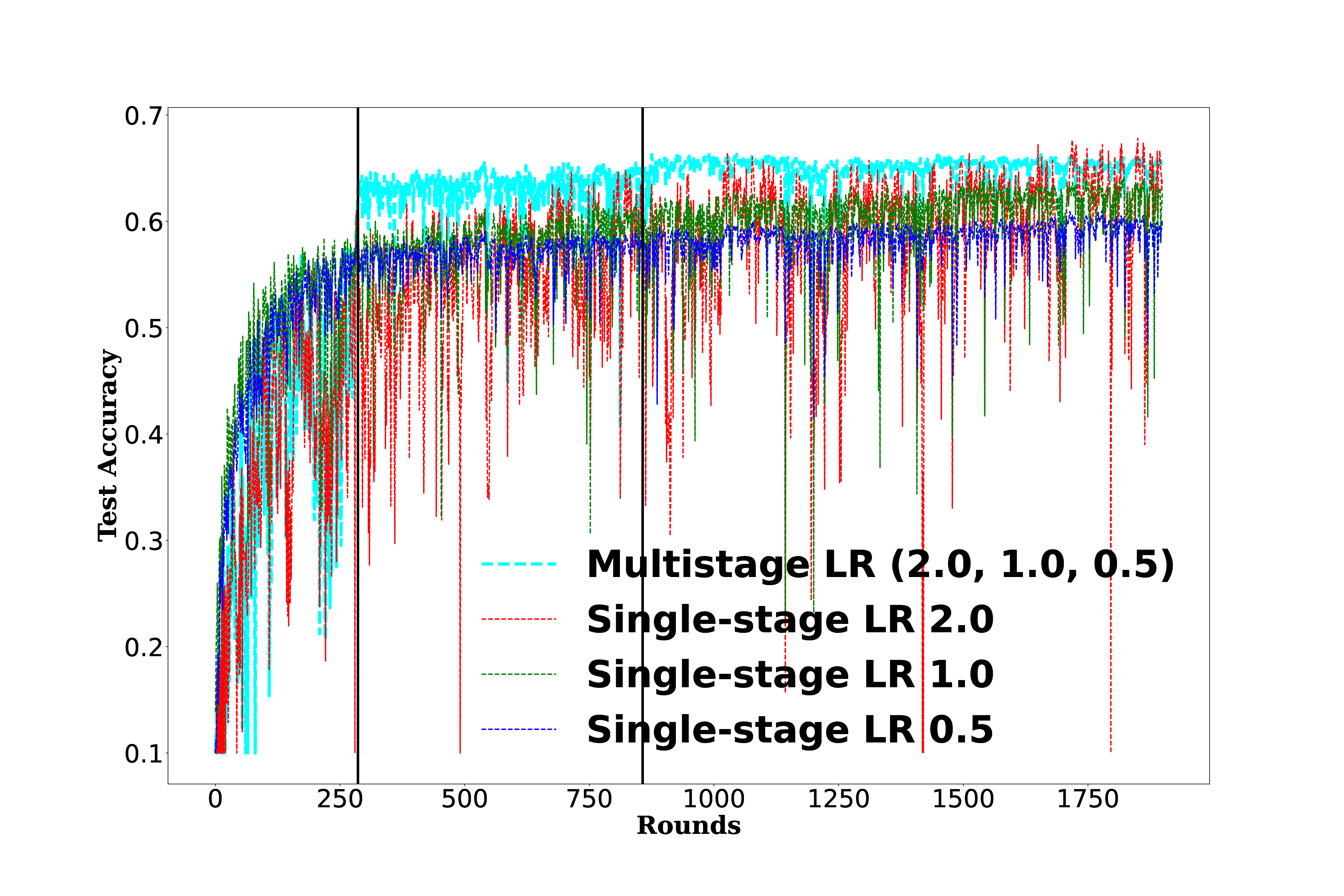}
\label{subfig:2000r_multistage_resnet_cifar10_test}
}
\vspace*{-6pt}
\caption{\ref{subfig:2000r_multistage_resnet_cifar10_train} Training and \ref{subfig:2000r_multistage_resnet_cifar10_test} Testing Curves for Multistage FedGM vs. Single-stage FedGM for 2000 rounds.}
\label{fig:2000r_multistage_resnet_cifar10}
\end{figure}

Figure \ref{fig:2000r_multistage_resnet_cifar10} presents the results of running multistage FedGM for 2000 rounds, to see whether the advantage of multistage disappears with a longer training time. The two black vertical lines at round 286 and 857 mark the end of 1st/2nd stage. As we could observe from Figure \ref{fig:2000r_multistage_resnet_cifar10}, the superiority of multistage FedGM is consistent with longer training time.

\subsection{More Experiments in Section \ref{subsec:exp_autonomous_fedgm}}
\label{subsec:appendix_more_exp_autonomous}

\begin{figure}[h]
\vspace*{-12pt}
\centering
\subfigure{
\hspace{0pt}
\includegraphics[width=.35\textwidth]{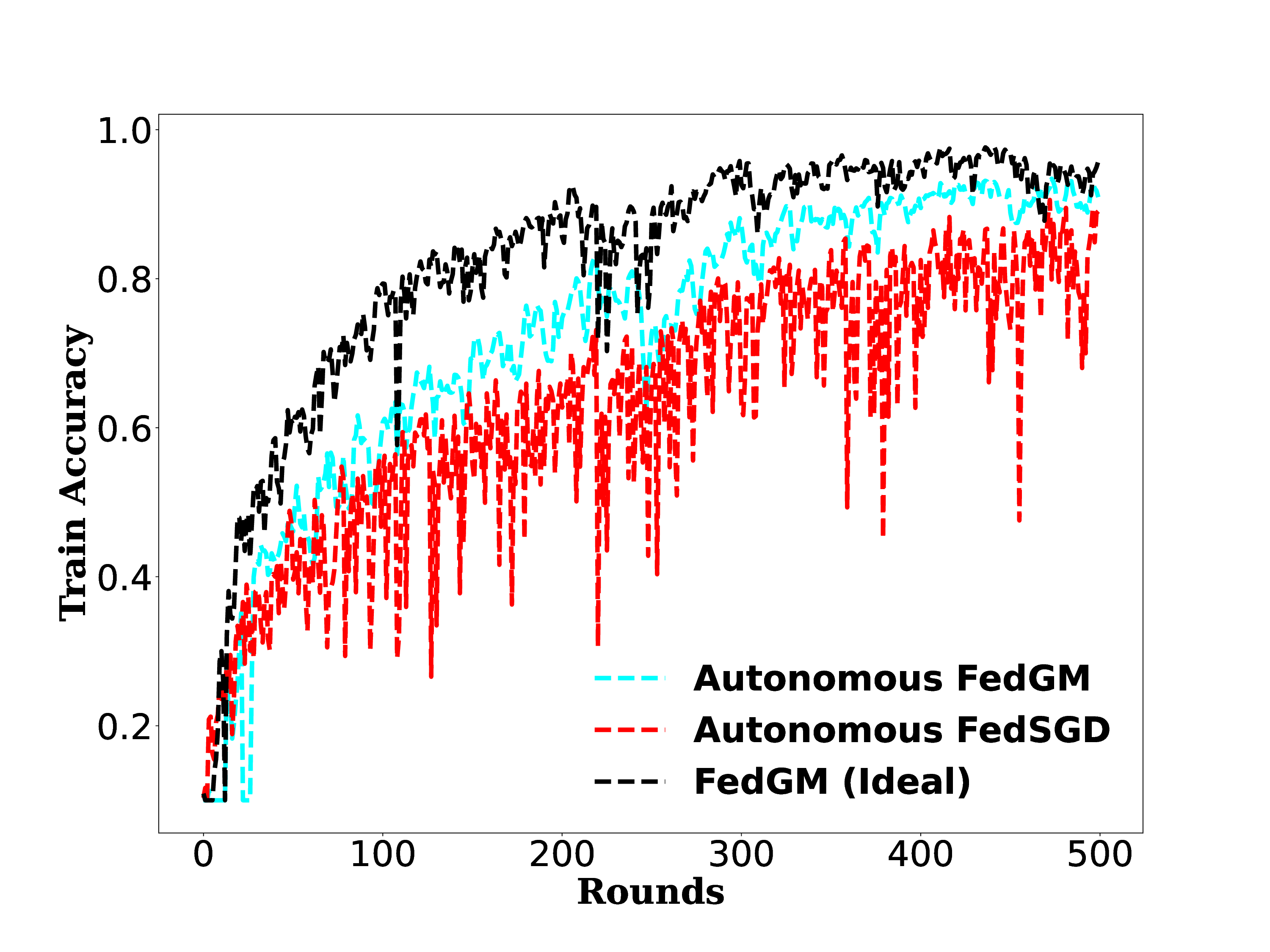}
\label{subfig:autonomous_varying_delay_train}
}
\subfigure{
\hspace{0pt}
\includegraphics[width=.35\textwidth]{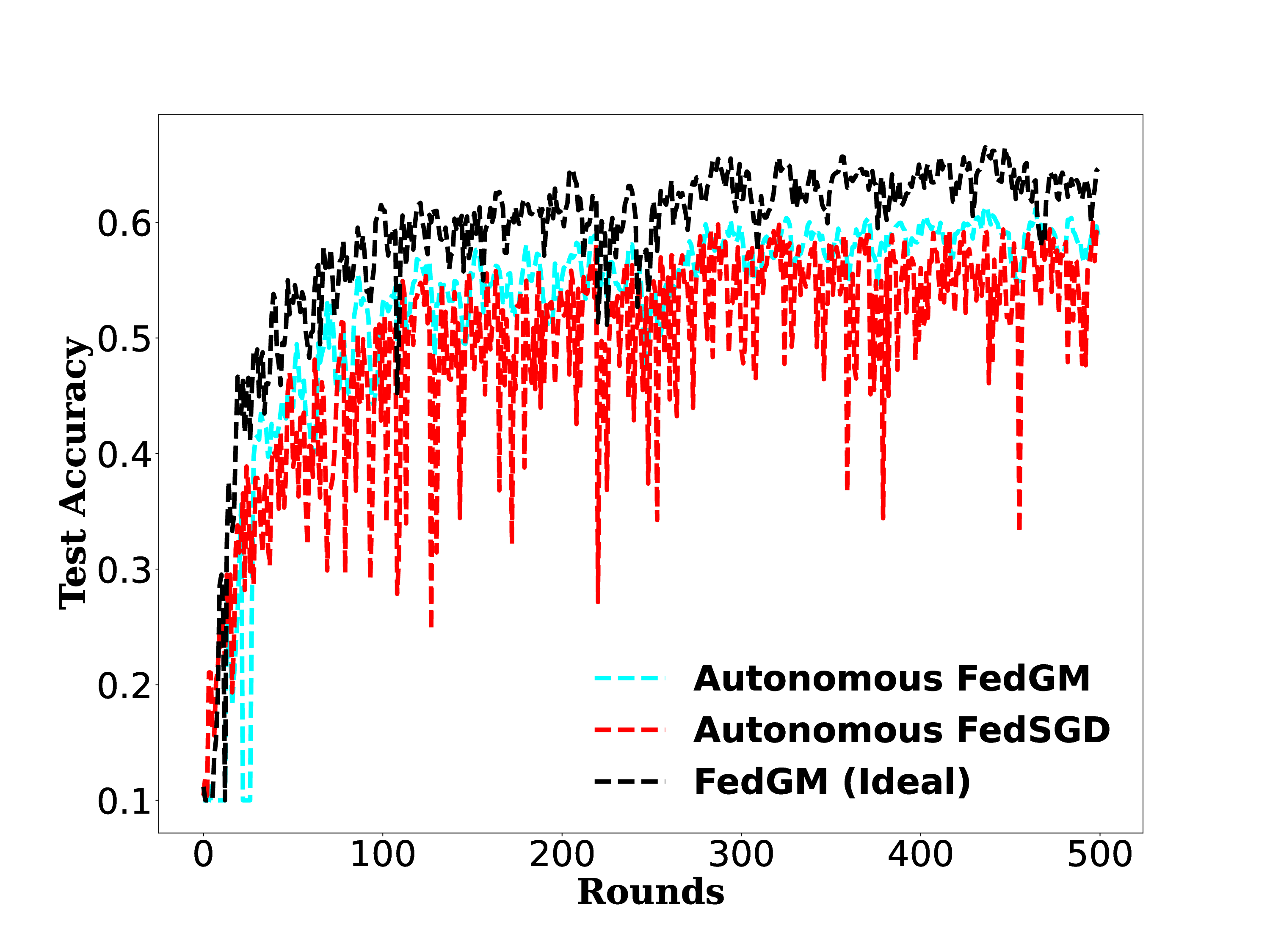}
\label{subfig:autonomous_varying_delay_test}
}
\vspace*{-6pt}
\caption{\ref{subfig:autonomous_varying_delay_train} Training and \ref{subfig:autonomous_varying_delay_test} Testing Curve for ResNet on CIFAR-10 with Random Delay = 10.}
\label{fig:autonomous_varying_delay_result}
\end{figure}

Figure \ref{fig:autonomous_varying_delay_result} shows the results for ResNet on CIFAR-10 with Autonomous FedGM and Autonomous FedAvg. The experimental settings are exactly same as Figure \ref{fig:resnet_cifar10} except the random delay is 10 instead of 5. Specifically, in  Figure \ref{fig:resnet_cifar10_autonomous} we allow each worker to select one global model randomly from the last recent 5 global models, while in Figure \ref{fig:resnet_cifar10} we allow each worker to select one global model randomly from the last recent 10 global models. The objective is to mimic different levels of asynchrony. We report the curves with best final test accuracy. We plot a FedGM (i.e. no random delay and identical local epochs) as a baseline. Similarly as Figure \ref{fig:resnet_cifar10}, we observe momentum is crucial as Autonomous FedGM outperforms Autonomous FedAvg with system heterogeneity. Autonomous FedGM does experience a slowdown compared to the ideal FedGM, but the difference is within a manageable margin, which validates the effectiveness of our proposed Autonomous FedGM.

%% file: main.bbl
\begin{thebibliography}{81}
\providecommand{\natexlab}[1]{#1}

\bibitem[{{An} et~al.(2018){An}, {Wang}, {Sun}, {Xu}, {Dai}, and {Zhang}}]{An18PID}
{An}, W.; {Wang}, H.; {Sun}, Q.; {Xu}, J.; {Dai}, Q.; and {Zhang}, L. 2018.
\newblock A PID Controller Approach for Stochastic Optimization of Deep Networks.
\newblock In \emph{2018 IEEE/CVF Conference on Computer Vision and Pattern Recognition}, 8522--8531.

\bibitem[{Avdiukhin and Kasiviswanathan(2021)}]{Avdiukhin21arbitrarycommunication}
Avdiukhin, D.; and Kasiviswanathan, S. 2021.
\newblock Federated Learning under Arbitrary Communication Patterns.
\newblock In Meila, M.; and Zhang, T., eds., \emph{Proceedings of the 38th International Conference on Machine Learning}, volume 139 of \emph{Proceedings of Machine Learning Research}, 425--435. PMLR.

\bibitem[{Bao, Gu, and Huang(2020)}]{bao2020fast}
Bao, R.; Gu, B.; and Huang, H. 2020.
\newblock Fast OSCAR and OWL regression via safe screening rules.
\newblock In \emph{International Conference on Machine Learning}, 653--663. PMLR.

\bibitem[{Bao et~al.(2022)Bao, Wu, Xian, and Huang}]{bao2022doubly}
Bao, R.; Wu, X.; Xian, W.; and Huang, H. 2022.
\newblock Doubly sparse asynchronous learning for stochastic composite optimization.
\newblock In \emph{Proceedings of the Thirty-First International Joint Conference on Artificial Intelligence, IJCAI}, 1916--1922.

\bibitem[{Bao et~al.(2023)Bao, Wei, Wang, and He}]{bao2023adaptive}
Bao, W.; Wei, T.; Wang, H.; and He, J. 2023.
\newblock Adaptive Test-Time Personalization for Federated Learning.
\newblock In \emph{Thirty-seventh Conference on Neural Information Processing Systems}.

\bibitem[{Basu et~al.(2019)Basu, Data, Karakus, and Diggavi}]{Basu19Qsparse-Local-SGD}
Basu, D.; Data, D.; Karakus, C.; and Diggavi, S. 2019.
\newblock \emph{Qsparse-Local-SGD: Distributed SGD with Quantization, Sparsification, and Local Computations}.
\newblock Red Hook, NY, USA: Curran Associates Inc.

\bibitem[{BUKATY(2019)}]{California_Consumer_Privacy_Act_CCPA}
BUKATY, P. 2019.
\newblock \emph{The California Consumer Privacy Act (CCPA): An implementation guide}.
\newblock IT Governance Publishing.
\newblock ISBN 9781787781320.

\bibitem[{Chen, Horv{\'a}th, and Richt{\'a}rik(2020)}]{Chen2020ClientSampling}
Chen, W.; Horv{\'a}th, S.; and Richt{\'a}rik, P. 2020.
\newblock Optimal Client Sampling for Federated Learning.
\newblock \emph{ArXiv}, abs/2010.13723.

\bibitem[{Cheng et~al.(2016)Cheng, Koc, Harmsen, Shaked, Chandra, Aradhye, Anderson, Corrado, Chai, Ispir, Anil, Haque, Hong, Jain, Liu, and Shah}]{google16deep&wide}
Cheng, H.-T.; Koc, L.; Harmsen, J.; Shaked, T.; Chandra, T.; Aradhye, H.; Anderson, G.; Corrado, G.; Chai, W.; Ispir, M.; Anil, R.; Haque, Z.; Hong, L.; Jain, V.; Liu, X.; and Shah, H. 2016.
\newblock Wide \& Deep Learning for Recommender Systems.
\newblock In \emph{Proceedings of the 1st Workshop on Deep Learning for Recommender Systems}, DLRS 2016, 7–10. New York, NY, USA: Association for Computing Machinery.
\newblock ISBN 9781450347952.

\bibitem[{Cutkosky and Mehta(2020)}]{Cutkosky2020MomentumIN}
Cutkosky, A.; and Mehta, H. 2020.
\newblock Momentum Improves Normalized SGD.
\newblock In \emph{International Conference on Machine Learning}.

\bibitem[{Devlin et~al.(2018)Devlin, Chang, Lee, and Toutanova}]{devlin2018bert}
Devlin, J.; Chang, M.-W.; Lee, K.; and Toutanova, K. 2018.
\newblock Bert: Pre-training of deep bidirectional transformers for language understanding.
\newblock \emph{arXiv preprint arXiv:1810.04805}.

\bibitem[{Devlin et~al.(2019)Devlin, Chang, Lee, and Toutanova}]{Devlin2019BERT}
Devlin, J.; Chang, M.-W.; Lee, K.; and Toutanova, K. 2019.
\newblock BERT: Pre-training of Deep Bidirectional Transformers for Language Understanding.
\newblock \emph{ArXiv}, abs/1810.04805.

\bibitem[{Dosovitskiy et~al.(2021)Dosovitskiy, Beyer, Kolesnikov, Weissenborn, Zhai, Unterthiner, Dehghani, Minderer, Heigold, Gelly, Uszkoreit, and Houlsby}]{dosovitskiy2021VIT}
Dosovitskiy, A.; Beyer, L.; Kolesnikov, A.; Weissenborn, D.; Zhai, X.; Unterthiner, T.; Dehghani, M.; Minderer, M.; Heigold, G.; Gelly, S.; Uszkoreit, J.; and Houlsby, N. 2021.
\newblock An Image is Worth 16x16 Words: Transformers for Image Recognition at Scale.
\newblock In \emph{International Conference on Learning Representations}.

\bibitem[{{European Commission}(2016)}]{european_commission_regulation_2016}
{European Commission}. 2016.
\newblock Regulation ({EU}) 2016/679 of the {European} {Parliament} and of the {Council} of 27 {April} 2016 on the protection of natural persons with regard to the processing of personal data and on the free movement of such data, and repealing {Directive} 95/46/{EC} ({General} {Data} {Protection} {Regulation}) ({Text} with {EEA} relevance).

\bibitem[{Fallah, Mokhtari, and Ozdaglar(2020)}]{fallah2020personalized}
Fallah, A.; Mokhtari, A.; and Ozdaglar, A. 2020.
\newblock Personalized federated learning: A meta-learning approach.
\newblock \emph{arXiv preprint arXiv:2002.07948}.

\bibitem[{Ge et~al.(2019{\natexlab{a}})Ge, Kakade, Kidambi, and Netrapalli}]{Ge2019TheSD}
Ge, R.; Kakade, S.~M.; Kidambi, R.; and Netrapalli, P. 2019{\natexlab{a}}.
\newblock The Step Decay Schedule: A Near Optimal, Geometrically Decaying Learning Rate Procedure.
\newblock In \emph{NeurIPS}.

\bibitem[{Ge et~al.(2019{\natexlab{b}})Ge, Kakade, Kidambi, and Netrapalli}]{ge19stepdecay}
Ge, R.; Kakade, S.~M.; Kidambi, R.; and Netrapalli, P. 2019{\natexlab{b}}.
\newblock \emph{The Step Decay Schedule: A near Optimal, Geometrically Decaying Learning Rate Procedure for Least Squares}.
\newblock Red Hook, NY, USA: Curran Associates Inc.

\bibitem[{Goetz et~al.(2019)Goetz, Malik, Bui, Moon, Liu, and Kumar}]{Goetz2019ActiveFL}
Goetz, J.; Malik, K.; Bui, D.~V.; Moon, S.; Liu, H.; and Kumar, A. 2019.
\newblock Active Federated Learning.
\newblock \emph{ArXiv}, abs/1909.12641.

\bibitem[{Goyal et~al.(2017)Goyal, Doll{\'{a}}r, Girshick, Noordhuis, Wesolowski, Kyrola, Tulloch, Jia, and He}]{GoyalDGNWKTJH17LargeMinibatch}
Goyal, P.; Doll{\'{a}}r, P.; Girshick, R.~B.; Noordhuis, P.; Wesolowski, L.; Kyrola, A.; Tulloch, A.; Jia, Y.; and He, K. 2017.
\newblock Accurate, Large Minibatch {SGD:} Training ImageNet in 1 Hour.
\newblock \emph{CoRR}, abs/1706.02677.

\bibitem[{Gu et~al.(2021)Gu, Huang, Zhang, and Huang}]{gu2021arbitraryunavailable}
Gu, X.; Huang, K.; Zhang, J.; and Huang, L. 2021.
\newblock Fast Federated Learning in the Presence of Arbitrary Device Unavailability.
\newblock In Beygelzimer, A.; Dauphin, Y.; Liang, P.; and Vaughan, J.~W., eds., \emph{Advances in Neural Information Processing Systems}.

\bibitem[{Hamilton, Ying, and Leskovec(2017)}]{Jure2017GNN}
Hamilton, W.; Ying, Z.; and Leskovec, J. 2017.
\newblock Inductive Representation Learning on Large Graphs.
\newblock In Guyon, I.; Luxburg, U.~V.; Bengio, S.; Wallach, H.; Fergus, R.; Vishwanathan, S.; and Garnett, R., eds., \emph{Advances in Neural Information Processing Systems}, volume~30. Curran Associates, Inc.

\bibitem[{He, Liu, and Tao(2019)}]{He19ControlBatch}
He, F.; Liu, T.; and Tao, D. 2019.
\newblock Control Batch Size and Learning Rate to Generalize Well: Theoretical and Empirical Evidence.
\newblock In \emph{Advances in Neural Information Processing Systems 32}, 1143--1152. Curran Associates, Inc.

\bibitem[{{He} et~al.(2016){He}, {Zhang}, {Ren}, and {Sun}}]{He16Res}
{He}, K.; {Zhang}, X.; {Ren}, S.; and {Sun}, J. 2016.
\newblock Deep Residual Learning for Image Recognition.
\newblock In \emph{2016 IEEE Conference on Computer Vision and Pattern Recognition (CVPR)}, 770--778.

\bibitem[{He et~al.(2016)He, Zhang, Ren, and Sun}]{He2016DeepResNet}
He, K.; Zhang, X.; Ren, S.; and Sun, J. 2016.
\newblock Deep Residual Learning for Image Recognition.
\newblock \emph{2016 IEEE Conference on Computer Vision and Pattern Recognition (CVPR)}, 770--778.

\bibitem[{Hsu, Qi, and Brown(2019)}]{Hsu2019MeasuringTE}
Hsu, T.-M.~H.; Qi; and Brown, M. 2019.
\newblock Measuring the Effects of Non-Identical Data Distribution for Federated Visual Classification.
\newblock \emph{ArXiv}, abs/1909.06335.

\bibitem[{Hu, Wu, and Huang(2023)}]{hu2023beyond}
Hu, Z.; Wu, X.; and Huang, H. 2023.
\newblock Beyond Lipschitz smoothness: a tighter analysis for nonconvex optimization.
\newblock In \emph{International Conference on Machine Learning}, 13652--13678. PMLR.

\bibitem[{{Huang} et~al.(2017){Huang}, {Liu}, {Van Der Maaten}, and {Weinberger}}]{Huang2017DenseNet}
{Huang}, G.; {Liu}, Z.; {Van Der Maaten}, L.; and {Weinberger}, K.~Q. 2017.
\newblock Densely Connected Convolutional Networks.
\newblock In \emph{2017 IEEE Conference on Computer Vision and Pattern Recognition (CVPR)}, 2261--2269.

\bibitem[{Jee~Cho, Wang, and Joshi(2022)}]{cho22biased_selection}
Jee~Cho, Y.; Wang, J.; and Joshi, G. 2022.
\newblock Towards Understanding Biased Client Selection in Federated Learning.
\newblock In Camps-Valls, G.; Ruiz, F. J.~R.; and Valera, I., eds., \emph{Proceedings of The 25th International Conference on Artificial Intelligence and Statistics}, volume 151 of \emph{Proceedings of Machine Learning Research}, 10351--10375. PMLR.

\bibitem[{Kairouz et~al.(2021)Kairouz, McMahan, Avent, Bellet, Bennis, Bhagoji, Bonawitz, Charles, Cormode, Cummings, D'Oliveira, Eichner, Rouayheb, Evans, Gardner, Garrett, Gascón, Ghazi, Gibbons, Gruteser, Harchaoui, He, He, Huo, Hutchinson, Hsu, Jaggi, Javidi, Joshi, Khodak, Konecný, Korolova, Koushanfar, Koyejo, Lepoint, Liu, Mittal, Mohri, Nock, Özgür, Pagh, Qi, Ramage, Raskar, Raykova, Song, Song, Stich, Sun, Suresh, Tramèr, Vepakomma, Wang, Xiong, Xu, Yang, Yu, Yu, and Zhao}]{Kairouz21AdvancesProblems}
Kairouz, P.; McMahan, H.~B.; Avent, B.; Bellet, A.; Bennis, M.; Bhagoji, A.~N.; Bonawitz, K.~A.; Charles, Z.; Cormode, G.; Cummings, R.; D'Oliveira, R. G.~L.; Eichner, H.; Rouayheb, S.~E.; Evans, D.; Gardner, J.; Garrett, Z.; Gascón, A.; Ghazi, B.; Gibbons, P.~B.; Gruteser, M.; Harchaoui, Z.; He, C.; He, L.; Huo, Z.; Hutchinson, B.; Hsu, J.; Jaggi, M.; Javidi, T.; Joshi, G.; Khodak, M.; Konecný, J.; Korolova, A.; Koushanfar, F.; Koyejo, S.; Lepoint, T.; Liu, Y.; Mittal, P.; Mohri, M.; Nock, R.; Özgür, A.; Pagh, R.; Qi, H.; Ramage, D.; Raskar, R.; Raykova, M.; Song, D.; Song, W.; Stich, S.~U.; Sun, Z.; Suresh, A.~T.; Tramèr, F.; Vepakomma, P.; Wang, J.; Xiong, L.; Xu, Z.; Yang, Q.; Yu, F.~X.; Yu, H.; and Zhao, S. 2021.
\newblock Advances and Open Problems in Federated Learning.
\newblock \emph{Found. Trends Mach. Learn.}, 14(1-2): 1--210.

\bibitem[{Karimireddy et~al.(2020)Karimireddy, Kale, Mohri, Reddi, Stich, and Suresh}]{karimireddy2020scaffold}
Karimireddy, S.~P.; Kale, S.; Mohri, M.; Reddi, S.; Stich, S.; and Suresh, A.~T. 2020.
\newblock Scaffold: Stochastic controlled averaging for federated learning.
\newblock In \emph{International Conference on Machine Learning}, 5132--5143. PMLR.

\bibitem[{Khanduri et~al.(2021)Khanduri, Sharma, Yang, Hong, Liu, Rajawat, and Varshney}]{khanduri2021stem}
Khanduri, P.; Sharma, P.; Yang, H.; Hong, M.; Liu, J.; Rajawat, K.; and Varshney, P. 2021.
\newblock Stem: A stochastic two-sided momentum algorithm achieving near-optimal sample and communication complexities for federated learning.
\newblock \emph{Advances in Neural Information Processing Systems}, 34.

\bibitem[{Kidambi et~al.(2018)Kidambi, Netrapalli, Jain, and Kakade}]{Kidambi18Insufficiency}
Kidambi, R.; Netrapalli, P.; Jain, P.; and Kakade, S.~M. 2018.
\newblock On the insufficiency of existing momentum schemes for Stochastic Optimization.
\newblock \emph{CoRR}, abs/1803.05591.

\bibitem[{Krizhevsky(2009)}]{Krizhevsky2009CIFAR}
Krizhevsky, A. 2009.
\newblock Learning Multiple Layers of Features from Tiny Images.

\bibitem[{Krizhevsky, Sutskever, and Hinton(2012)}]{Krizhevsky12ImageNet}
Krizhevsky, A.; Sutskever, I.; and Hinton, G.~E. 2012.
\newblock ImageNet Classification with Deep Convolutional Neural Networks.
\newblock 1097--1105.

\bibitem[{Lessard, Recht, and Packard(2014)}]{Lessard14SNV}
Lessard, L.; Recht, B.; and Packard, A. 2014.
\newblock Analysis and Design of Optimization Algorithms via Integral Quadratic Constraints.
\newblock \emph{SIAM Journal on Optimization}, 26.

\bibitem[{Li et~al.(2022)Li, Diao, Chen, and He}]{li2022NIIDBenchmark}
Li, Q.; Diao, Y.; Chen, Q.; and He, B. 2022.
\newblock Federated Learning on Non-IID Data Silos: An Experimental Study.
\newblock In \emph{IEEE International Conference on Data Engineering}.

\bibitem[{Li, He, and Song(2021)}]{li2021model}
Li, Q.; He, B.; and Song, D. 2021.
\newblock Model-Contrastive Federated Learning.
\newblock In \emph{Proceedings of the IEEE/CVF Conference on Computer Vision and Pattern Recognition}.

\bibitem[{Li et~al.(2020{\natexlab{a}})Li, Sahu, Zaheer, Sanjabi, Talwalkar, and Smith}]{Li20FedProx}
Li, T.; Sahu, A.~K.; Zaheer, M.; Sanjabi, M.; Talwalkar, A.; and Smith, V. 2020{\natexlab{a}}.
\newblock Federated Optimization in Heterogeneous Networks.
\newblock In Dhillon, I.; Papailiopoulos, D.; and Sze, V., eds., \emph{Proceedings of Machine Learning and Systems}, volume~2, 429--450.

\bibitem[{Li et~al.(2020{\natexlab{b}})Li, Huang, Yang, Wang, and Zhang}]{Li2020Fed-Non-IID}
Li, X.; Huang, K.; Yang, W.; Wang, S.; and Zhang, Z. 2020{\natexlab{b}}.
\newblock On the Convergence of FedAvg on Non-IID Data.
\newblock In \emph{International Conference on Learning Representations}.

\bibitem[{Lian et~al.(2015)Lian, Huang, Li, and Liu}]{Lian15Asynchronous}
Lian, X.; Huang, Y.; Li, Y.; and Liu, J. 2015.
\newblock Asynchronous Parallel Stochastic Gradient for Nonconvex Optimization.
\newblock In \emph{Proceedings of the 28th International Conference on Neural Information Processing Systems - Volume 2}, NIPS'15, 2737–2745. Cambridge, MA, USA: MIT Press.

\bibitem[{Lin et~al.(2020)Lin, Stich, Patel, and Jaggi}]{lin2020dont}
Lin, T.; Stich, S.~U.; Patel, K.~K.; and Jaggi, M. 2020.
\newblock Don't Use Large Mini-batches, Use Local {SGD}.
\newblock In \emph{ICLR - International Conference on Learning Representations}.

\bibitem[{Liu, Gao, and Yin(2020)}]{liu2020improved}
Liu, Y.; Gao, Y.; and Yin, W. 2020.
\newblock An Improved Analysis of Stochastic Gradient Descent with Momentum.
\newblock arXiv:2007.07989.

\bibitem[{Ma and Yarats(2019)}]{ma2018quasihyperbolic}
Ma, J.; and Yarats, D. 2019.
\newblock Quasi-hyperbolic momentum and Adam for deep learning.
\newblock In \emph{International Conference on Learning Representations}.

\bibitem[{McMahan et~al.(2017)McMahan, Moore, Ramage, Hampson, and y~Arcas}]{McMahan2017FedAvg}
McMahan, H.~B.; Moore, E.; Ramage, D.; Hampson, S.; and y~Arcas, B.~A. 2017.
\newblock Communication-Efficient Learning of Deep Networks from Decentralized Data.
\newblock In \emph{International Conference on Artificial Intelligence and Statistics}.

\bibitem[{Mnih et~al.(2013)Mnih, Kavukcuoglu, Silver, Graves, Antonoglou, Wierstra, and Riedmiller}]{Mnih2013PlayingAW}
Mnih, V.; Kavukcuoglu, K.; Silver, D.; Graves, A.; Antonoglou, I.; Wierstra, D.; and Riedmiller, M.~A. 2013.
\newblock Playing Atari with Deep Reinforcement Learning.
\newblock \emph{ArXiv}, abs/1312.5602.

\bibitem[{Nguyen et~al.(2021)Nguyen, Malik, Zhan, Yousefpour, Rabbat, Malek, and Huba}]{Nguyen2021FedBuff}
Nguyen, J.; Malik, K.; Zhan, H.; Yousefpour, A.; Rabbat, M.~G.; Malek, M.; and Huba, D. 2021.
\newblock Federated Learning with Buffered Asynchronous Aggregation.
\newblock In \emph{International Conference on Artificial Intelligence and Statistics}.

\bibitem[{Nishio and Yonetani(2018)}]{Nishio2018ClientSelection}
Nishio, T.; and Yonetani, R. 2018.
\newblock Client Selection for Federated Learning with Heterogeneous Resources in Mobile Edge.
\newblock \emph{ICC 2019 - 2019 IEEE International Conference on Communications (ICC)}, 1--7.

\bibitem[{Reddi et~al.(2020)Reddi, Charles, Zaheer, Garrett, Rush, Kone{\v{c}}n{\`y}, Kumar, and McMahan}]{reddi2020adaptive}
Reddi, S.; Charles, Z.; Zaheer, M.; Garrett, Z.; Rush, K.; Kone{\v{c}}n{\`y}, J.; Kumar, S.; and McMahan, H.~B. 2020.
\newblock Adaptive federated optimization.
\newblock \emph{arXiv preprint arXiv:2003.00295}.

\bibitem[{Reddi, Kale, and Kumar(2018)}]{reddi18adam_convergence}
Reddi, S.~J.; Kale, S.; and Kumar, S. 2018.
\newblock On the Convergence of Adam and Beyond.
\newblock In \emph{International Conference on Learning Representations}.

\bibitem[{Ribero and Vikalo(2020)}]{Ribero2020clientsampling}
Ribero, M.; and Vikalo, H. 2020.
\newblock Communication-Efficient Federated Learning via Optimal Client Sampling.
\newblock \emph{ArXiv}, abs/2007.15197.

\bibitem[{Rothchild et~al.(2020)Rothchild, Panda, Ullah, Ivkin, Stoica, Braverman, Gonzalez, and Arora}]{rothchild20fetchsgd}
Rothchild, D.; Panda, A.; Ullah, E.; Ivkin, N.; Stoica, I.; Braverman, V.; Gonzalez, J.; and Arora, R. 2020.
\newblock FetchSGD: Communication-Efficient Federated Learning with Sketching.
\newblock In \emph{Proceedings of the 37th International Conference on Machine Learning}, ICML'20. JMLR.org.

\bibitem[{Simonyan and Zisserman(2015)}]{Simonyan14VGG}
Simonyan, K.; and Zisserman, A. 2015.
\newblock Very Deep Convolutional Networks for Large-Scale Image Recognition.
\newblock In \emph{3rd International Conference on Learning Representations, {ICLR} 2015, San Diego, CA, USA, May 7-9, 2015, Conference Track Proceedings}.

\bibitem[{{Smith}(2017)}]{Smith17Cyclic}
{Smith}, L.~N. 2017.
\newblock Cyclical Learning Rates for Training Neural Networks.
\newblock In \emph{2017 IEEE Winter Conference on Applications of Computer Vision (WACV)}, 464--472.

\bibitem[{Smith and Le(2018)}]{Smith18Bayesian}
Smith, S.; and Le, Q.~V. 2018.
\newblock A Bayesian Perspective on Generalization and Stochastic Gradient Descent.

\bibitem[{Smith, Kindermans, and Le(2018)}]{Smith18DontDecay}
Smith, S.~L.; Kindermans, P.-J.; and Le, Q.~V. 2018.
\newblock Don't Decay the Learning Rate, Increase the Batch Size.
\newblock In \emph{International Conference on Learning Representations}.

\bibitem[{Sun et~al.(2022)Sun, Huai, Jha, and Zhang}]{Sun22Hyperparameters}
Sun, J.; Huai, M.; Jha, K.; and Zhang, A. 2022.
\newblock Demystify Hyperparameters for Stochastic Optimization with Transferable Representations.
\newblock In \emph{Proceedings of the 28th ACM SIGKDD Conference on Knowledge Discovery and Data Mining}, KDD '22, 1706–1716. New York, NY, USA: Association for Computing Machinery.
\newblock ISBN 9781450393850.

\bibitem[{Sun, Sinha, and Zhang(2023)}]{Sun23Diffusion}
Sun, J.; Sinha, S.; and Zhang, A. 2023.
\newblock Enhance Diffusion to Improve Robust Generalization.
\newblock In \emph{Proceedings of the 29th ACM SIGKDD Conference on Knowledge Discovery and Data Mining}, KDD '23, 2083–2095. New York, NY, USA: Association for Computing Machinery.
\newblock ISBN 9798400701030.

\bibitem[{Sun et~al.(2021)Sun, Yang, Xun, and Zhang}]{sun21stagewise}
Sun, J.; Yang, Y.; Xun, G.; and Zhang, A. 2021.
\newblock A Stagewise Hyperparameter Scheduler to Improve Generalization.
\newblock In \emph{Proceedings of the 27th ACM SIGKDD Conference on Knowledge Discovery \& Data Mining}, KDD '21, 1530–1540. New York, NY, USA: Association for Computing Machinery.
\newblock ISBN 9781450383325.

\bibitem[{Sun et~al.(2023)Sun, Yang, Xun, and Zhang}]{Sun23TKDD}
Sun, J.; Yang, Y.; Xun, G.; and Zhang, A. 2023.
\newblock Scheduling Hyperparameters to Improve Generalization: From Centralized SGD to Asynchronous SGD.
\newblock \emph{ACM Trans. Knowl. Discov. Data}, 17(2).

\bibitem[{Suo et~al.(2019)Suo, Yao, Xun, Sun, and Zhang}]{SuoICHI19}
Suo, Q.; Yao, L.; Xun, G.; Sun, J.; and Zhang, A. 2019.
\newblock Recurrent Imputation for Multivariate Time Series with Missing Values.
\newblock In \emph{2019 {IEEE} International Conference on Healthcare Informatics, {ICHI} 2019, Xi'an, China, June 10-13, 2019}, 1--3. {IEEE}.

\bibitem[{Sutskever et~al.(2013)Sutskever, Martens, Dahl, and Hinton}]{Sutskever13Init}
Sutskever, I.; Martens, J.; Dahl, G.; and Hinton, G. 2013.
\newblock On the Importance of Initialization and Momentum in Deep Learning.
\newblock In \emph{Proceedings of the 30th International Conference on International Conference on Machine Learning - Volume 28}, ICML'13, III–1139–III–1147.

\bibitem[{{Van Scoy}, {Freeman}, and {Lynch}(2018)}]{VanScoy18Triple}
{Van Scoy}, B.; {Freeman}, R.~A.; and {Lynch}, K.~M. 2018.
\newblock The Fastest Known Globally Convergent First-Order Method for Minimizing Strongly Convex Functions.
\newblock \emph{IEEE Control Systems Letters}, 2(1): 49--54.

\bibitem[{Wang et~al.(2020)Wang, Liu, Liang, Joshi, and Poor}]{Wang20FedNova}
Wang, J.; Liu, Q.; Liang, H.; Joshi, G.; and Poor, H.~V. 2020.
\newblock Tackling the Objective Inconsistency Problem in Heterogeneous Federated Optimization.
\newblock In \emph{Proceedings of the 34th International Conference on Neural Information Processing Systems}, NIPS'20. Red Hook, NY, USA: Curran Associates Inc.
\newblock ISBN 9781713829546.

\bibitem[{Wang and Ji(2022)}]{wang2022arbitraryparticipation}
Wang, S.; and Ji, M. 2022.
\newblock A Unified Analysis of Federated Learning with Arbitrary Client Participation.
\newblock In Oh, A.~H.; Agarwal, A.; Belgrave, D.; and Cho, K., eds., \emph{Advances in Neural Information Processing Systems}.

\bibitem[{Wang, Magn{\'u}sson, and Johansson(2021)}]{wang21stepdecay}
Wang, X.; Magn{\'u}sson, S.; and Johansson, M. 2021.
\newblock On the Convergence of Step Decay Step-Size for Stochastic Optimization.
\newblock In Beygelzimer, A.; Dauphin, Y.; Liang, P.; and Vaughan, J.~W., eds., \emph{Advances in Neural Information Processing Systems}.

\bibitem[{Wang, Lin, and Chen(2022)}]{wang22adaptive}
Wang, Y.; Lin, L.; and Chen, J. 2022.
\newblock Communication-Efficient Adaptive Federated Learning.
\newblock In \emph{Proceedings of the 39th International Conference on Machine Learning}, volume 162 of \emph{Proceedings of Machine Learning Research}, 22802--22838. PMLR.

\bibitem[{Wilson et~al.(2017)Wilson, Roelofs, Stern, Srebro, and Recht}]{Wilson2017Generalization}
Wilson, A.~C.; Roelofs, R.; Stern, M.; Srebro, N.; and Recht, B. 2017.
\newblock The Marginal Value of Adaptive Gradient Methods in Machine Learning.
\newblock In \emph{Advances in Neural Information Processing Systems 30}, 4148--4158. Curran Associates, Inc.

\bibitem[{Wu et~al.(2023{\natexlab{a}})Wu, Huang, Hu, and Huang}]{wu2023faster}
Wu, X.; Huang, F.; Hu, Z.; and Huang, H. 2023{\natexlab{a}}.
\newblock Faster adaptive federated learning.
\newblock In \emph{Proceedings of the AAAI Conference on Artificial Intelligence}, volume~37, 10379--10387.

\bibitem[{Wu et~al.(2023{\natexlab{b}})Wu, Sun, Hu, Li, Zhang, and Huang}]{wu2023federated}
Wu, X.; Sun, J.; Hu, Z.; Li, J.; Zhang, A.; and Huang, H. 2023{\natexlab{b}}.
\newblock Federated Conditional Stochastic Optimization.
\newblock In \emph{Thirty-seventh Conference on Neural Information Processing Systems}.

\bibitem[{Wu et~al.(2023{\natexlab{c}})Wu, Sun, Hu, Zhang, and Huang}]{wu2023solving}
Wu, X.; Sun, J.; Hu, Z.; Zhang, A.; and Huang, H. 2023{\natexlab{c}}.
\newblock Solving a Class of Non-Convex Minimax Optimization in Federated Learning.
\newblock In \emph{Thirty-seventh Conference on Neural Information Processing Systems}.

\bibitem[{Wu et~al.(2023{\natexlab{d}})Wu, Hu, Zhang, and Huang}]{Wu2023DiPmarkAS}
Wu, Y.; Hu, Z.; Zhang, H.; and Huang, H. 2023{\natexlab{d}}.
\newblock DiPmark: A Stealthy, Efficient and Resilient Watermark for Large Language Models.
\newblock \emph{ArXiv}, abs/2310.07710.

\bibitem[{Xie, Koyejo, and Gupta(2019)}]{Xie2019AsynchronousFO}
Xie, C.; Koyejo, O.; and Gupta, I. 2019.
\newblock Asynchronous Federated Optimization.
\newblock \emph{ArXiv}, abs/1903.03934.

\bibitem[{Xun et~al.(2020)Xun, Jha, Sun, and Zhang}]{Xun2020CorrelationNF}
Xun, G.; Jha, K.; Sun, J.; and Zhang, A. 2020.
\newblock Correlation Networks for Extreme Multi-label Text Classification.
\newblock \emph{Proceedings of the 26th ACM SIGKDD International Conference on Knowledge Discovery \& Data Mining}.

\bibitem[{Yan et~al.(2020)Yan, Niu, Ding, Zheng, Wu, Chen, Tang, and Wu}]{Yan2020DistributedClient}
Yan, Y.; Niu, C.; Ding, Y.; Zheng, Z.; Wu, F.; Chen, G.; Tang, S.; and Wu, Z. 2020.
\newblock Distributed Non-Convex Optimization with Sublinear Speedup under Intermittent Client Availability.
\newblock \emph{ArXiv}, abs/2002.07399.

\bibitem[{Yang, Fang, and Liu(2021)}]{yang2021achieving}
Yang, H.; Fang, M.; and Liu, J. 2021.
\newblock Achieving Linear Speedup with Partial Worker Participation in Non-{IID} Federated Learning.
\newblock In \emph{International Conference on Learning Representations}.

\bibitem[{Yang et~al.(2021)Yang, Zhang, Khanduri, and Liu}]{Yang2021AnarchicFL}
Yang, H.; Zhang, X.; Khanduri, P.; and Liu, J. 2021.
\newblock Anarchic Federated Learning.
\newblock In \emph{International Conference on Machine Learning}.

\bibitem[{Yurochkin et~al.(2019)Yurochkin, Agarwal, Ghosh, Greenewald, Hoang, and Khazaeni}]{Yurochkin2019BayesianNF}
Yurochkin, M.; Agarwal, M.; Ghosh, S.~S.; Greenewald, K.~H.; Hoang, T.~N.; and Khazaeni, Y. 2019.
\newblock Bayesian Nonparametric Federated Learning of Neural Networks.
\newblock In \emph{International Conference on Machine Learning}.

\bibitem[{Zhang et~al.(2020)Zhang, Karimireddy, Veit, Kim, Reddi, Kumar, and Sra}]{Zhang20Adam_Attention}
Zhang, J.; Karimireddy, S.~P.; Veit, A.; Kim, S.; Reddi, S.; Kumar, S.; and Sra, S. 2020.
\newblock Why are Adaptive Methods Good for Attention Models?
\newblock In Larochelle, H.; Ranzato, M.; Hadsell, R.; Balcan, M.; and Lin, H., eds., \emph{Advances in Neural Information Processing Systems}, volume~33, 15383--15393. Curran Associates, Inc.

\bibitem[{Zhang, Choromańska, and LeCun(2014)}]{Zhang2014DeepElasticAvg}
Zhang, S.; Choromańska, A.; and LeCun, Y. 2014.
\newblock Deep learning with Elastic Averaging SGD.
\newblock In \emph{NIPS}.

\bibitem[{Zhao et~al.(2018)Zhao, Li, Lai, Suda, Civin, and Chandra}]{zhao2018federated-noniid}
Zhao, Y.; Li, M.; Lai, L.; Suda, N.; Civin, D.; and Chandra, V. 2018.
\newblock Federated learning with non-iid data.
\newblock \emph{arXiv preprint arXiv:1806.00582}.

\bibitem[{Zheng et~al.(2017)Zheng, Meng, Wang, Chen, Yu, Ma, and Liu}]{Zheng17ASGD}
Zheng, S.; Meng, Q.; Wang, T.; Chen, W.; Yu, N.; Ma, Z.-M.; and Liu, T.-Y. 2017.
\newblock Asynchronous Stochastic Gradient Descent with Delay Compensation.
\newblock In \emph{Proceedings of the 34th International Conference on Machine Learning - Volume 70}, ICML'17, 4120–4129. JMLR.org.

\end{thebibliography}
